\documentclass{article}

\usepackage[preprint, nonatbib]{neurips_2026}
\usepackage{comment}

\usepackage[utf8]{inputenc} % allow utf-8 input
\usepackage[T1]{fontenc}    % use 8-bit T1 fonts
\usepackage{hyperref}       % hyperlinks

\usepackage{url}            % simple URL typesetting
\usepackage{booktabs}       % professional-quality tables
\usepackage{amsfonts}       % blackboard math symbols
\usepackage{nicefrac}       % compact symbols for 1/2, etc.
\usepackage{microtype}      % microtypography
\usepackage{xcolor}         % colors
\usepackage{amsmath, amssymb, amsthm}
\usepackage{adjustbox}
\usepackage{array}
\usepackage{caption}
\usepackage{enumerate}
\usepackage{graphicx}
\usepackage{enumitem}
\usepackage[numbers]{natbib}
\usepackage{multirow}
\usepackage[table]{xcolor}
\usepackage{etoc} 

\usepackage[nameinlink,noabbrev]{cleveref}
\crefformat{section}{\S#2#1#3}
\crefmultiformat{section}{\S#2#1#3}{ and \S#2#1#3}{, \S#2#1#3}{ and \S#2#1#3}
\crefrangeformat{section}{\S#3#1#4--\S#5#2#6}
\crefname{equation}{Eq.}{Eqs.}
\crefname{figure}{Fig.}{Figs.}
  
\definecolor{royalblue}{rgb}{.25, .41, .88}

\hypersetup{colorlinks,
citecolor=royalblue,
linkcolor=royalblue,
urlcolor=royalblue}

\usepackage{amsthm}
\theoremstyle{definition}
\newtheorem{definition}{Definition}

\definecolor{baserow}{RGB}{235,242,252}   % soft blue for base-model rows
\definecolor{grpcolor}{gray}{0.93}         % light gray for column group shading

\usepackage{nicefrac}
\usepackage{pgfplots}
\usepackage{pgfplotstable}
\usepgfplotslibrary{groupplots, fillbetween}
\usetikzlibrary{patterns, arrows.meta}
\usepackage{ulem}
\pgfplotsset{compat=newest}
\pgfplotsset{every tick label/.append style={font=\footnotesize}}

\definecolor{colorRandom}{HTML}{DC267F}     % Random placement / knockout
\definecolor{colorFisher}{HTML}{648FFF}     % F-hat-scored (Fisher) circuit
\definecolor{colorSignal}{HTML}{785EF0}     % S-hat-scored (signal) circuit
\definecolor{colorFullLoRA}{HTML}{FE6100}   % Full LoRA baseline
\definecolor{colorBase}{HTML}{555555}       % Base (pre-trained) model

\pgfplotsset{
    tinylora/.style={
        enlarge y limits=0.02,
        every axis plot/.append style={semithick},
        grid=both,
        grid style={gray!15},
        tick label style={font=\small},
        label style={font=\small},
        title style={font=\small\bfseries, yshift=-4pt},
        legend style={
            font=\scriptsize,
            fill=white,
            fill opacity=1,
            draw=black,
            text opacity=1,
            row sep=2pt,
        },
    },
}

\usepgfplotslibrary{colorbrewer}

\usepackage[most,skins,theorems]{tcolorbox}
\usepackage{mathtools}
\tcbset{
 aibox/.style={
 width=\linewidth,
 top=8pt,
 bottom=4pt,
 colback=blue!6!white,
 colframe=black,
 colbacktitle=black,
 enhanced,
 center,
 attach boxed title to top left={yshift=-0.1in,xshift=0.15in},
 boxed title style={boxrule=0pt,colframe=white,},
 }
}
\newtcolorbox{dmbox}[2][]{aibox,title=#2,#1}

\title{Not How Many, But Which: Parameter Placement in Low-Rank Adaptation}

\author{                                 
    Arijit Sehanobish\thanks{Equal contribution.} \quad Charles Lovering\footnotemark[1] \\
    Kensho Technologies\\                                      
    \texttt{\{arijit.sehanobish, charles.lovering\}@kensho.com}
}

\begin{document}

\maketitle

\begin{abstract}
We study the \textit{parameter placement problem}: given a fixed budget of $k$ trainable entries within the B matrix of a LoRA adapter (A frozen), does the choice of which $k$ matter? Under supervised fine-tuning, random and informed subsets achieve comparable performance. Under GRPO on base models, random placement fails to improve over the base model, while gradient-informed placement recovers standard LoRA accuracy. This regime dependence traces to gradient structure: SFT gradients are low-rank and directionally stable, so any subset accumulates coherent updates; GRPO gradients are high-rank and near-orthogonal across steps, so only elements with consistently signed gradients retain the learning signal. Our scoring procedure identifies these critical parameters in under 10 seconds at less than 0.5\% of training cost. Selected parameters concentrate on residual-stream–writing projections (V, O, Down), stable across model families and scales (1.5B--8B).
\end{abstract}

\section{Introduction}
\label{sec:intro}
Large language models are increasingly ubiquitous, underpinning applications from code generation to scientific reasoning, but adapting them to new tasks remains costly: full fine-tuning updates billions of parameters and requires substantial compute.  Parameter-efficient fine-tuning (PEFT) methods address this by training a smaller set of parameters while freezing the rest. Among these, Low-Rank Adaptation (LoRA; \cite{hu2022lora}) has become widely adopted, decomposing weight updates into low-rank matrices $BA$ and optimizing only the adapter parameters.  

Recent works shows that number of trainable parameters in such adapters can be dramatically reduced, via shared random bases \citep{kopiczko2024vera}, spectral parameterizations \citep{gao2024fourierft}, learned scaling vectors \citep{kalajdzievski2023rslora}, or low-dimensional projections \citep{morris2026tinylora}. These results suggest that strong performance can be achieved with surprisingly small parameter budgets. However, they implicitly assume that performance depends primarily on how many parameters are trained, rather than which ones. We revisit this assumption:

\begin{dmbox}{Our Research Question}
 We study what we term the \textbf{Parameter Placement Problem}: \textbf{given a fixed budget of trainable parameters within a LoRA adapter, how does parameter choice affect downstream performance?} Concretely, in a standard LoRA layer $\mathbf{W}_0 \mathbf{x} + \mathbf{BAx}$, where $\mathbf{B} \in \mathbb{R}^{h \times r}$ is initialized to zero and $\mathbf{A} \in \mathbb{R}^{r \times d}$ is drawn randomly, we keep $\mathbf{A}$ frozen, and restrict training to a subset of $k$ entries of $\mathbf{B}$. This isolates parameter selection from parameter count, allowing us to ask whether different subsets of equal size behave differently during optimization. \end{dmbox}

Donoway et. al \citep{donoway2025elicitation} showed that by searching across hyperparameter sweeps and seeds, subsets of LoRA neurons can yield most of the performance gain. We aimed to directly recover these ``circuits''.

We made two empirical observations, which align with existing research on the nature of the optimization signal in LoRA \citep{schulman2025lora}, which together motivate our parameter selection proposal, and explain our results. First, SFT gradients are highly aligned across training steps and admit a low-dimensional structure, so that most parameter subsets receive a coherent update direction. Second, in contrast, RL gradients exhibit substantially higher variance, lower magnitude, and weak alignment across steps. In this regime, many parameter subsets experience updates that largely cancel over time, while a small subset of parameters with consistently large gradients accumulates meaningful updates.

Motivated by these observations, we propose a simple gradient-based selection procedure that identifies parameters with consistently high gradient magnitude during a short scoring phase, producing stable subsets across data samples with negligible overhead. To isolate the role of gradient noise, we study a controlled synthetic setting where the signal-to-noise ratio can be varied directly. Consistent with our hypothesis, when gradients are clean, most parameter subsets learn effectively; under high noise, performance depends on selecting  parameters aligned with the dominant signal.

Our contributions are as follows:
\begin{itemize}[leftmargin=*,itemsep=1pt]
\item We formalize the parameter placement problem for LoRA, separating the effect of parameter count from parameter identity and showing the critical impact under RL (\Cref{sec:method}). 

\item We propose a lightweight gradient-based selection procedure that identifies effective parameter subsets in seconds and yields competitive performance at extreme parameter budgets (\Cref{sec:experiments}).

\item We conduct a detailed investigation on the location, causal effects and the training dynamics of these parameters (\Cref{sec:analysis_main}).
\end{itemize}

\paragraph{Organization.}
\Cref{sec:related} reviews related work; 
\Cref{sec:method} introduces our methodology, and analyzes when placement matters (\Cref{sec:circuit_discovery,sec:synthetic}); 
\Cref{sec:experiments} presents experiments on fine-tuning, reinforcement learning, and out-of-distribution generalization, followed by a discussion of circuit structure (\Cref{sec:analysis_main}).

\section{Related Work}
\label{sec:related}

Our paper touches on several lines of research, discussed below. More background is in \Cref{app:related}.

\paragraph{Low-rank adaptation and its variants.}
LoRA \citep{hu2022lora} decomposes weight updates into low-rank matrices $BA$, training only the adapter while freezing the pre-trained model. Subsequent work seeks to improve LoRA along several axes: initialization \citep{meng2024pissa, yang2024corda}, training dynamics \citep{hayou2024loraplus, liu2024dora, kalajdzievski2023rslora}, and rank allocation \citep{zhang2023adalora, valipour2023dylora, ding2023sora}. Instead of modifying LoRA, we study the complementary question: given a fixed adapter, \textit{which elements} should receive gradient updates.

\paragraph{Parameter budget and fixed-basis methods.}
Several methods reduce adapter size by fixing most parameters and training a small residual.
  VeRA \citep{kopiczko2024vera} shares frozen random $A$ and $B$ across all layers and trains per-layer scaling vectors.  FourierFT \citep{gao2024fourierft} represents updates via learnable spectral coefficient over a fixed Fourier basis.  RandLoRA~\citep{albert2025randlora} extends VeRA with element-wise learned scaling to recover full-rank expressiveness.  LoRA-XS \citep{balazy2024loraxs} inserts a tiny $r \times r$ trainable matrix between frozen SVD-initialized factors.  Most recently, \citet{morris2026tinylora} show that RL can succeed with as few as 13 trainable parameters on \textit{instruction-tuned} models, using random low-dimensional projections. 
  
  These methods share a common structure: fix a basis (random, Fourier, SVD) and learn a small set of coefficients.  Our approach differs in two respects.  First, we do not change the LoRA parameterization, we use standard $\mathbf{BA}$ with a standard optimizer and simply mask which elements of $\mathbf{B}$ receive gradients. Second, our selection is informed: we use gradient statistics from the training distribution to choose elements, rather than relying on a fixed or random basis.

\paragraph{Importance scoring and parameter selection.}
Using gradient or curvature information to identify important parameters has a long history, from Optimal Brain Damage \citep{lecun1990obd} and Optimal Brain Surgeon \citep{hassibi1993obs} to recent methods like Wanda \citep{sun2024wanda} and SparseGPT \citep{frantar2023sparsegpt}.  Movement pruning \citep{sanh2020movement} selects parameters during fine-tuning based on gradient-weight products.
Other methods (SNIP, GRASP, SynFlow) \cite{DBLP:journals/corr/abs-1810-02340,DBLP:journals/corr/abs-2002-07376,DBLP:journals/corr/abs-2006-05467} addresses issues like layer collapse. The lottery ticket hypothesis \citep{frankle2019lottery} demonstrates that sparse trainable subnetworks exist within dense networks.  SPT \citep{he2023spt} uses sensitivity scores to decide where to place adapters; we assume adapters are placed and select which elements within them to train.  Our circuit scores are instances of this family, specialized to the LoRA setting where $\mathbf{B} = \mathbf{0}$ at initialization provides a clean starting point for gradient-based scoring.

\paragraph{Elicitation and minimal adaptation.}
\citet{aghajanyan2021intrinsic} show that fine-tuning has low intrinsic dimensionality: randomly projected subspaces of a few   hundred dimensions suffice.  \citet{donoway2025elicitation} find that training as few as 10--100 randomly chosen parameters can recover 50\% of the performance gap, framing adaptation as elicitation of latent capabilities.  Both studies use random parameter selection.  Our results are consistent with this picture for SFT, where random placement remains a reasonable baseline.  However, we show that under RL on base models, random placement fails entirely, suggesting elicitation via RL is harder and requires from informed parameter selection.

\paragraph{Reinforcement learning for reasoning.}
% eliminates the critic model by 
GRPO \citep{shao2024deepseekmath} uses group-relative rewards to enable efficient RL fine-tuning.  DeepSeek-R1 \citep{deepseek2025r1} demonstrates that RL alone can elicit strong reasoning.  \citet{morris2026tinylora} show that RL succeeds with far fewer parameters than SFT on instruction-tuned models.  Our work complements these findings: on base models, RL with sparse adapters requires informed parameter placement, random placement, which suffices for SFT, fails under GRPO.

\section{Method}
\label{sec:method}

We formalize the parameter placement problem (\Cref{sec:problem_statement}), establish why the budget should target $\mathbf{B}$ rather than $\mathbf{A}$ (\Cref{sec:why_b}), describe a lightweight scoring method to select entries of $\mathbf{B}$ (\Cref{sec:circuit_discovery}), analyze when placement matters  (\Cref{sec:signal_retention}), and validate this analysis in a controlled setting (\Cref{sec:synthetic}).
\subsection{Problem Statement}
\label{sec:problem_statement}
Consider a model with LoRA-adapted layers $\{\mathbf{B}^{(\ell)}, \mathbf{A}^{(\ell)}\}_\ell$, where each layer outputs
$\mathbf{y}^{(\ell)} = \mathbf{W}_0^{(\ell)} \mathbf{x}^{(\ell)} + \mathbf{B}^{(\ell)} \mathbf{A}^{(\ell)} \mathbf{x}^{(\ell)}$.
We fix $\mathbf{A}^{(\ell)}$ and initialize $\mathbf{B}^{(\ell)} = 0$. We define the \textbf{parameter placement problem} under a global budget: given $k$ trainable parameters,
select  per-layer masks $\mathbf{m}^{(\ell)} \in \{0,1\}^{h_\ell \times r_\ell}$  such that
$\sum_\ell \|\mathbf{m}^{(\ell)}\|_0 = k$, and training only the selected entries
$\{\mathbf{m}^{(\ell)} \odot \mathbf{B}^{(\ell)}\}_\ell$ maximizes performance. We consider two strategies:
(1) \textbf{Random placement}, where $k$ entries are sampled uniformly across all layers, and
(2) \textbf{Informed placement}, where a scoring function scores each parameter across all layers and the top-$k$ global entries are selected.

\subsection{Why $\mathbf{B}$, Not $\mathbf{A}$}
\label{sec:why_b}

At initialization $\mathbf{B} = \mathbf{0}$, changes to $\mathbf{A}$ produce no change in output ($\mathbf{BAx} = \mathbf{0}$ regardless of $\mathbf{A}$) and receives zero gradient. If half the budget goes to $\mathbf{A}$, half the budget is idle until $\mathbf{B}$ isn't zero.  This is avoided by fixing $\mathbf{A}$ as a random projection and allocating the entire budget to $\mathbf{B}$ \citep{zhang2023lorafa, zhu2024asymmetry}. 
Empirically, $\mathbf{B}$-only training dominates training both $\mathbf{A}$ and $\mathbf{B}$ at matched budgets (\Cref{app:a_vs_ab}).

\subsection{Circuit Discovery}
\label{sec:circuit_discovery}

With the budget allocated to $\mathbf{B}$, we need a method to identify which of its $h \times r$ entries are most useful for learning.  We score each element using gradient statistics from $N$ forward-backward passes at initialization.  Since $\mathbf{B} = \mathbf{0}$, the model's predictions are identical to the pre-trained model, and the gradient $\nabla_{\mathbf{B}} \mathcal{L}$ directly measures each element's utility for reducing the loss.

For each $x_k$, let  $g_{k,ij} = [\nabla_\mathbf{B} \mathcal{L}(x_k)]_{ij}$ be per-element gradient.  We define two scoring functions:

  \begin{minipage}[t]{0.45\linewidth}
  \begin{equation}                                                   \hat{S}_{ij} = \left|\frac{1}{N}\sum_{k=1}^{N} g_{k,ij}\right|,
  \label{eq:s_score}
  \end{equation}
  \end{minipage}%                                                    \hfill
  \begin{minipage}[t]{0.45\linewidth}
  \begin{equation}                                                   \hat{F}_{ij} = \frac{1}{N}\sum_{k=1}^{N} g_{k,ij}^2.               \label{eq:f_score}
  \end{equation}
  \end{minipage}

The \textbf{Gradient magnitude score} $\hat{S}_{ij}$ (left, \Cref{eq:s_score}) estimates $|\mathbb{E}[g_{ij}]|$, the magnitude of the population mean gradient: elements with high $\hat{S}$ have gradients that are directionally consistent across examples, i.e. the loss surface has a persistent preference for how that element should change.

The \textbf{Diagonal Fisher score} $\hat{F}_{ij}$ (right, \Cref{eq:f_score}) estimates the diagonal of the empirical Fisher $\mathbb{E}[g_{ij}^2]$: elements where the loss is most sensitive to perturbations. The two scores are related by the bias-variance decomposition $\mathbb{E}[g_{ij}^2] = \mu_{ij}^2 + \sigma_{ij}^2$; Fisher captures both mean and variance, while $\hat{S}$ captures only the mean.  The distinction matters: an element can have high $\hat{F}$ but low $\hat{S}$ if its gradients are large but flip sign across examples. Given budget $k$, the \textbf{circuit} is the set of $k$ elements with the highest scores.  Both estimators converge at rate $O(1/\sqrt{N})$, and because score distributions are heavy-tailed (top elements have scores orders of magnitude above the median), small $N$ suffices (see \Cref{sec:exp_stability}).

\subsection{When Does Placement Matter?}
\label{sec:signal_retention}
Circuit discovery selects a subset of parameters, but this only improves performance when those parameters receive and accumulate useful gradient signal. In other words, placement matters when the chosen elements capture significant gradient mass at each step and do so consistently over training.

\paragraph{Signal retention under sparse masking.}
Let $g \in \mathbb{R}^d$ be the gradient over all $B$ elements (flattened), and let $m \in \{0,1\}^d$ be a mask with $\|m\|_0 = k$.  The effective update is $m \odot g$, retaining only the masked entries.  The fraction of gradient energy retained is:  $\rho(m, g) = \nicefrac{\|m \odot g\|^2}{\|g\|^2}.$

For a random mask: $\mathbb{E}[\rho] = k/d$.  For an informed mask selecting the top-$k$ elements by $|g_i|$: $\rho = \sum_{i \in \mathrm{top\text{-}k}} g_i^2 / \|g\|^2$, which can be much larger for heavy-tailed gradient.  Empirically, the top 0.05\% of $B$ elements capture $\sim$8--15\% of gradient energy, 160 to 300$\times$ amplification over random.

Informed placement achieves higher signal retention for SFT and GRPO, but random placement only succeeds for SFT (\Cref{sec:experiments}).  The missing factor is how updates accumulate across training steps.
\paragraph{Gradient stability determines accumulation.}
Over $T$ steps, $\Delta w_i = \eta \sum_{t} g_{t,i}$. When gradients align, updates accumulate coherently; when they are noisy, updates behave like a random walk and mostly cancel. As a result, only parameters with consistently large, same-signed gradients across steps undergo meaningful change. Under low alignment, most elements have near-zero accumulated signal, while a sparse subset remains large (heavy-tailed). Circuit discovery identifies this subset, whereas random placement samples mostly near-zero elements, explaining its failure under GRPO.

\begin{table}[t]
\centering
\caption{{\small{\textbf{Gradient structure under SFT vs.\ GRPO} (Qwen2.5-1.5B/3B/7B, MATH, matched gradient steps at initialization).  Across all three scales, SFT gradients are low-rank, directionally stable, and accumulate coherently.  GRPO gradients are high-rank, near-orthogonal across steps, and mostly cancel, only informed placement recovers the signal.  GSM8K results in \Cref{app:gradient_full}.}}}
\label{tab:gradient_properties}
\vspace{4pt}
\small
\setlength{\tabcolsep}{5pt}
\begin{tabular}{@{}l cc cc cc@{}}
\toprule
& \multicolumn{2}{c}{\textbf{Eff.\ Gradient Rank}} & \multicolumn{2}{c}{\textbf{Cosine Similarity}} & \multicolumn{2}{c}{\textbf{Accum.\ Efficiency}} \\
\cmidrule(lr){2-3} \cmidrule(lr){4-5} \cmidrule(lr){6-7}
\textbf{Model} & \textbf{SFT} & \textbf{GRPO} & \textbf{SFT} & \textbf{GRPO} & \textbf{SFT} & \textbf{GRPO} \\
\midrule
Qwen2.5-1.5B & 2.6 & 21.8 & 0.76 & 0.04 & 0.90 & 0.28 \\
Qwen2.5-3B   & 5.8 & 7.7  & 0.61 & $-$0.03 & 0.81 & 0.22 \\
Qwen2.5-7B   & 2.0 & 8.8  & 0.85 & $-$0.03 & 0.93 & 0.18 \\
\bottomrule
\end{tabular}
\end{table}

\Cref{tab:gradient_properties} quantifies this interaction for SFT and GRPO.  Under SFT, the gradient is nearly rank-1 and highly stable across steps ($\cos \approx 0.87$): 93\% of gradient energy accumulates coherently.  Any $k$-element subset projects onto the dominant direction with high probability, so placement has little effect.  Under GRPO, the gradient is rank-24, noisy ($\cos \approx 0.08$), and spread across many dimensions---despite per-step gradient norms that are comparable to SFT, only 34\% of gradient energy survives accumulation.  A random subset receives negligible projection of each signal dimension, and the noise causes individual elements to random-walk rather than accumulate displacement.

Placement importance depends on the gradient signal. SFT produces a concentrated, stable signal that most subsets may extract; GRPO produces a distributed, noisy signal only the right subset  extracts.

\subsection{Synthetic Validation}
\label{sec:synthetic}

The analysis predicts that placement matters when gradient signal is concentrated. We test this with a 2-layer MLP using LoRA under (i) a dense target (rank-2 residual, SFT-like) and (ii) a sparse target (5\% large entries, RL-like). Full LoRA trains both $\mathbf{A}$ and $\mathbf{B}$. For both random and informed placements $\mathbf{A}$ is kept frozen. See \Cref{app:synthetic} for details.

\begin{figure}[t]
\centering
% Auto-generated by pgf_synthetic.py — do not edit by hand.
% Usage: \input{figures/synthetic_placement.tex}
%   Requires: tinylora_preamble.tex, pgfplots, groupplots, fillbetween
\begin{tikzpicture}
\begin{groupplot}[
    tinylora,
    group style={
        group size=2 by 1,
        horizontal sep=1.05cm,
    },
    width=0.40\textwidth,
    height=0.35\textwidth,
    legend style={at={(1.05,0.5)}, anchor=west},
]

    \nextgroupplot[
        title={(a) Supervised (dense signal)},
        ylabel={Relative MSE\\($1$ = no adaptation)}, ylabel style={align=center},
        xlabel={Fraction of $B$ trained ($k/|B|$)},
        xmode=log,
        xmin=0.015, xmax=1.3,
        ymin=-0.1, ymax=1.1,
        ytick={0, 0.25, 0.5, 0.75, 1.0},
    ]
    \addplot[black, dashed, thin, forget plot, domain=0.01:1.1] {1.0};
        \addplot[draw=none, name path=uppersftrandom, forget plot]
            coordinates {(0.02,0.983007) (0.05,0.937714) (0.1,0.874710) (0.2,0.758267) (0.5,0.459548) (1.0,0.019809)};
        \addplot[draw=none, name path=lowersftrandom, forget plot]
            coordinates {(0.02,0.954609) (0.05,0.913537) (0.1,0.838466) (0.2,0.723170) (0.5,0.423401) (1.0,0.006074)};
        \addplot[colorRandom!20, forget plot]
            fill between[of=uppersftrandom and lowersftrandom];
        \addplot[colorRandom, mark=o, mark size=2.5pt, thick, forget plot]
            coordinates {(0.02,0.968808) (0.05,0.925626) (0.1,0.856588) (0.2,0.740719) (0.5,0.441475) (1.0,0.012942)};
        \addplot[draw=none, name path=uppersftinformed, forget plot]
            coordinates {(0.02,0.813359) (0.05,0.698904) (0.1,0.567214) (0.2,0.437283) (0.5,0.220156) (1.0,0.019809)};
        \addplot[draw=none, name path=lowersftinformed, forget plot]
            coordinates {(0.02,0.811317) (0.05,0.695542) (0.1,0.563337) (0.2,0.434739) (0.5,0.216047) (1.0,0.006074)};
        \addplot[colorFisher!20, forget plot]
            fill between[of=uppersftinformed and lowersftinformed];
        \addplot[colorFisher, mark=square*, mark size=2.5pt, thick, forget plot]
            coordinates {(0.02,0.812338) (0.05,0.697223) (0.1,0.565275) (0.2,0.436011) (0.5,0.218101) (1.0,0.012942)};
        \addplot[draw=none, name path=uppersftfull, forget plot]
            coordinates {(0.02,0.019809) (0.05,0.019809) (0.1,0.019809) (0.2,0.019809) (0.5,0.019809) (1.0,0.019809)};
        \addplot[draw=none, name path=lowersftfull, forget plot]
            coordinates {(0.02,0.006074) (0.05,0.006074) (0.1,0.006074) (0.2,0.006074) (0.5,0.006074) (1.0,0.006074)};
        \addplot[colorFullLoRA!20, forget plot]
            fill between[of=uppersftfull and lowersftfull];
        \addplot[colorFullLoRA, mark=triangle*, mark size=2.5pt, thick, forget plot]
            coordinates {(0.02,0.012942) (0.05,0.012942) (0.1,0.012942) (0.2,0.012942) (0.5,0.012942) (1.0,0.012942)};

    \nextgroupplot[
        title={(b) RL-like (concentrated signal)},
        ylabel={},
        xlabel={Fraction of $B$ trained ($k/|B|$)},
        xmode=log,
        xmin=0.015, xmax=1.3,
        ymin=-0.1, ymax=1.1,
        ytick={0, 0.25, 0.5, 0.75, 1.0},
    ]
    \addplot[black, dashed, thin, forget plot, domain=0.01:1.1] {1.0};
        \addplot[draw=none, name path=upperrlrandom, forget plot]
            coordinates {(0.02,0.990251) (0.05,0.957085) (0.1,0.904597) (0.2,0.734710) (0.5,0.465660) (1.0,0.008512)};
        \addplot[draw=none, name path=lowerrlrandom, forget plot]
            coordinates {(0.02,0.943285) (0.05,0.875132) (0.1,0.741468) (0.2,0.548306) (0.5,0.292384) (1.0,0.008183)};
        \addplot[colorRandom!20, forget plot]
            fill between[of=upperrlrandom and lowerrlrandom];
        \addplot[colorRandom, mark=o, mark size=2.5pt, thick]
            coordinates {(0.02,0.966768) (0.05,0.916109) (0.1,0.823032) (0.2,0.641508) (0.5,0.379022) (1.0,0.008348)};
        \addlegendentry{Random}
        \addplot[draw=none, name path=upperrlinformed, forget plot]
            coordinates {(0.02,0.206121) (0.05,0.076477) (0.1,0.044466) (0.2,0.026210) (0.5,0.016010) (1.0,0.008512)};
        \addplot[draw=none, name path=lowerrlinformed, forget plot]
            coordinates {(0.02,0.206054) (0.05,0.076331) (0.1,0.044248) (0.2,0.025946) (0.5,0.015731) (1.0,0.008183)};
        \addplot[colorFisher!20, forget plot]
            fill between[of=upperrlinformed and lowerrlinformed];
        \addplot[colorFisher, mark=square*, mark size=2.5pt, thick]
            coordinates {(0.02,0.206087) (0.05,0.076404) (0.1,0.044357) (0.2,0.026078) (0.5,0.015871) (1.0,0.008348)};
        \addlegendentry{Informed (circuit)}
        \addplot[draw=none, name path=upperrlfull, forget plot]
            coordinates {(0.02,0.008512) (0.05,0.008512) (0.1,0.008512) (0.2,0.008512) (0.5,0.008512) (1.0,0.008512)};
        \addplot[draw=none, name path=lowerrlfull, forget plot]
            coordinates {(0.02,0.008183) (0.05,0.008183) (0.1,0.008183) (0.2,0.008183) (0.5,0.008183) (1.0,0.008183)};
        \addplot[colorFullLoRA!20, forget plot]
            fill between[of=upperrlfull and lowerrlfull];
        \addplot[colorFullLoRA, mark=triangle*, mark size=2.5pt, thick]
            coordinates {(0.02,0.008348) (0.05,0.008348) (0.1,0.008348) (0.2,0.008348) (0.5,0.008348) (1.0,0.008348)};
        \addlegendentry{LoRA}

\end{groupplot}
\end{tikzpicture}
\caption{{\small{\textbf{Synthetic validation.}  (a)~Under dense signal, random and informed placement perform similarly---any parameter subset captures the distributed gradient.  (b)~Under concentrated signal, informed placement at 2\% of $B$ nearly matches Full LoRA, while random placement barely improves even at 50\%.  
}}}
\label{fig:synthetic}
\end{figure}
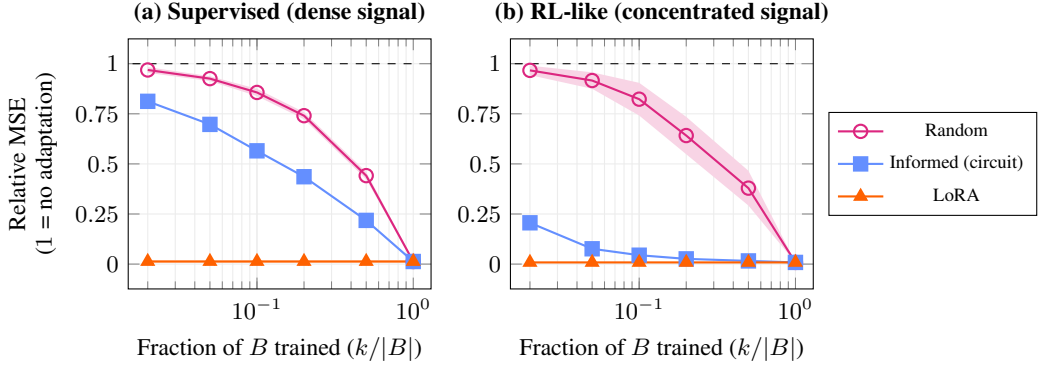

\Cref{fig:synthetic} confirms the prediction. Under dense signal (a), random and informed placement perform similarly across budgets, as gradients are spread across parameters. Under concentrated signal (b), informed placement at 2\% of $\mathbf{B}$ nearly matches LoRA, while random placement fails, with a large gap persisting up to $\sim$50\%. We next test this mechanism in real language models (\Cref{sec:experiments}).

\section{Experiments}
\label{sec:experiments}

We first show that circuit discovery is stable and practical (\Cref{sec:exp_stability}).  Sparse circuits are evaluated under supervised fine-tuning (\Cref{sec:exp_sft}) and reinforcement learning (\Cref{sec:exp_grpo}), its  effect is most pronounced.

We evaluate on base models from 1.5B--7B parameters (Qwen2.5-\{1.5B, 3B, 7B\}~\citep{qwen2.5} and instruct models at 3B and 8B (Llama-3.2-3B-Instruct, Llama-3.1-8B-Instruct)~\citep{grattafiori2024llama} as well as Qwen2.5-7b-Instruct~\citep{qwen2.5} and Qwen2.5-VL-7B-Instruct~\citep{qwen2.5-VL}.  LoRA is applied to all linear layers.  We compare two gradient-based selection methods, $\hat{S}$ (gradient magnitude) and $\hat{F}$ (diagonal Fisher) against random placement, under both SFT and GRPO.  The instruct models test whether placement matters when the base model already provides dense reward signal.  Full hyperparameters are in \Cref{app:hyperparams}.

\subsection{Circuit Discovery Validation}
\label{sec:exp_stability}

Before evaluating downstream performance, we verify that the discovery procedure produces stable circuits.  We report results on Qwen2.5-1.5B and -3B / GSM8K with 10K-element circuits.  For reference, two random 10K-element subsets of the ${\sim}$20M adapter parameters overlap by $< 0.1\%$ in expectation, so any substantial overlap reflects structure in the scores (details in \Cref{app:mc_convergence,app:a_perturbation}).

\paragraph{Stability and cost.} Circuit discovery is stable to sampling. Using as few as $N{=}10$ examples recovers $>90\%$ of the top-10K elements of the $N{=}100$ reference circuit, reaching $\approx 98\%$ at $N{=}50$ across scales. Independent data subsets yield $95\%$ mutual overlap, and downstream SFT accuracy is unchanged ($60.0 \pm 1.5\%$), indicating that small samples recover the population circuit. The exact elements of the circuit are sensitive to $\mathbf{A}$ initialization, but the selected modules and layers remain constant across random seeds (\S\ref{app:seed_stability}). Discovery is inexpensive, requiring $N$ forward-backward passes with no optimizer state and completing in $<10$s on a single GPU ($<0.5\%$ of SFT cost). The resulting circuit is highly compact (under 40\,KB for 10K elements, $\sim\!4000\times$ smaller than LoRA) and reusable across runs (\Cref{tab:discovery_cost}).

%% ============================================================
\subsection{Supervised Fine-Tuning}
\label{sec:exp_sft}
We evaluate circuit placement under SFT, where gradients are concentrated and stable (\Cref{sec:signal_retention}).

We fine-tune on Alpaca~\citep{alpaca} and Bespoke-Stratos~\citep{bespoke_stratos}, evaluating on instruction following benchmarks (7-task average; \Cref{app:sft_7bench}) and GPQADiamond~\citep{rein2024gpqa}. We also test vision-language setting by fine-tuning Qwen2.5-VL-7B-Instruct~\citep{qwen2.5-VL} on MathV360K~\citep{shihu2024mathllava} and evaluating on MathVista~\citep{lu2024mathvista}.

Under SFT, differences between methods are small across most settings, indicating that arbitrary placements already capture useful gradient signal. On MathVista, all methods converge to similar accuracy, with $\hat{S}$ at k=10K (67.8\%) marginally edging Full LoRA (66.9\%, \Cref{tab:sft_vlm}). The exception is GPQA Diamond, where informed placement pulls +10pp ahead of random at the same budget, suggesting that circuit placement may become important at extreme parameter budgets on hard tasks.

\begin{table}[t]
\centering
\caption{{\small{\textbf{7-benchmark average accuracy over 5 seeds (\%) after SFT on Alpaca.}  Informed placement provides a consistent edge over random, though the gap is modest under SFT's stable gradients.  Per-benchmark results in \Cref{app:sft_7bench}; budget ablations in \Cref{app:budget_selection}.}}}
\label{tab:sft_7bench}
\vspace{4pt}
\small
\setlength{\tabcolsep}{4.5pt}
\renewcommand{\arraystretch}{0.87}
\begin{tabular}{l r r r r r r@{}}
\toprule
\textbf{Model} & \textbf{Pretrained} & \textbf{Random} & $\hat{S}$ & $\hat{F}$ & \textbf{Full LoRA} & \textbf{$k$} \\
\midrule
\multicolumn{7}{l}{\emph{SFT on Alpaca (Instruction Following); Accuracy over 7-benchmarks.}} \\
\addlinespace[2pt]
\rowcolor{baserow}
Qwen2.5-1.5B  & 53.0 & 61.6 & 61.9 & 62.1 & \textbf{62.4} & 50K \\
\rowcolor{baserow}
Qwen2.5-3B    & 65.6 & 65.8 & 66.2 & 66.0 & \textbf{66.7} & 50K \\
\rowcolor{baserow}
Qwen2.5-7B    & 70.9  & 71.1 & 71.4 & \textbf{71.5} & \textbf{71.5} & 100K \\
Llama-3.2-3B-Inst  & 62.3  & 62.9 & 63.8 & 63.7 & \textbf{63.9} & 50K \\
Llama-3.1-8B-Inst  & 68.7  & 69.1 & 69.6 & 69.7 & \textbf{69.9} & 100K \\
\midrule
\multicolumn{7}{l}{\emph{SFT on Bespoke-Stratos (Reasoning); Accuracy on GPQADiamond.}} \\
\addlinespace[2pt]
\rowcolor{baserow}
Qwen2.5-7B-Inst & 31.6 & {41.1} & {43.3} & 31.1 & \textbf{50.7} & 5K \\
\bottomrule
\end{tabular}
\end{table}

%% ============================================================
\subsection{Reinforcement Learning (GRPO)}
\label{sec:exp_grpo}

We compare circuit methods ($\hat{S}, \hat{F}$) to random placement and full LoRA under GRPO (\Cref{tab:grpo}). In contrast to SFT, placement has a drastic effect. For all Qwen models, random placement stays at base level and is insensitive to budget or scaling (\Cref{tab:random_failure} in \Cref{app:random_grpo}). In contrast, both $\hat{S}$ and $\hat{F}$ at the same budget (k=10K--
50K, under 0.1\% of LoRA parameters) match or exceed full LoRA across model sizes (additional budget scaling in \Cref{tab:grpo_budget_all} in \Cref{app:budget_selection}). Similar trends hold for instruct models, though gains are smaller.

\begin{table}[t]
\centering
\caption{\small{%
  \textbf{GRPO accuracy (\%) on GSM8K and MATH-500.}
Random placement fails for base models (Random $\approx$ Pretrained) while circuit placement recovers full LoRA accuracy.
  $^\star$$k{=}10\text{K}$; otherwise $k{=}50\text{K}$. Additional ablations in \Cref{app:budget_selection,app:random_grpo}.
}}
\label{tab:grpo}
\vspace{3pt}
\setlength{\tabcolsep}{4.2pt}
\renewcommand{\arraystretch}{0.87}
\small
\begin{tabular}{@{}l r r r r r@{}}
\toprule
& & & \multicolumn{2}{c}{\textbf{Circuit ($k$ params)}} & \\
\cmidrule(lr){4-5}
\textbf{Model} & \textbf{Pretrained} & \textbf{Random} & $\hat{S}$ & $\hat{F}$ & \textbf{Full LoRA} \\
\midrule
\multicolumn{6}{@{}l@{}}{\footnotesize\textit{GSM8K}} \\[1.5pt]
\rowcolor{baserow}
\quad Qwen2.5-1.5B$^\star$ & 9.5  & 9.8  & \textbf{64.2} & 63.2 & 62.8 \\
\rowcolor{baserow}
\quad Qwen2.5-3B$^\star$   & 62.1 & 62.3 & 70.6 & \textbf{73.6} & 71.9 \\
\rowcolor{baserow}
\quad Qwen2.5-7B           & 79.3 & 79.6 & 85.2 & 86.3 & \textbf{87.6} \\
\quad Llama-3.2-3B-Inst & 70.4 & 67.9 & \textbf{73.9} & 72.6 & 74.7 \\
\quad Llama-3.1-8B-Inst    & 82.9 & 83.4 & 86.2 & 85.9 & \textbf{86.4} \\
\midrule
\multicolumn{6}{@{}l@{}}{\footnotesize\textit{MATH-500}} \\[1.5pt]
\rowcolor{baserow}
\quad Qwen2.5-1.5B$^\star$ & 17.4 & 18.8 & \textbf{43.6} & 42.2 & 41.4 \\
\rowcolor{baserow}
\quad Qwen2.5-3B           & 50.6 & 51.0 & 60.4 & \textbf{63.2} & 62.6 \\
\rowcolor{baserow}
\quad Qwen2.5-7B           & 53.6 & 55.2 & 74.8 & \textbf{76.0} & 74.6 \\
\quad Llama-3.2-3B-Inst    & 43.6 & 50.2 & 50.8 & \textbf{51.0} & 50.8 \\
\quad Llama-3.1-8B-Inst    & 50.4 & 53.2 & 54.6 & 55.8 & \textbf{56.6} \\
\bottomrule
\end{tabular}
\end{table}

GRPO gradients are high-rank and near-orthogonal across steps (\Cref{tab:gradient_properties,tab:gradient_full}), so most coordinates accumulate little net displacement, random placement samples these incoherent directions and fails. $\hat{S}$ selects coordinates with consistent gradient sign, recovering the small coherent component; circuit elements move
$1.4-2.1\times$ more than random under GRPO (\Cref{tab:alignment}). $\hat{F}$ weights by expected gradient magnitude rather than sign alone; both methods concentrate on the same functional regions (V, O, Down), which may account for their similar performance despite distinct selection criteria.

%% ============================================================
\begin{figure}[t!]
\centering
% Auto-generated by pgf_ood_passk.py — do not edit by hand.
% Usage: \input{figures/ood_passk.tex}
%   Requires: tinylora_preamble.tex, pgfplots, groupplots, patterns, calc
\begin{tikzpicture}

\begin{groupplot}[
    tinylora,
    group style={
        group size=2 by 1,
        horizontal sep=0.4cm,
    },
    width=0.32\textwidth,
    height=0.40\textwidth,
]

    \nextgroupplot[
        title={(a) Trained on MATH},
        xlabel={$k$ (number of attempts)},
        ylabel={pass@$k$ (\%)},
        xmode=log, log basis x=2,
        xtick={1, 4, 8, 16, 32, 64},
        xticklabels={1, 4, 8, 16, 32, 64},
        ymin=-3, ymax=44,
        legend to name=passlegend,
        legend columns=3,
    ]
        \addplot[colorBase, mark=o, mark size=2.5pt, thick, solid,
            every mark/.append style={solid},
            mark options={fill=white, draw=colorBase}, forget plot]
            coordinates {(1,0.0) (4,3.3) (8,10.0) (16,10.0) (32,20.0) (64,23.3)};
        \addplot[colorRandom, mark=square*, mark size=2.5pt, thick, solid,
            every mark/.append style={solid},
            mark options={fill=white, draw=colorRandom}, forget plot]
            coordinates {(1,0.0) (4,3.3) (8,6.7) (16,13.3) (32,30.0) (64,36.7)};
        \addplot[colorSignal, mark=triangle*, mark size=2.5pt, thick, solid,
            every mark/.append style={solid},
            mark options={fill=colorSignal}, forget plot]
            coordinates {(1,0.0) (4,6.7) (8,6.7) (16,20.0) (32,20.0) (64,30.0)};
        \addplot[colorFisher, mark=diamond*, mark size=2.5pt, thick, solid,
            every mark/.append style={solid},
            mark options={fill=colorFisher}, forget plot]
            coordinates {(1,3.3) (4,10.0) (8,16.7) (16,23.3) (32,30.0) (64,36.7)};
        \addplot[colorFullLoRA, mark=star, mark size=2.5pt, thick, solid,
            every mark/.append style={solid},
            mark options={fill=white, draw=colorFullLoRA}, forget plot]
            coordinates {(1,3.3) (4,10.0) (8,10.0) (16,20.0) (32,26.7) (64,36.7)};
        \addlegendimage{colorFisher, mark=diamond*, mark size=2.5pt, thick, solid,
            every mark/.append style={solid}, mark options={fill=colorFisher}}
        \addlegendentry{$\hat{F}$ 50K}
        \addlegendimage{colorSignal, mark=triangle*, mark size=2.5pt, thick, solid,
            every mark/.append style={solid}, mark options={fill=colorSignal}}
        \addlegendentry{$\hat{S}$ 50K}
        \addlegendimage{colorRandom, mark=square*, mark size=2.5pt, thick, solid,
            every mark/.append style={solid}, mark options={fill=white, draw=colorRandom}}
        \addlegendentry{Random 50K}
        \addlegendimage{colorFullLoRA, mark=star, mark size=2.5pt, thick, solid,
            every mark/.append style={solid}, mark options={fill=white, draw=colorFullLoRA}}
        \addlegendentry{Full LoRA}
        \addlegendimage{colorBase, mark=o, mark size=2.5pt, thick, solid,
            every mark/.append style={solid}, mark options={fill=white, draw=colorBase}}
        \addlegendentry{Base}

    \nextgroupplot[
        title={(b) Trained on GSM8K},
        xlabel={$k$ (number of attempts)},
        ylabel={},
        yticklabels={},
        xmode=log, log basis x=2,
        xtick={1, 4, 8, 16, 32, 64},
        xticklabels={1, 4, 8, 16, 32, 64},
        ymin=-3, ymax=44,
    ]
        \addplot[colorBase, mark=o, mark size=2.5pt, thick, solid,
            every mark/.append style={solid},
            mark options={fill=white, draw=colorBase}, forget plot]
            coordinates {(1,0.0) (4,3.3) (8,10.0) (16,10.0) (32,20.0) (64,23.3)};
        \addplot[colorRandom, mark=square*, mark size=2.5pt, thick, solid,
            every mark/.append style={solid},
            mark options={fill=white, draw=colorRandom}, forget plot]
            coordinates {(1,0.0) (4,0.0) (8,6.7) (16,10.0) (32,20.0) (64,30.0)};
        \addplot[colorSignal, mark=triangle*, mark size=2.5pt, thick, solid,
            every mark/.append style={solid},
            mark options={fill=colorSignal}, forget plot]
            coordinates {(1,6.7) (4,16.7) (8,20.0) (16,20.0) (32,23.3) (64,30.0)};
        \addplot[colorFisher, mark=diamond*, mark size=2.5pt, thick, solid,
            every mark/.append style={solid},
            mark options={fill=colorFisher}, forget plot]
            coordinates {(1,6.7) (4,16.7) (8,20.0) (16,20.0) (32,23.3) (64,26.7)};
        \addplot[colorFullLoRA, mark=star, mark size=2.5pt, thick, solid,
            every mark/.append style={solid},
            mark options={fill=white, draw=colorFullLoRA}, forget plot]
            coordinates {(1,3.3) (4,3.3) (8,3.3) (16,6.7) (32,10.0) (64,16.7)};

\end{groupplot}

% Pass@k legend (2 rows) centered under panels (a) and (b)
\node[below, yshift=-10mm] at
    ($(group c1r1.south)!0.5!(group c2r1.south)$)
    {\begingroup\ifdefined\hypersetup\hypersetup{hidelinks}\fi
     \pgfplotslegendfromname{passlegend}\endgroup};

\begin{axis}[
    tinylora,
    name=panelc,
    at={($(group c2r1.north east)+(1.4cm,0)$)},
    anchor=north west,
    width=0.40\textwidth,
    height=0.40\textwidth,
    title={(c) Cross-task transfer (greedy)},
    ylabel={Accuracy (\%)},
    xtick={0, 1, 2, 3},
    xticklabels={Rand., $\hat{S}$, $\hat{F}$, {Full\\LoRA}},
    x tick label style={align=center},
    xmin=-0.6, xmax=3.6,
    ymin=0, ymax=92,
    enlarge y limits={upper, value=0.02},
    ytick={0, 20, 40, 60, 80},
    xmajorgrids=false,
    legend to name=barlegend,
    legend columns=2,
]
    \addplot[colorBase, dashed, very thick, forget plot]
        coordinates {(-0.6,62.1) (3.6,62.1)};
    \addplot[colorBase, densely dotted, thick, forget plot]
        coordinates {(-0.6,49.0) (3.6,49.0)};
    \addplot[ybar, bar width=10pt, bar shift=-5pt,
        fill=colorRandom, draw=none, forget plot]
        coordinates {(0,63.6)};
    \addplot[ybar, bar width=10pt, bar shift=-5pt,
        fill=colorSignal, draw=none, forget plot]
        coordinates {(1,71.7)};
    \addplot[ybar, bar width=10pt, bar shift=-5pt,
        fill=colorFisher, draw=none, forget plot]
        coordinates {(2,74.4)};
    \addplot[ybar, bar width=10pt, bar shift=-5pt,
        fill=colorFullLoRA, draw=none, forget plot]
        coordinates {(3,73.8)};
    \addplot[ybar, bar width=10pt, bar shift=5pt,
        fill=colorRandom, fill opacity=0.55, draw=none,
        postaction={pattern=north east lines, pattern color=black!60},
        forget plot]
        coordinates {(0,46.2)};
    \addplot[ybar, bar width=10pt, bar shift=5pt,
        fill=colorSignal, fill opacity=0.55, draw=none,
        postaction={pattern=north east lines, pattern color=black!60},
        forget plot]
        coordinates {(1,53.8)};
    \addplot[ybar, bar width=10pt, bar shift=5pt,
        fill=colorFisher, fill opacity=0.55, draw=none,
        postaction={pattern=north east lines, pattern color=black!60},
        forget plot]
        coordinates {(2,52.8)};
    \addplot[ybar, bar width=10pt, bar shift=5pt,
        fill=colorFullLoRA, fill opacity=0.55, draw=none,
        postaction={pattern=north east lines, pattern color=black!60},
        forget plot]
        coordinates {(3,34.2)};
    \addlegendimage{area legend, fill=gray!70}
    \addlegendentry{MATH$\rightarrow$GSM8K}
    \addlegendimage{area legend, fill=gray!70, fill opacity=0.55,
        postaction={pattern=north east lines, pattern color=black!60}}
    \addlegendentry{GSM8K$\rightarrow$MATH}
    \addlegendimage{colorBase, dashed, very thick, no markers}
    \addlegendentry{Base (GSM8K)}
    \addlegendimage{colorBase, densely dotted, thick, no markers}
    \addlegendentry{Base (MATH)}
\end{axis}

% Bar chart legend (2x2) centered under panel (c)
\node[below, yshift=-10mm] at (panelc.south)
    {\begingroup\ifdefined\hypersetup\hypersetup{hidelinks}\fi
     \pgfplotslegendfromname{barlegend}\endgroup};
\end{tikzpicture}
\caption{\textbf{OOD generalization} (Qwen2.5-3B, GRPO).  (a,\,b) AIME 2024 pass@$k$ for models trained on MATH and GSM8K respectively.  (c) Cross-task greedy transfer: solid bars = MATH$\rightarrow$GSM8K, hatched = GSM8K$\rightarrow$MATH; dashed/dotted lines show base-model accuracy.  $\hat{F}$ at $<$1\% of adapter parameters matches or exceeds Full LoRA across all settings.}
\label{fig:ood}
\end{figure}
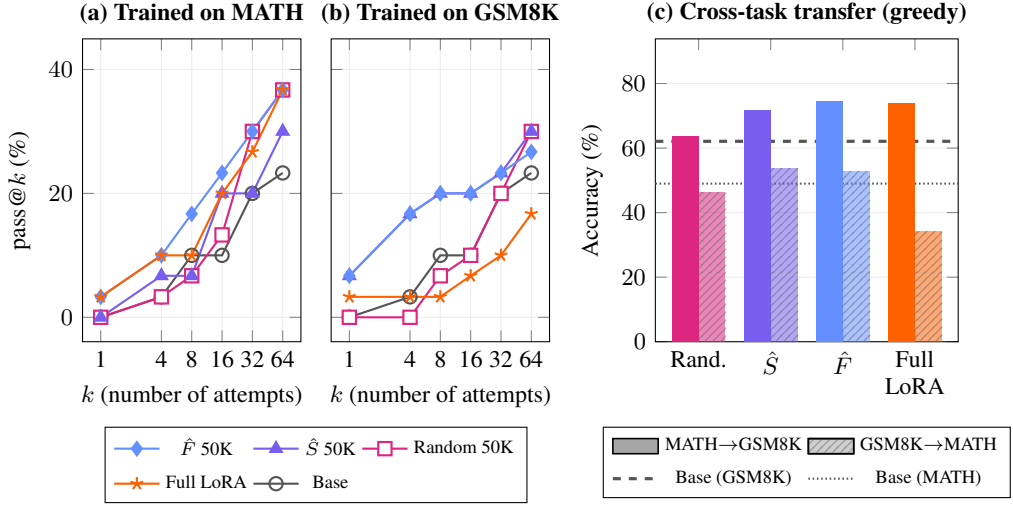

\subsection{Out-of-Distribution Generalization}
\label{sec:ood}

Does informed placement generalize beyond the training distribution?  We evaluate our Qwen2.5-3B models RL trained on MATH on held-out benchmarks: GSM8K (cross-task transfer), AMC 2023 (40 competition problems)~\citep{amc2023dataset}, and AIME 2024 (30 problems)~\citep{aime_2024} (additional details in \Cref{app:ood}).

Circuits discovered on different math datasets share $\sim 30$\% of their elements (12 $\times$ above chance), suggesting a stable, task-general subnetwork rather than dataset-specific artifacts (\Cref{app:cross_dataset_overlap}). This structural overlap predicts that informed placement should transfer out-of-distribution (\Cref{fig:ood}). $\hat{F}$
at 50K parameters matches Full LoRA on cross-task transfer to GSM8K (74.4\% vs.\ 73.8\%) and exceeds it on AMC (35.0\% vs.\ 32.5\%), while random placement barely improves over the base model. The advantage is sharpest where overfitting risk is highest: GSM8K-trained $\hat{F}$ circuits outperform Full LoRA on AIME (\Cref{fig:ood_gsm_passk}), and reaches 100\% pass@64 on AMC versus 87.5\% for random. Concentrating updates on a small set of functionally stable parameters appears to act as implicit regularization, preserving generality that dense adapters lose.

\section{Analysis}
\label{sec:analysis_main}
Here, we analyze the location as well as causal effects and behavior of the circuits during training. 
\subsection{Anatomy of Selected Circuits}\label{sec:anatomy}
\Cref{tab:circuit_composition:main} reports the placement density of $\hat{S}$, quantifying the module-level structure of $\hat{S}$ circuits (top 10K entries) across all four configurations.  V is overrepresented in every setting (21--39\% of circuit from ${\sim}$1\% of parameters).  At 7B scale, O absorbs most of the remaining circuit budget (38--60\%), while at 3B, the budget is more distributed across O, Down, and MLP modules.  Gate and Up together hold ${\sim}$76\% of all $\mathbf{B}$ parameters but capture only 16--44\% of circuit entries.  $\hat{F}$ circuits show the same ordering with budget shifting slightly from Gate/Up towards V and O. Both $\hat{S}$ and $\hat{F}$ circuits concentrate in layers 0–4 (70--98\% of budget), with concentration increasing at larger scale. Additional details are presented in \Cref{app:circuit_anatomy,app:perlayer_concentration}.
\begin{table}
\centering
\caption{\textbf{$\hat{S}$-circuit composition} (top 10K entries, \% of circuit).  ``Param'' shows each module type's share of all $\mathbf{B}$ parameters.  V is overrepresented in all configurations; O dominates at 7B.  Gate and Up are consistently underrepresented despite holding the majority of parameters.}
\label{tab:circuit_composition:main}
\vspace{4pt}
\small
\setlength{\tabcolsep}{4.5pt}
\begin{tabular}{@{}l c cccc@{}}
\toprule
& & \multicolumn{4}{c}{\textbf{\% of circuit (top 10K)}} \\
\cmidrule(lr){3-6}
\textbf{Module} & \textbf{Param \%} & \textbf{3B MATH} & \textbf{3B GSM8K} & \textbf{7B MATH} & \textbf{7B GSM8K} \\
\midrule
Q    & \phantom{0}7.1 & \phantom{0}2.9 & \phantom{0}2.2 & \phantom{0}0.5 & \phantom{0}0.6 \\
K    & \phantom{0}0.9 & \phantom{0}6.1 & \phantom{0}8.5 & \phantom{0}1.9 & \phantom{0}2.4 \\
\rowcolor{black!4}
V    & \phantom{0}0.9 & 23.6 & 29.2 & 21.3 & 39.2 \\
\rowcolor{black!4}
O    & \phantom{0}7.1 & 11.1 & \phantom{0}2.1 & 59.9 & 38.2 \\
Gate & 38.4 & 17.3 & 27.8 & \phantom{0}7.7 & \phantom{0}8.2 \\
Up   & 38.4 & 19.8 & 16.4 & \phantom{0}8.5 & 11.1 \\
\rowcolor{black!4}
Down & \phantom{0}7.1 & 19.1 & 13.9 & \phantom{0}0.1 & \phantom{0}0.2 \\
\bottomrule
\end{tabular}
\end{table}

This concentration is not arbitrary: gradient-consistency and SVD-alignment analyses (\Cref{app:module_dissociation}) reveal that selected modules fall into two functional classes: \textbf{attention readers} (V, K) that align with the pretrained weight's dominant singular vectors, and \textbf{residual writers} (O, Down) that produce spectrally concentrated updates, both at the interface of the residual stream.

\noindent\textbf{Knockout validation.}  Our results show that circuits learn effectively, but do the scores identify parameters that are already functional in a trained adapter?  We answer with a training-free knockout sweep: starting from a fully trained LoRA, we zero out the top fraction of $\mathbf{B}$ entries by $\hat{F}$ score and measure accuracy after merging the residual adapter into the base model.  If $\hat{F}$ identifies critical parameters, circuit-ordered knockout should degrade accuracy faster than removing entries at random.

\begin{figure}[t]
\centering
\input{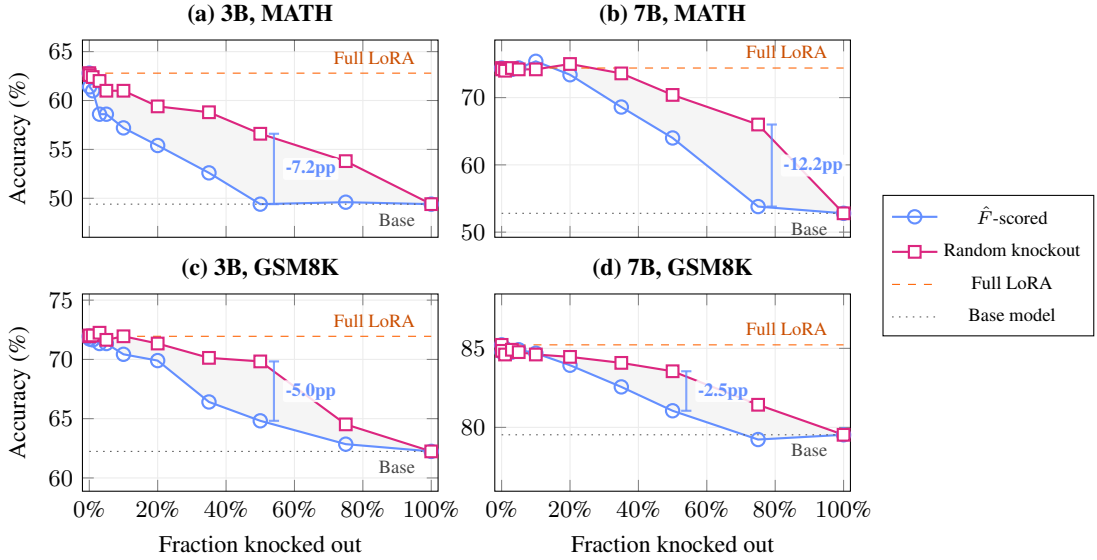}
\vspace{-2mm}
\caption{\small{\textbf{$\hat{F}$ knockout sweep} (3B/7B $\times$ MATH/GSM8K).  Accuracy vs.\ fraction of $\mathbf{B}$ entries zeroed by $\hat{F}$ score (circuit, red) or at random (blue).  Dashed: full LoRA; dotted: base model.  Circuit-ordered knockout degrades faster in all configurations, with the gap scaling with LoRA effect size.}}
\label{fig:knockout}
\end{figure}

Zeroing $\hat{F}$-ranked entries degrades accuracy faster than random removal in all four configurations (\Cref{fig:knockout}), causally confirming that scored parameters mediate the adapter's effect. The gap scales predictably with the adapter's total contribution: where LoRA spans 21.6pp (7B MATH), peak degradation reaches -12.2pp; where it spans 9.8pp (3B GSM8K), -5.0pp.  $\hat{S}$-scored knockout replicates similar qualitative pattern (\Cref{app:knockout}).

\begin{figure}[t]
\centering
\input{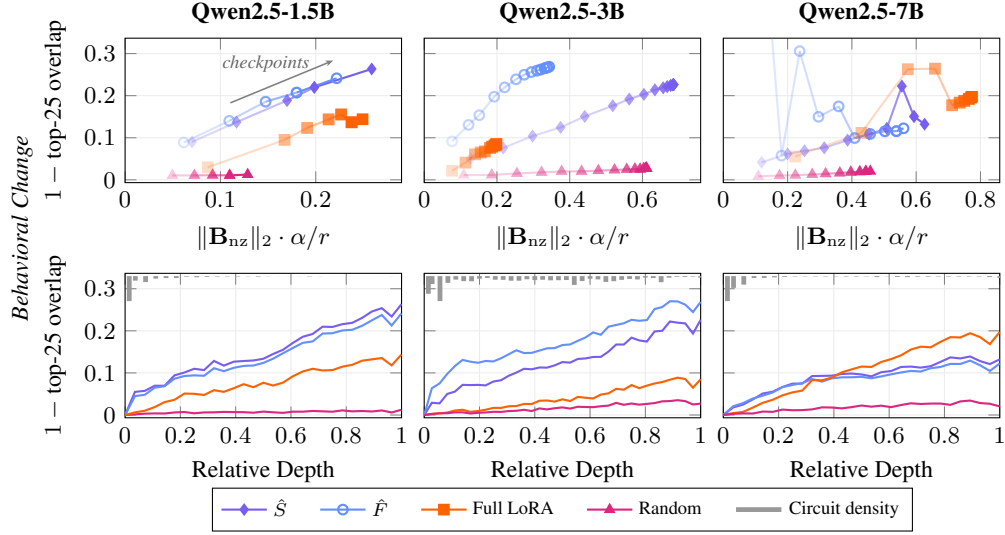}
\caption{\textbf{Divergence vs.\ effective update norm} (Qwen2.5-1.5B/3B/7B, MATH-500, GRPO). For both rows, the $y$-axis is $1 - \textit{top-25 token overlap}$, measuring behavioral change via logit lens relative to the base model. Top: each trajectory traces training checkpoints, opacity increases with step. $x$-axis: $|\mathbf{B}_{\mathrm{nz}}|_2 \cdot \alpha/r$, the scaled norm of non-zero $\mathbf{B}$ weights. Bottom: per-layer divergence at the final checkpoint. $x$-axis: relative layer depth. Gray bars report the relative circuit density.  
}
\label{fig:analysis}
\end{figure}
\subsection{Weight Norm and Behavior Change Dynamics}
\label{sec:dynamics}
In \Cref{fig:analysis}, we examine the dynamics between how selected parameters change and how much model predictions change, across training checkpoints (top) and across model layers (bot).

We measure the dynamics of Qwen2.5-1.5/3/7B (GRPO on MATH as compared to the respective base models) over 100 test examples from MATH. We report $1 -$ \textit{top-25 overlap}, where overlap is the Jaccard overlap of the top-25 predicted tokens between the fine-tuned and base model, averaged over positions and examples including only the prompts and problem text. (Results using Llama models, or KL instead of Jaccard, are similar (\Cref{app:dynamics}).) The x-axis reports the effective norm of the LoRA trainable weights, $\|\mathbf{B}_{\mathrm{nz}}\|_2 \cdot \alpha/r$. Relatedly, since $\mathbf{B} = 0$ at initialization, the non-zero $\mathbf{B}_{\mathrm{nz}}$ weights are the total displacement. In \Cref{app:update_alignment}, we show that GRPO moves the weights at higher rates than SFT. 

\Cref{fig:analysis} (top) shows the projection onto the vocabulary across training checkpoints (as indicated by opacity). Random circuits serve as a clean baseline: despite similar levels of effective weight change, behavioral change remains near zero, confirming that arbitrary parameter subsets fail to route updates into the output distribution. Compared to both LoRA (and random circuits), selected circuits achieve higher behavioral change per parameter than full LoRA, indicating that $\hat{S}$ and $\hat{F}$ identify parameters with disproportionate leverage on the output token distribution.

\Cref{fig:analysis} (bottom) shows the behavioral change of vocabulary projections across model layers (x-axis) of last checkpoint models using logit lens~\citep{belrose2023eliciting}, which maps hidden states to vocabulary logits. Despite most selected parameters being in the lower layers, shown by  \Cref{tab:gradient_properties} and the gray bars in the bottom row denoting the circuit density, change is highest in later layers. This complements \citet{meng2022mass}'s findings, which show that rank-one edits to early layers were often corrected by downstream computation; training, rather than editing, produces updates that compound through depth.

\section{Conclusion}
\label{sec:conclusion}
We studied the parameter placement problem in LoRA: given a fixed budget of trainable parameters, does it matter which are trained?  The answer depends on the training regime.  Under SFT, gradients are concentrated and directionally stable, so arbitrary subsets accumulate coherent updates and placement has limited effect.  Under GRPO on base models, gradients are distributed across many dimensions and nearly orthogonal across steps so most parameter subsets random-walk rather than learn. 
A lightweight procedure--50 forward-backward passes at initialization--identifies a sparse subset of $\mathbf{B}$ elements with consistently signed gradients.  At less than 0.05\% of adapter parameters, these circuits recover full LoRA performance under GRPO across models from 1.5B to 8B.

\section*{Author Contributions}
AS ran instruction tuning SFT and the RL experiments and did the anatomy and knockout analysis of the circuits. CL ran the reasoning SFT experiments and investigated the dynamics of the circuits during training. Both authors contributed to the writing of the manuscript. 

\bibliographystyle{plainnat}
\bibliography{references}

@article{meng2022mass,
  title={Mass-editing memory in a transformer},
  author={Meng, Kevin and Sharma, Arnab Sen and Andonian, Alex and Belinkov, Yonatan and Bau, David},
  journal={arXiv preprint arXiv:2210.07229},
  year={2022}
}

@article{belrose2023eliciting,
  title={Eliciting latent predictions from transformers with the tuned lens},
  author={Belrose, Nora and Ostrovsky, Igor and McKinney, Lev and Furman, Zach and Smith, Logan and Halawi, Danny and Biderman, Stella and Steinhardt, Jacob},
  journal={arXiv preprint arXiv:2303.08112},
  year={2023}
}

@article{schulman2025lora,
  author = {John Schulman and Thinking Machines Lab},
  title = {LoRA Without Regret},
  journal = {Thinking Machines Lab: Connectionism},
  year = {2025},
  note = {https://thinkingmachines.ai/blog/lora/},
  doi = {10.64434/tml.20250929},
}

@article{DBLP:journals/corr/abs-1810-02340,
  author       = {Namhoon Lee and
                  Thalaiyasingam Ajanthan and
                  Philip H. S. Torr},
  title        = {{SNIP:} Single-shot Network Pruning based on Connection Sensitivity},
  journal      = {CoRR},
  volume       = {abs/1810.02340},
  year         = {2018},
  url          = {http://arxiv.org/abs/1810.02340},
  eprinttype   = {arXiv},
  eprint       = {1810.02340},
  bibsource    = {dblp computer science bibliography, https://dblp.org}
}

@article{DBLP:journals/corr/abs-2002-07376,
  author       = {Chaoqi Wang and
                  Guodong Zhang and
                  Roger Baker Grosse},
  title        = {Picking Winning Tickets Before Training by Preserving Gradient Flow},
  journal      = {CoRR},
  volume       = {abs/2002.07376},
  year         = {2020},
  url          = {https://arxiv.org/abs/2002.07376},
  eprinttype   = {arXiv},
  eprint       = {2002.07376},
  bibsource    = {dblp computer science bibliography, https://dblp.org}
}

@article{DBLP:journals/corr/abs-2006-05467,
  author       = {Hidenori Tanaka and
                  Daniel Kunin and
                  Daniel L. K. Yamins and
                  Surya Ganguli},
  title        = {Pruning neural networks without any data by iteratively conserving
                  synaptic flow},
  journal      = {CoRR},
  volume       = {abs/2006.05467},
  year         = {2020},
  url          = {https://arxiv.org/abs/2006.05467},
  eprinttype   = {arXiv},
  eprint       = {2006.05467},
  bibsource    = {dblp computer science bibliography, https://dblp.org}
}

@article{gurnee2025when,
  author={Gurnee, Wes and Ameisen, Emmanuel and Kauvar, Isaac and Tarng ,Julius and Pearce, Adam and Olah, Chris and Batson, Joshua},
  title={When Models Manipulate Manifolds: The Geometry of a Counting Task},
  journal={Transformer Circuits Thread},
  year={2025},
  url={https://transformer-circuits.pub/2025/linebreaks/index.html}
}

@article{cammarata2021curve,
  author = {Cammarata, Nick and Goh, Gabriel and Carter, Shan and Voss, Chelsea and Schubert, Ludwig and Olah, Chris},
  title = {Curve Circuits},
  journal = {Distill},
  year = {2021},
  note = {https://distill.pub/2020/circuits/curve-circuits},
  doi = {10.23915/distill.00024.006}
}

@article{wang2022interpretability,
  title={Interpretability in the wild: a circuit for indirect object identification in gpt-2 small},
  author={Wang, Kevin and Variengien, Alexandre and Conmy, Arthur and Shlegeris, Buck and Steinhardt, Jacob},
  journal={arXiv preprint arXiv:2211.00593},
  year={2022}
}

@misc{ge2024automaticallyidentifyinglocalglobal,
      title={Automatically Identifying Local and Global Circuits with Linear Computation Graphs}, 
      author={Xuyang Ge and Fukang Zhu and Wentao Shu and Junxuan Wang and Zhengfu He and Xipeng Qiu},
      year={2024},
      eprint={2405.13868},
      archivePrefix={arXiv},
      primaryClass={cs.LG},
      url={https://arxiv.org/abs/2405.13868}, 
}

@misc{haklay2025positionawareautomaticcircuitdiscovery,
      title={Position-aware Automatic Circuit Discovery}, 
      author={Tal Haklay and Hadas Orgad and David Bau and Aaron Mueller and Yonatan Belinkov},
      year={2025},
      eprint={2502.04577},
      archivePrefix={arXiv},
      primaryClass={cs.LG},
      url={https://arxiv.org/abs/2502.04577}, 
}

@misc{kharlapenko2025scalingsparsefeaturecircuit,
      title={Scaling sparse feature circuit finding for in-context learning}, 
      author={Dmitrii Kharlapenko and Stepan Shabalin and Fazl Barez and Arthur Conmy and Neel Nanda},
      year={2025},
      eprint={2504.13756},
      archivePrefix={arXiv},
      primaryClass={cs.LG},
      url={https://arxiv.org/abs/2504.13756}, 
}

@article{nanda2023progress,
  title={Progress measures for grokking via mechanistic interpretability},
  author={Nanda, Neel and Chan, Lawrence and Lieberum, Tom and Smith, Jess and Steinhardt, Jacob},
  journal={arXiv preprint arXiv:2301.05217},
  year={2023}
}

@misc{kramr2024atpefficientscalablemethod,
      title={AtP*: An efficient and scalable method for localizing LLM behaviour to components}, 
      author={János Kramár and Tom Lieberum and Rohin Shah and Neel Nanda},
      year={2024},
      eprint={2403.00745},
      archivePrefix={arXiv},
      primaryClass={cs.LG},
      url={https://arxiv.org/abs/2403.00745}, 
}

@article{hu2022lora,
  title={Lora: Low-rank adaptation of large language models.},
  author={Hu, Edward J and Shen, Yelong and Wallis, Phillip and Allen-Zhu, Zeyuan and Li, Yuanzhi and Wang, Shean and Wang, Liang and Chen, Weizhu and others},
  journal={Iclr},
  volume={1},
  number={2},
  pages={3},
  year={2022}
}

@inproceedings{kopiczko2024vera,
  title={VeRA: Vector-based Random Matrix Adaptation},
  author={Kopiczko, Dawid Jan and Blankevoort, Tijmen and Asano, Yuki M},
  booktitle={The Twelfth International Conference on Learning Representations},
  year={2024},
}

@inproceedings{gao2024fourierft,
  title={Parameter-Efficient Fine-Tuning with Discrete Fourier Transform},
  author={Gao, Ziqi and Wang, Qichao and Chen, Aochuan and Liu, Zijing and Wu, Bingzhe and Chen, Liang and Li, Jia},
  booktitle={International Conference on Machine Learning},
  pages={14884--14901},
  year={2024},
  organization={PMLR}
}

@inproceedings{
albert2025randlora,
title={RandLo{RA}: Full rank parameter-efficient fine-tuning of large models},
author={Paul Albert and Frederic Z. Zhang and Hemanth Saratchandran and Cristian Rodriguez-Opazo and Anton van den Hengel and Ehsan Abbasnejad},
booktitle={The Thirteenth International Conference on Learning Representations},
year={2025},
url={https://openreview.net/forum?id=Hn5eoTunHN}
}

@article{kalajdzievski2023rslora,
  title={A rank stabilization scaling factor for fine-tuning with lora},
  author={Kalajdzievski, Damjan},
  journal={arXiv preprint arXiv:2312.03732},
  year={2023}
}

@article{morris2026tinylora,
  title={Learning to Reason in 13 Parameters},
  author={Morris, John X and Mireshghallah, Niloofar and Ibrahim, Mark and Mahloujifar, Saeed},
  journal={arXiv preprint arXiv:2602.04118},
  year={2026}
}

@article{meng2024pissa,
  title={Pissa: Principal singular values and singular vectors adaptation of large language models},
  author={Meng, Fanxu and Wang, Zhaohui and Zhang, Muhan},
  journal={Advances in Neural Information Processing Systems},
  volume={37},
  pages={121038--121072},
  year={2024}
}

@inproceedings{
yang2024corda,
title={Cor{DA}: Context-Oriented Decomposition Adaptation of Large Language Models for Task-Aware Parameter-Efficient Fine-tuning},
author={Yibo Yang and Xiaojie Li and Zhongzhu Zhou and Shuaiwen Leon Song and Jianlong Wu and Liqiang Nie and Bernard Ghanem},
booktitle={The Thirty-eighth Annual Conference on Neural Information Processing Systems},
year={2024},
url={https://openreview.net/forum?id=Gi00NVru6n}
}

@article{hayou2024loraplus,
  title={Lora+: Efficient low rank adaptation of large models},
  author={Hayou, Soufiane and Ghosh, Nikhil and Yu, Bin},
  journal={arXiv preprint arXiv:2402.12354},
  year={2024}
}

@inproceedings{liu2024dora,
  title={Dora: Weight-decomposed low-rank adaptation},
  author={Liu, Shih-Yang and Wang, Chien-Yi and Yin, Hongxu and Molchanov, Pavlo and Wang, Yu-Chiang Frank and Cheng, Kwang-Ting and Chen, Min-Hung},
  booktitle={Forty-first International Conference on Machine Learning},
  year={2024}
}

@inproceedings{
zhang2023adalora,
title={Adaptive Budget Allocation for Parameter-Efficient Fine-Tuning },
author={Qingru Zhang and Minshuo Chen and Alexander Bukharin and Pengcheng He and Yu Cheng and Weizhu Chen and Tuo Zhao},
booktitle={The Eleventh International Conference on Learning Representations },
year={2023},
url={https://openreview.net/forum?id=lq62uWRJjiY}
}

@inproceedings{valipour2023dylora,
    title = "{D}y{L}o{RA}: Parameter-Efficient Tuning of Pre-trained Models using Dynamic Search-Free Low-Rank Adaptation",
    author = "Valipour, Mojtaba  and
      Rezagholizadeh, Mehdi  and
      Kobyzev, Ivan  and
      Ghodsi, Ali",
    editor = "Vlachos, Andreas  and
      Augenstein, Isabelle",
    booktitle = "Proceedings of the 17th Conference of the European Chapter of the Association for Computational Linguistics",
    month = may,
    year = "2023",
    address = "Dubrovnik, Croatia",
    publisher = "Association for Computational Linguistics",
    url = "https://aclanthology.org/2023.eacl-main.239/",
    doi = "10.18653/v1/2023.eacl-main.239",
    pages = "3274--3287",
}

@inproceedings{ding2023sora,
    title = "Sparse Low-rank Adaptation of Pre-trained Language Models",
    author = "Ding, Ning  and
      Lv, Xingtai  and
      Wang, Qiaosen  and
      Chen, Yulin  and
      Zhou, Bowen  and
      Liu, Zhiyuan  and
      Sun, Maosong",
    editor = "Bouamor, Houda  and
      Pino, Juan  and
      Bali, Kalika",
    booktitle = "Proceedings of the 2023 Conference on Empirical Methods in Natural Language Processing",
    month = dec,
    year = "2023",
    address = "Singapore",
    publisher = "Association for Computational Linguistics",
    url = "https://aclanthology.org/2023.emnlp-main.252/",
    doi = "10.18653/v1/2023.emnlp-main.252",
    pages = "4133--4145",
}

@misc{
balazy2024loraxs,
title={Lo{RA}-{XS}: Low-Rank Adaptation with Extremely Small Number of Parameters},
author={Klaudia Ba{\l}azy and Mohammadreza Banaei and Karl Aberer and Jacek Tabor},
year={2024},
url={https://openreview.net/forum?id=l80AgHoRaN}
}

@misc{
zhang2023lorafa,
title={Lo{RA}-{FA}: Memory-efficient Low-rank Adaptation for Large Language Models Fine-tuning},
author={Longteng Zhang and Lin Zhang and Shaohuai Shi and Xiaowen Chu and Bo Li},
year={2024},
url={https://openreview.net/forum?id=RbKThNNFxr}
}

@inproceedings{
zhu2024asymmetry,
title={Asymmetry in Low-Rank Adapters of Foundation Models},
author={Jiacheng Zhu and Kristjan Greenewald and Kimia Nadjahi and Haitz S{\'a}ez de Oc{\'a}riz Borde and Rickard Br{\"u}el Gabrielsson and Leshem Choshen and Marzyeh Ghassemi and Mikhail Yurochkin and Justin Solomon},
booktitle={Forty-first International Conference on Machine Learning},
year={2024},
url={https://openreview.net/forum?id=txRZBD8tBV}
}

@inproceedings{he2022sparseadapter,
    title = "{S}parse{A}dapter: An Easy Approach for Improving the Parameter-Efficiency of Adapters",
    author = "He, Shwai  and
      Ding, Liang  and
      Dong, Daize  and
      Zhang, Jeremy  and
      Tao, Dacheng",
    editor = "Goldberg, Yoav  and
      Kozareva, Zornitsa  and
      Zhang, Yue",
    booktitle = "Findings of the Association for Computational Linguistics: EMNLP 2022",
    month = dec,
    year = "2022",
    address = "Abu Dhabi, United Arab Emirates",
    publisher = "Association for Computational Linguistics",
    url = "https://aclanthology.org/2022.findings-emnlp.160/",
    doi = "10.18653/v1/2022.findings-emnlp.160",
    pages = "2184--2190",
}

@inproceedings{zhou2024loradrop,
  title={Lora-drop: Efficient lora parameter pruning based on output evaluation},
  author={Zhou, Hongyun and Lu, Xiangyu and Xu, Wang and Zhu, Conghui and Zhao, Tiejun and Yang, Muyun},
  booktitle={Proceedings of the 31st International Conference on Computational Linguistics},
  pages={5530--5543},
  year={2025}
}

@inproceedings{zhang2024loraprune,
    title = "{L}o{RAP}rune: Structured Pruning Meets Low-Rank Parameter-Efficient Fine-Tuning",
    author = "Zhang, Mingyang  and
      Chen, Hao  and
      Shen, Chunhua  and
      Yang, Zhen  and
      Ou, Linlin  and
      Yu, Xinyi  and
      Zhuang, Bohan",
    editor = "Ku, Lun-Wei  and
      Martins, Andre  and
      Srikumar, Vivek",
    booktitle = "Findings of the Association for Computational Linguistics: ACL 2024",
    month = aug,
    year = "2024",
    address = "Bangkok, Thailand",
    publisher = "Association for Computational Linguistics",
    url = "https://aclanthology.org/2024.findings-acl.178/",
    doi = "10.18653/v1/2024.findings-acl.178",
    pages = "3013--3026",
}

@article{yang2024s2ft,
  title={S2FT: Efficient, scalable and generalizable LLM fine-tuning by structured sparsity},
  author={Yang, Xinyu and Leng, Jixuan and Guo, Geyang and Zhao, Jiawei and Nakada, Ryumei and Zhang, Linjun and Yao, Huaxiu and Chen, Beidi},
  journal={Advances in Neural Information Processing Systems},
  volume={37},
  pages={59912--59947},
  year={2024}
}

@article{lecun1990obd,
  title={Optimal brain damage},
  author={LeCun, Yann and Denker, John and Solla, Sara},
  journal={Advances in neural information processing systems},
  volume={2},
  year={1989}
}

@inproceedings{hassibi1993obs,
  title={Optimal brain surgeon and general network pruning},
  author={Hassibi, Babak and Stork, David G and Wolff, Gregory J},
  booktitle={IEEE international conference on neural networks},
  pages={293--299},
  year={1993},
  organization={IEEE}
}

@inproceedings{
sun2024wanda,
title={A Simple and Effective Pruning Approach for Large Language Models},
author={Mingjie Sun and Zhuang Liu and Anna Bair and J Zico Kolter},
booktitle={The Twelfth International Conference on Learning Representations},
year={2024},
url={https://openreview.net/forum?id=PxoFut3dWW}
}

@inproceedings{frantar2023sparsegpt,
  title={Sparsegpt: Massive language models can be accurately pruned in one-shot},
  author={Frantar, Elias and Alistarh, Dan},
  booktitle={International conference on machine learning},
  pages={10323--10337},
  year={2023},
  organization={PMLR}
}

@article{sanh2020movement,
  title={Movement pruning: Adaptive sparsity by fine-tuning},
  author={Sanh, Victor and Wolf, Thomas and Rush, Alexander},
  journal={Advances in neural information processing systems},
  volume={33},
  pages={20378--20389},
  year={2020}
}

@inproceedings{
frankle2019lottery,
title={The Lottery Ticket Hypothesis: Finding Sparse, Trainable Neural Networks},
author={Jonathan Frankle and Michael Carbin},
booktitle={International Conference on Learning Representations},
year={2019},
url={https://openreview.net/forum?id=rJl-b3RcF7},
}

@inproceedings{he2023spt,
  title={Sensitivity-aware visual parameter-efficient fine-tuning},
  author={He, Haoyu and Cai, Jianfei and Zhang, Jing and Tao, Dacheng and Zhuang, Bohan},
  booktitle={Proceedings of the IEEE/CVF international conference on computer vision},
  pages={11825--11835},
  year={2023}
}

@inproceedings{aghajanyan2021intrinsic,
    title = "Intrinsic Dimensionality Explains the Effectiveness of Language Model Fine-Tuning",
    author = "Aghajanyan, Armen  and
      Gupta, Sonal  and
      Zettlemoyer, Luke",
    editor = "Zong, Chengqing  and
      Xia, Fei  and
      Li, Wenjie  and
      Navigli, Roberto",
    booktitle = "Proceedings of the 59th Annual Meeting of the Association for Computational Linguistics and the 11th International Joint Conference on Natural Language Processing (Volume 1: Long Papers)",
    month = aug,
    year = "2021",
    address = "Online",
    publisher = "Association for Computational Linguistics",
    url = "https://aclanthology.org/2021.acl-long.568/",
    doi = "10.18653/v1/2021.acl-long.568",
    pages = "7319--7328",
}

@inproceedings{donoway2025elicitation,
  title={Quantifying elicitation of latent capabilities in language models},
  author={Donoway, Elizabeth and Joren, Hailey and Somani, Arushi and Sleight, Henry and Michael, Julian and DeWeese, Michael R and Schulman, John and Perez, Ethan and Roger, Fabien and Leike, Jan},
  booktitle={The Thirty-ninth Annual Conference on Neural Information Processing Systems},
  year={2025}
}

@article{shao2024deepseekmath,
  title={Deepseekmath: Pushing the limits of mathematical reasoning in open language models},
  author={Shao, Zhihong and Wang, Peiyi and Zhu, Qihao and Xu, Runxin and Song, Junxiao and Bi, Xiao and Zhang, Haowei and Zhang, Mingchuan and Li, YK and others},
  journal={arXiv preprint arXiv:2402.03300},
  year={2024}
}

@article{deepseek2025r1,
  title={DeepSeek-R1 incentivizes reasoning in LLMs through reinforcement learning},
  author={Guo, Daya and Yang, Dejian and Zhang, Haowei and Song, Junxiao and Wang, Peiyi and Zhu, Qihao and Xu, Runxin and Zhang, Ruoyu and Ma, Shirong and Bi, Xiao and others},
  journal={Nature},
  volume={645},
  number={8081},
  pages={633--638},
  year={2025},
  publisher={Nature Publishing Group UK London}
}

@article{buyukakyuz2024olora,
  title={Olora: Orthonormal low-rank adaptation of large language models},
  author={B{\"u}y{\"u}kaky{\"u}z, Kerim},
  journal={arXiv preprint arXiv:2406.01775},
  year={2024}
}

@inproceedings{
adilova2026olora,
title={{OL}o{RA}+: A Hybrid Approach to Parameter-Efficient Fine-Tuning of Large Language Models},
author={F.T. Adilova and Samariddin Kushmuratov},
booktitle={Conference of Mathematics of AI},
year={2026},
url={https://openreview.net/forum?id=c75JefyklT}
}

@inproceedings{
zhang2026loram,
title={The Primacy of Magnitude in Low-Rank Adaptation},
author={Zicheng Zhang and Haoran Li and Yifeng Zhang and Guoqiang Gong and Jiaxing Wang and Junxing Hu and pengzhang liu and Qixia Jiang},
booktitle={The Thirty-ninth Annual Conference on Neural Information Processing Systems},
year={2026},
url={https://openreview.net/forum?id=s4LnWgjacg}
}

@inproceedings{
hao2024flora,
title={Flora: Low-Rank Adapters Are Secretly Gradient Compressors},
author={Yongchang Hao and Yanshuai Cao and Lili Mou},
booktitle={Forty-first International Conference on Machine Learning},
year={2024},
url={https://openreview.net/forum?id=uubBZKM99Y}
}

@inproceedings{ren2024melora,
    title = "{MEL}o{RA}: Mini-Ensemble Low-Rank Adapters for Parameter-Efficient Fine-Tuning",
    author = "Ren, Pengjie  and
      Shi, Chengshun  and
      Wu, Shiguang  and
      Zhang, Mengqi  and
      Ren, Zhaochun  and
      de Rijke, Maarten  and
      Chen, Zhumin  and
      Pei, Jiahuan",
    editor = "Ku, Lun-Wei  and
      Martins, Andre  and
      Srikumar, Vivek",
    booktitle = "Proceedings of the 62nd Annual Meeting of the Association for Computational Linguistics (Volume 1: Long Papers)",
    month = aug,
    year = "2024",
    address = "Bangkok, Thailand",
    publisher = "Association for Computational Linguistics",
    url = "https://aclanthology.org/2024.acl-long.168/",
    doi = "10.18653/v1/2024.acl-long.168",
    pages = "3052--3064",
}

@inproceedings{li2024vblora,
      title={VB-LoRA: Extreme Parameter Efficient Fine-Tuning with Vector Banks}, 
      author={Yang Li and Shaobo Han and Shihao Ji},
      booktitle={The 38th Conference on Neural Information Processing Systems (NeurIPS)},
      year={2024}
}

@inproceedings{zaken2022bitfit,
    title = "{B}it{F}it: Simple Parameter-efficient Fine-tuning for Transformer-based Masked Language-models",
    author = "Ben Zaken, Elad  and
      Goldberg, Yoav  and
      Ravfogel, Shauli",
    editor = "Muresan, Smaranda  and
      Nakov, Preslav  and
      Villavicencio, Aline",
    booktitle = "Proceedings of the 60th Annual Meeting of the Association for Computational Linguistics (Volume 2: Short Papers)",
    month = may,
    year = "2022",
    address = "Dublin, Ireland",
    publisher = "Association for Computational Linguistics",
    url = "https://aclanthology.org/2022.acl-short.1/",
    doi = "10.18653/v1/2022.acl-short.1",
    pages = "1--9",
}

@inproceedings{guo2021diffpruning,
    title = "Parameter-Efficient Transfer Learning with Diff Pruning",
    author = "Guo, Demi  and
      Rush, Alexander  and
      Kim, Yoon",
    editor = "Zong, Chengqing  and
      Xia, Fei  and
      Li, Wenjie  and
      Navigli, Roberto",
    booktitle = "Proceedings of the 59th Annual Meeting of the Association for Computational Linguistics and the 11th International Joint Conference on Natural Language Processing (Volume 1: Long Papers)",
    month = aug,
    year = "2021",
    address = "Online",
    publisher = "Association for Computational Linguistics",
    url = "https://aclanthology.org/2021.acl-long.378/",
    doi = "10.18653/v1/2021.acl-long.378",
    pages = "4884--4896",
}

@article{pan2024lisa,
title={LISA: Layerwise Importance Sampling for Memory-Efficient Large Language Model Fine-Tuning},
author={Rui Pan and Xiang Liu and Shizhe Diao and Renjie Pi and Jipeng Zhang and Chi Han and Tong Zhang},
journal={arXiv preprint arXiv:2403.17919},
year={2024}
}

@article{dettmers2023qlora,
  title={Qlora: Efficient finetuning of quantized llms},
  author={Dettmers, Tim and Pagnoni, Artidoro and Holtzman, Ari and Zettlemoyer, Luke},
  journal={Advances in neural information processing systems},
  volume={36},
  pages={10088--10115},
  year={2023}
}

@inproceedings{
zhao2024galore,
title={GaLore: Memory-Efficient {LLM} Training by Gradient Low-Rank Projection},
author={Jiawei Zhao and Zhenyu Zhang and Beidi Chen and Zhangyang Wang and Anima Anandkumar and Yuandong Tian},
booktitle={Forty-first International Conference on Machine Learning},
year={2024},
url={https://openreview.net/forum?id=hYHsrKDiX7}
}

@misc{
jiang2024mora,
title={Mo{RA}: High-Rank Updating for Parameter-Efficient Fine-Tuning},
author={Ting Jiang and Shaohan Huang and Shengyue Luo and Zihan Zhang and Haizhen Huang and Furu Wei and Weiwei Deng and Feng Sun and Qi Zhang and Songtao Wang and deqing wang and Fuzhen Zhuang},
year={2024},
url={https://openreview.net/forum?id=SxOrhLuuVz}
}

@inproceedings{
rafailov2023dpo,
title={Direct Preference Optimization: Your Language Model is Secretly a Reward Model},
author={Rafael Rafailov and Archit Sharma and Eric Mitchell and Christopher D Manning and Stefano Ermon and Chelsea Finn},
booktitle={Thirty-seventh Conference on Neural Information Processing Systems},
year={2023},
url={https://openreview.net/forum?id=HPuSIXJaa9}
}

@inproceedings{li2024remax,
author = {Li, Ziniu and Xu, Tian and Zhang, Yushun and Lin, Zhihang and Yu, Yang and Sun, Ruoyu and Luo, Zhi-Quan},
title = {ReMax: a simple, effective, and efficient reinforcement learning method for aligning large language models},
year = {2024},
publisher = {JMLR.org},
booktitle = {Proceedings of the 41st International Conference on Machine Learning},
articleno = {1172},
numpages = {36},
location = {Vienna, Austria},
series = {ICML'24}
}

@article{santacroce2023efficient_rlhf,
  title={Efficient rlhf: Reducing the memory usage of ppo},
  author={Santacroce, Michael and Lu, Yadong and Yu, Han and Li, Yuanzhi and Shen, Yelong},
  journal={arXiv preprint arXiv:2309.00754},
  year={2023}
}

@inproceedings{
li2018intrinsic,
title={Measuring the Intrinsic Dimension of Objective Landscapes},
author={Chunyuan Li and Heerad Farkhoor and Rosanne Liu and Jason Yosinski},
booktitle={International Conference on Learning Representations},
year={2018},
url={https://openreview.net/forum?id=ryup8-WCW},
}

@inproceedings{jang2024lora_ntk,
  title={LoRA Training in the NTK Regime has No Spurious Local Minima},
  author={Jang, Uijeong and Lee, Jason D and Ryu, Ernest K},
  booktitle={International Conference on Machine Learning},
  pages={21306--21328},
  year={2024},
  organization={PMLR}
}

@article{
biderman2024lora_learns_less,
title={Lo{RA} Learns Less and Forgets Less},
author={Dan Biderman and Jacob Portes and Jose Javier Gonzalez Ortiz and Mansheej Paul and Philip Greengard and Connor Jennings and Daniel King and Sam Havens and Vitaliy Chiley and Jonathan Frankle and Cody Blakeney and John Patrick Cunningham},
journal={Transactions on Machine Learning Research},
issn={2835-8856},
year={2024},
url={https://openreview.net/forum?id=aloEru2qCG},
note={Featured Certification}
}

@article{
kirkpatrick2017ewc,
author = {James Kirkpatrick  and Razvan Pascanu  and Neil Rabinowitz  and Joel Veness  and Guillaume Desjardins  and Andrei A. Rusu  and Kieran Milan  and John Quan  and Tiago Ramalho  and Agnieszka Grabska-Barwinska  and Demis Hassabis  and Claudia Clopath  and Dharshan Kumaran  and Raia Hadsell },
title = {Overcoming catastrophic forgetting in neural networks},
journal = {Proceedings of the National Academy of Sciences},
volume = {114},
number = {13},
pages = {3521-3526},
year = {2017},
doi = {10.1073/pnas.1611835114},
URL = {https://www.pnas.org/doi/abs/10.1073/pnas.1611835114},
eprint = {https://www.pnas.org/doi/pdf/10.1073/pnas.1611835114},
}

@article{matena2022fisher_merge,
  title={Merging models with fisher-weighted averaging},
  author={Matena, Michael S and Raffel, Colin A},
  journal={Advances in Neural Information Processing Systems},
  volume={35},
  pages={17703--17716},
  year={2022}
}

@inproceedings{
zeng2024expressive,
title={The Expressive Power of Low-Rank Adaptation},
author={Yuchen Zeng and Kangwook Lee},
booktitle={The Twelfth International Conference on Learning Representations},
year={2024},
url={https://openreview.net/forum?id=likXVjmh3E}
}

@inproceedings{
sharma2024laser,
title={The Truth is in There: Improving Reasoning in Language Models with Layer-Selective Rank Reduction},
author={Pratyusha Sharma and Jordan T. Ash and Dipendra Misra},
booktitle={The Twelfth International Conference on Learning Representations},
year={2024},
url={https://openreview.net/forum?id=ozX92bu8VA}
}

@article{elhage2021mathematical,
   title={A Mathematical Framework for Transformer Circuits},
   author={Elhage, Nelson and Nanda, Neel and Olsson, Catherine and Henighan, Tom and Joseph, Nicholas and Mann, Ben and Askell, Amanda and Bai, Yuntao and Chen, Anna and Conerly, Tom and DasSarma, Nova and Drain, Dawn and Ganguli, Deep and Hatfield-Dodds, Zac and Hernandez, Danny and Jones, Andy and Kernion, Jackson and Lovitt, Liane and Ndousse, Kamal and Amodei, Dario and Brown, Tom and Clark, Jack and Kaplan, Jared and McCandlish, Sam and Olah, Chris},
   year={2021},
   journal={Transformer Circuits Thread},
   note={https://transformer-circuits.pub/2021/framework/index.html}
}

@misc{unsloth,
  author = {Daniel Han, Michael Han and Unsloth team},
  title = {Unsloth},
  year = {2023},
  url = {https://github.com/unslothai/unsloth},
}

@misc{trl,
  title   = {{TRL: Transformers Reinforcement Learning}},
  author  = {von Werra, Leandro and Belkada, Younes and Tunstall, Lewis and Beeching, Edward and Thrush, Tristan and Lambert, Nathan and Huang, Shengyi and Rasul, Kashif and Gallouédec, Quentin},
  license = {Apache-2.0},
  url     = {https://github.com/huggingface/trl},
  year    = {2020}
}

@inproceedings{
dapo,
title={{DAPO}: An Open-Source {LLM} Reinforcement Learning System at Scale},
author={Qiying Yu and Zheng Zhang and Ruofei Zhu and Yufeng Yuan and Xiaochen Zuo and YuYue and Weinan Dai and Tiantian Fan and Gaohong Liu and Juncai Liu and LingJun Liu and Xin Liu and Haibin Lin and Zhiqi Lin and Bole Ma and Guangming Sheng and Yuxuan Tong and Chi Zhang and Mofan Zhang and Ru Zhang and Wang Zhang and Hang Zhu and Jinhua Zhu and Jiaze Chen and Jiangjie Chen and Chengyi Wang and Hongli Yu and Yuxuan Song and Xiangpeng Wei and Hao Zhou and Jingjing Liu and Wei-Ying Ma and Ya-Qin Zhang and Lin Yan and Yonghui Wu and Mingxuan Wang},
booktitle={The Thirty-ninth Annual Conference on Neural Information Processing Systems},
year={2026},
url={https://openreview.net/forum?id=2a36EMSSTp}
}

@inproceedings{drgrpo,
  title={Understanding r1-zero-like training: A critical perspective},
  author={Liu, Zichen and Chen, Changyu and Li, Wenjun and Qi, Penghui and Pang, Tianyu and Du, Chao and Lee, Wee Sun and Lin, Min},
  booktitle={Conference on Language Modeling (COLM)},
  year={2025}
}

@article{cobbe2021gsm8k,
  title={Training Verifiers to Solve Math Word Problems},
  author={Cobbe, Karl and Kosaraju, Vineet and Bavarian, Mohammad and Chen, Mark and Jun, Heewoo and Kaiser, Lukasz and Plappert, Matthias and Tworek, Jerry and Hilton, Jacob and Nakano, Reiichiro and Hesse, Christopher and Schulman, John},
  journal={arXiv preprint arXiv:2110.14168},
  year={2021}
}

@article{hendrycks2021math,
  title={Measuring Mathematical Problem Solving With the MATH Dataset},
  author={Dan Hendrycks and Collin Burns and Saurav Kadavath and Akul Arora and Steven Basart and Eric Tang and Dawn Song and Jacob Steinhardt},
  journal={NeurIPS},
  year={2021}
}

@misc{
bnpo,
title={{BNPO}: Beta Normalization Policy Optimization},
author={Changyi Xiao and Mengdi Zhang and Yixin Cao},
year={2026},
url={https://openreview.net/forum?id=eXigrPOI2G}
}

@misc{alpaca,
  author = {Rohan Taori and Ishaan Gulrajani and Tianyi Zhang and Yann Dubois and Xuechen Li and Carlos Guestrin and Percy Liang and Tatsunori B. Hashimoto },
  title = {Stanford Alpaca: An Instruction-following LLaMA model},
  year = {2023},
  publisher = {GitHub},
  journal = {GitHub repository},
  howpublished = {\url{https://github.com/tatsu-lab/stanford_alpaca}},
}

@inproceedings{chen2020lottery_bert,
 author = {Chen, Tianlong and Frankle, Jonathan and Chang, Shiyu and Liu, Sijia and Zhang, Yang and Wang, Zhangyang and Carbin, Michael},
 booktitle = {Advances in Neural Information Processing Systems},
 editor = {H. Larochelle and M. Ranzato and R. Hadsell and M.F. Balcan and H. Lin},
 pages = {15834--15846},
 publisher = {Curran Associates, Inc.},
 title = {The Lottery Ticket Hypothesis for Pre-trained BERT Networks},
 url = {https://proceedings.neurips.cc/paper_files/paper/2020/file/b6af2c9703f203a2794be03d443af2e3-Paper.pdf},
 volume = {33},
 year = {2020}
}

@inproceedings{lu2024mathvista,
  title = {MathVista: Evaluating Mathematical Reasoning of Foundation Models in Visual Contexts},
  author = {Lu, Pan and others},
  booktitle = {ICLR},
  year = {2024}
}

@misc{shihu2024mathllava,
      title={Math-LLaVA: Bootstrapping Mathematical Reasoning for Multimodal Large Language Models}, 
      author={Wenhao Shi and Zhiqiang Hu and Yi Bin and Junhua Liu and Yang Yang and See-Kiong Ng and Lidong Bing and Roy Ka-Wei Lee},
      year={2024},
      eprint={2406.17294},
      archivePrefix={arXiv},
      primaryClass={cs.CL}
}

@misc{qwen2.5,
    title = {Qwen2.5: A Party of Foundation Models},
    url = {https://qwenlm.github.io/blog/qwen2.5/},
    author = {Qwen Team},
    month = {September},
    year = {2024}
}

@article{grattafiori2024llama,
  title={The llama 3 herd of models},
  author={Grattafiori, Aaron and Dubey, Abhimanyu and Jauhri, Abhinav and Pandey, Abhinav and Kadian, Abhishek and Al-Dahle, Ahmad and Letman, Aiesha and Mathur, Akhil and Schelten, Alan and Vaughan, Alex and others},
  journal={arXiv preprint arXiv:2407.21783},
  year={2024}
}

@misc{qwen2.5-VL,
    title = {Qwen2.5-VL},
    url = {https://qwenlm.github.io/blog/qwen2.5-vl/},
    author = {Qwen Team},
    month = {January},
    year = {2025}
}

@misc{eval-harness,
  author       = {Gao, Leo and Tow, Jonathan and Abbasi, Baber and Biderman, Stella and Black, Sid and DiPofi, Anthony and Foster, Charles and Golding, Laurence and Hsu, Jeffrey and Le Noac'h, Alain and Li, Haonan and McDonell, Kyle and Muennighoff, Niklas and Ociepa, Chris and Phang, Jason and Reynolds, Laria and Schoelkopf, Hailey and Skowron, Aviya and Sutawika, Lintang and Tang, Eric and Thite, Anish and Wang, Ben and Wang, Kevin and Zou, Andy},
  title        = {The Language Model Evaluation Harness},
  month        = 07,
  year         = 2024,
  publisher    = {Zenodo},
  version      = {v0.4.3},
  doi          = {10.5281/zenodo.12608602},
  url          = {https://zenodo.org/records/12608602}
}

@article{mmlu,
  title={Measuring Massive Multitask Language Understanding},
  author={Dan Hendrycks and Collin Burns and Steven Basart and Andy Zou and Mantas Mazeika and Dawn Song and Jacob Steinhardt},
  journal={Proceedings of the International Conference on Learning Representations (ICLR)},
  year={2021}
}

@inproceedings{piqa,
    author = {Yonatan Bisk and Rowan Zellers and
            Ronan Le Bras and Jianfeng Gao
            and Yejin Choi},
    title = {PIQA: Reasoning about Physical Commonsense in
           Natural Language},
    booktitle = {Thirty-Fourth AAAI Conference on
               Artificial Intelligence},
    year = {2020},
}

@misc{aime_2024,
  author = {Maxwell Jia},
  title = {AIME Problem Set 2024},
  year = {2024},
  publisher = {Huggingface},
  url = {https://huggingface.co/datasets/Maxwell-Jia/AIME_2024}
}

@misc{aime_2025,
  author = {math-ai},
  title = {AIME Problem Set 2025},
  year = {2025},
  publisher = {Huggingface},
  url = {https://huggingface.co/datasets/math-ai/aime25}
}

@article{arc,
  title={Think you have Solved Question Answering? Try ARC, the AI2 Reasoning Challenge},
  author={Peter Clark and Isaac Cowhey and Oren Etzioni and Tushar Khot and Ashish Sabharwal and Carissa Schoenick and Oyvind Tafjord},
  journal={ArXiv},
  year={2018},
  volume={abs/1803.05457}
}

@inproceedings{hellaswag,
    title={HellaSwag: Can a Machine Really Finish Your Sentence?},
    author={Zellers, Rowan and Holtzman, Ari and Bisk, Yonatan and Farhadi, Ali and Choi, Yejin},
    booktitle ={Proceedings of the 57th Annual Meeting of the Association for Computational Linguistics},
    year={2019}
}

@article{winogrande,
    title={WinoGrande: An Adversarial Winograd Schema Challenge at Scale},
    author={Sakaguchi, Keisuke and Bras, Ronan Le and Bhagavatula, Chandra and Choi, Yejin},
    journal={arXiv preprint arXiv:1907.10641},
    year={2019}
}

@inproceedings{truthfulqa,
    title = "{T}ruthful{QA}: Measuring How Models Mimic Human Falsehoods",
    author = "Lin, Stephanie  and
      Hilton, Jacob  and
      Evans, Owain",
    booktitle = "Proceedings of the 60th Annual Meeting of the Association for Computational Linguistics (Volume 1: Long Papers)",
    month = may,
    year = "2022",
    address = "Dublin, Ireland",
    publisher = "Association for Computational Linguistics",
    url = "https://aclanthology.org/2022.acl-long.229",
    doi = "10.18653/v1/2022.acl-long.229",
    pages = "3214--3252",
}

@inproceedings{boolq,
  title =     {BoolQ: Exploring the Surprising Difficulty of Natural Yes/No Questions},
  author =    {Clark, Christopher and Lee, Kenton and Chang, Ming-Wei and Kwiatkowski, Tom and Collins, Michael and Toutanova, Kristina},
  booktitle = {NAACL},
  year =      {2019},
}

@inproceedings{rein2024gpqa,
      title={{GPQA}: A Graduate-Level Google-Proof Q\&A Benchmark},
      author={David Rein and Betty Li Hou and Asa Cooper Stickland and Jackson Petty and Richard Yuanzhe Pang and Julien Dirani and Julian Michael and Samuel R. Bowman},
      booktitle={First Conference on Language Modeling},
      year={2024},
      url={https://openreview.net/forum?id=Ti67584b98}
}

@misc{amc2023dataset,
  author = {Mathematical Association of America (MAA)},
  title = {AMC 2023: American Mathematics Competitions 2023 Dataset},
  year = {2023},
  howpublished = {\url{https://huggingface.co/datasets/math-ai/amc23}},
  note = {Accessed: 2024-05-05}
}

@misc{bespoke_stratos,  
    author = {Bespoke Labs},  
    title = {Bespoke-Stratos: The unreasonable effectiveness of reasoning distillation},  
    howpublished = {https://www.bespokelabs.ai/blog/bespoke-stratos-the-unreasonable-effectiveness-of-reasoning-distillation},  
    note = {Accessed: 2025-01-22},  
    year = {2025}
}

%%%%%%%%%%%%%%%%%%%%%%%%%%%%%%%%%%%%%%%%%%%%%%%%%%%%%%%%%%%%
\newpage
\appendix

\section*{Appendices}
\etocsetnexttocdepth{section} % section level only
\etocsetlocaltop.toc{part}    % make following sections one local block
\etocsetstyle{section}
  {}
  {\noindent}{Appendix~\etocnumber\quad\etocname\dotfill\etocpage\par\vspace{-0.4em}}
  {}
\localtableofcontents

\section{Extended Related Work}
\label{app:related}

We provide a broader survey of LoRA variants and parameter-efficient fine-tuning methods, organized by the axis of improvement each addresses.

\paragraph{Initialization.}
Standard LoRA initializes $A$ from $\mathcal{N}(0, \sigma^2)$ and $B = 0$.  PiSSA \citep{meng2024pissa} initializes both factors from the principal singular vectors of the pre-trained weight, training in the most important subspace.  OLoRA \citep{adilova2026olora, buyukakyuz2024olora} uses QR decomposition for orthogonal initialization.  CorDA \citep{yang2024corda} performs task-context-aware SVD, orienting the decomposition toward the target data distribution.  LoRAM \citep{zhang2026loram} incorporates weight magnitude information into initialization.  These methods improve where in weight space the adapter starts; our work addresses which elements of the adapter are active.

\paragraph{Training dynamics.}
LoRA+ \citep{hayou2024loraplus} sets different learning rates for $A$ and $B$, correcting an inefficiency in vanilla LoRA.  rsLoRA \citep{kalajdzievski2023rslora} introduces a $1/\sqrt{r}$ scaling factor to stabilize training across ranks.  DoRA \citep{liu2024dora} decomposes weights into magnitude and direction, applying LoRA only to the directional component.  FLoRA \citep{hao2024flora} resamples the $A$ projection matrix periodically, interpreting LoRA as gradient compression.  These are orthogonal to our element-level selection and could in principle be combined with circuit-based sparsity.

\paragraph{Rank allocation.}
AdaLoRA \citep{zhang2023adalora} parameterizes adapters in SVD form and prunes singular values based on importance, allocating higher rank to more critical layers.  DyLoRA \citep{valipour2023dylora} trains across a range of ranks simultaneously.  SoRA \citep{ding2023sora} learns sparse gates over rank-1 components.  MELoRA \citep{ren2024melora} uses mini-ensembles of very low-rank adapters.  These methods allocate budget at the rank or layer level; our circuits allocate at the individual element level.

\paragraph{Extreme parameter efficiency.}
VeRA \citep{kopiczko2024vera} achieves $\sim$10$\times$ reduction over LoRA by sharing frozen random matrices.  FourierFT \citep{gao2024fourierft} uses spectral coefficients over a Fourier basis.  LoRA-XS \citep{balazy2024loraxs} trains only a tiny $r \times r$ middle matrix.  VB-LoRA \citep{li2024vblora} composes adapters from a shared vector bank.  RandLoRA \citep{albert2025randlora} learns element-wise scaling over frozen random factors.  \citet{morris2026tinylora} scale LoRA down to a single parameter.  BitFit \citep{zaken2022bitfit} trains only bias terms ($<$0.1\% of parameters).  Our sparse circuits operate in the same ultra-low-parameter regime but with a standard LoRA parameterization and informed selection.

\paragraph{Adapter compression and pruning.}
SparseAdapter \citep{he2022sparseadapter} prunes adapter weights post-training.  LoRA-Drop \citep{zhou2024loradrop} removes entire LoRA layers based on output contribution.  LoRA-Prune \citep{zhang2024loraprune} jointly prunes the base model and fine-tunes with LoRA.  Diff pruning \citep{guo2021diffpruning} learns a sparse weight diff with $L_0$ regularization.  These are complementary to our pre-training circuit selection.

\paragraph{Parameter selection beyond adapters.}
LISA \citep{pan2024lisa} importance-samples which layers to update at each step.  LASER \citep{sharma2024laser} shows that removing higher-order weight components at specific layers can improve performance.  Movement pruning \citep{sanh2020movement} selects fine-tuning parameters based on gradient-weight products.  The lottery ticket hypothesis \citep{frankle2019lottery} and its extensions to pre-trained models \citep{chen2020lottery_bert} show that sparse trainable subnetworks exist within dense networks.  Our element-level circuit discovery within LoRA adapters is a new instantiation of this broad family.

\paragraph{Memory-efficient training.}
QLoRA \citep{dettmers2023qlora} combines 4-bit quantization with LoRA.  GaLore \citep{zhao2024galore} projects gradients into low-rank subspaces, enabling full-parameter training with LoRA-level memory.  MoRA \citep{jiang2024mora} achieves high-rank updates via a square matrix with non-parametric operators.  These address orthogonal memory concerns and could be combined with circuit-based sparsity.

\paragraph{RL fine-tuning.}
GRPO \citep{shao2024deepseekmath} uses group-relative rewards to eliminate the critic.  DeepSeek-R1 \citep{deepseek2025r1} applies GRPO at scale for reasoning.  DPO \citep{rafailov2023dpo} reformulates RLHF as a classification problem.  ReMax \citep{li2024remax} simplifies REINFORCE for LLM alignment.  Efficient RLHF methods \citep{santacroce2023efficient_rlhf} use LoRA for both policy and value networks.  Our work studies how parameter placement within LoRA interacts with the RL training signal.

\paragraph{Theoretical foundations.}
\citet{aghajanyan2021intrinsic} establish low intrinsic dimensionality of fine-tuning.  \citet{li2018intrinsic} introduce intrinsic dimension measurement for loss landscapes.  \citet{jang2024lora_ntk} prove LoRA training has no spurious local minima in the NTK regime.  \citet{zeng2024expressive} analyze the expressive power of low-rank adaptation.  \citet{biderman2024lora_learns_less} show LoRA learns less but forgets less than full fine-tuning.  Fisher information for parameter importance dates to Optimal Brain Damage \citep{lecun1990obd} and has been applied to continual learning \citep{kirkpatrick2017ewc} and model merging \citep{matena2022fisher_merge}.

 \paragraph{Circuits and Mechanistic Interpretability}
Parameter selection is tied methodologically to Mechanistic Interpretability, which aims to explain how models perform given tasks. Circuits, minimal subgraphs of a model’s computation graph that execute a specific task \citep{haklay2025positionawareautomaticcircuitdiscovery}, are found using ablations or interventions in the model activations, often requiring exhaustive searches. These circuits generally correspond to a specific task or capability \citep{cammarata2021curve,wang2022interpretability,gurnee2025when}. In our work, we mean something different by ``circuit''. Instead of a subgraph of the total computation graph, we aim to find, at a given budget, a set of neurons within LORA $\mathbf{BA}$ matrices that contribute most to efficient learning. This set is not necessarily unique, minimal, or complete.
% TODO: \citet not working?
\cite{ge2024automaticallyidentifyinglocalglobal,haklay2025positionawareautomaticcircuitdiscovery,kharlapenko2025scalingsparsefeaturecircuit} propose methods for finding circuits automatically (whereas initial methods require manual exploration \citep{nanda2023progress,kramr2024atpefficientscalablemethod}). Our work repurposes neuron attribution techniques to identify parameter subsets for more effective PEFT.

\paragraph{Sparse and pruned adapters.}
SparseAdapter \citep{he2022sparseadapter} prunes adapter weights after training, showing that adapters are highly compressible.  LoRA-Drop \citep{zhou2024loradrop} drops entire LoRA layers based on output contribution. LoRA-Prune \citep{zhang2024loraprune} jointly prunes the base model and adapter.  S$^2$FT \citep{yang2024s2ft} selects structured groups (attention heads and FFN channels) in the full weight matrices and permutes them into dense submatrices for efficient training; our work operates at individual elements within LoRA adapters rather than structured blocks in the full model.  These methods determine sparsity during or after training.  In contrast, our circuit discovery operates before training begins, selecting elements at initialization based on gradient statistics.  This means the sparsity pattern is fixed throughout training and requires no iterative pruning schedule.

\clearpage
\section{Detailed Results}\label{app:full_results}
\localtableofcontents

\subsection{Instruction Tuning: Full Results}
\label{app:sft_7bench}
We conduct instruction tuning on 7 well-known datasets : MMLU~\citep{mmlu}, PiQA~\citep{piqa}, TruthfulQA~\citep{truthfulqa}, HellaSwag~\citep{hellaswag}, Arc-Challenge~\citep{arc}, BoolQ~\citep{boolq} and Winogrande~\citep{winogrande}. \Cref{tab:sft_7bench_full} reports per-benchmark SFT accuracy on Alpaca (mean $\pm$ SE, 5 seeds). We see that the seeds are extremely stable across all configurations. In all cases the circuits edge out the random placement. 

\begin{table}[h]
\centering
\caption{\textbf{Per-benchmark SFT accuracy (\%) on Alpaca} (5 seeds).  Subscripts denote standard error.}
\label{tab:sft_7bench_full}
\vspace{4pt}
\scriptsize
\setlength{\tabcolsep}{3pt}
\begin{tabular}{ll ccccccc c}
\toprule
\textbf{Model} & \textbf{Method} & \textbf{ARC-C} & \textbf{BoolQ} & \textbf{HellaS} & \textbf{MMLU} & \textbf{PIQA} & \textbf{TQA} & \textbf{Wino} & \textbf{Avg} \\
\midrule
\multirow{4}{*}{\rotatebox[origin=c]{90}{\footnotesize Q2.5-1.5B}}
  & $\hat{F}$  & 46.4$_{\pm 0.2}$ & 73.2$_{\pm 0.6}$ & 67.9$_{\pm 0.0}$ & 59.6$_{\pm 0.1}$ & 75.9$_{\pm 0.1}$ & 47.9$_{\pm 0.2}$ & 63.6$_{\pm 0.3}$ & 62.1$_{\pm 0.1}$ \\
  & $\hat{S}$     & 46.7$_{\pm 0.8}$ & 73.3$_{\pm 0.6}$ & 67.9$_{\pm 0.0}$ & 59.0$_{\pm 0.2}$ & 76.1$_{\pm 0.1}$ & 47.2$_{\pm 0.5}$ & 63.4$_{\pm 0.2}$ & 61.9$_{\pm 0.2}$ \\
  & Random & 45.6$_{\pm 0.5}$ & 72.7$_{\pm 0.9}$ & 67.8$_{\pm 0.0}$ & 59.1$_{\pm 0.1}$ & 76.0$_{\pm 0.2}$ & 47.0$_{\pm 0.4}$ & 63.4$_{\pm 0.0}$ & 61.6$_{\pm 0.2}$ \\
  & Full LoRA     & 46.6$_{\pm 0.5}$ & 75.3$_{\pm 0.3}$ & 68.1$_{\pm 0.1}$ & 59.9$_{\pm 0.0}$ & 76.0$_{\pm 0.0}$ & 47.7$_{\pm 0.5}$ & 63.0$_{\pm 0.1}$ & \textbf{62.4$_{\pm 0.2}$} \\
\midrule
\multirow{4}{*}{\rotatebox[origin=c]{90}{\footnotesize Q2.5-3B}}
  & $\hat{F}$     & 48.8$_{\pm 0.1}$ & 76.7$_{\pm 0.1}$ & 73.8$_{\pm 0.0}$ & 65.0$_{\pm 0.0}$ & 78.9$_{\pm 0.1}$ & 49.7$_{\pm 0.1}$ & 69.1$_{\pm 0.1}$ & 66.0$_{\pm 0.0}$ \\
  & $\hat{S}$     & 48.2$_{\pm 0.1}$ & 78.7$_{\pm 0.2}$ & 74.0$_{\pm 0.0}$ & 65.0$_{\pm 0.1}$ & 78.9$_{\pm 0.0}$ & 49.9$_{\pm 0.2}$ & 68.9$_{\pm 0.1}$ & 66.2$_{\pm 0.0}$ \\
  & Random   & 47.7$_{\pm 0.2}$ & 77.5$_{\pm 0.1}$ & 73.7$_{\pm 0.0}$ & 65.0$_{\pm 0.0}$ & 78.8$_{\pm 0.1}$ & 49.7$_{\pm 0.2}$ & 68.4$_{\pm 0.2}$ & 65.8$_{\pm 0.1}$ \\
  & Full LoRA     & 48.7$_{\pm 0.4}$ & 79.4$_{\pm 0.4}$ & 74.0$_{\pm 0.0}$ & 65.0$_{\pm 0.1}$ & 78.8$_{\pm 0.1}$ & 51.2$_{\pm 0.3}$ & 69.3$_{\pm 0.3}$ & \textbf{66.7$_{\pm 0.2}$} \\
\midrule
\multirow{4}{*}{\rotatebox[origin=c]{90}{\footnotesize Q2.5-7B}}
  & $\hat{F}$     & 52.8$_{\pm 0.1}$ & 84.7$_{\pm 0.1}$ & 79.1$_{\pm 0.0}$ & 71.8$_{\pm 0.1}$ & 80.5$_{\pm 0.1}$ & 57.5$_{\pm 0.2}$ & 74.0$_{\pm 0.1}$ & \textbf{71.5}$_{\pm 0.0}$ \\
  & $\hat{S}$     & 52.5$_{\pm 0.2}$ & 85.2$_{\pm 0.1}$ & 79.1$_{\pm 0.0}$ & 71.9$_{\pm 0.1}$ & 79.9$_{\pm 0.0}$ & 57.5$_{\pm 0.2}$ & 73.4$_{\pm 0.1}$ & 71.4$_{\pm 0.1}$ \\
  & Random        & 51.8$_{\pm 0.1}$ & 85.1$_{\pm 0.0}$ & 78.9$_{\pm 0.0}$ & 71.8$_{\pm 0.1}$ & 80.0$_{\pm 0.0}$ & 57.1$_{\pm 0.1}$ & 73.2$_{\pm 0.1}$ & 71.1$_{\pm 0.0}$ \\
  & Full LoRA     & 52.8$_{\pm 0.1}$ & 85.3$_{\pm 0.1}$ & 79.1$_{\pm 0.0}$ & 71.7$_{\pm 0.1}$ & 80.4$_{\pm 0.1}$ & 58.3$_{\pm 0.2}$ & 72.5$_{\pm 0.2}$ & \textbf{71.5$_{\pm 0.0}$} \\
\midrule
\multirow{4}{*}{\rotatebox[origin=c]{90}{\footnotesize L3.2-1B}}
  & $\hat{F}$     & 37.7$_{\pm 0.3}$ & 65.4$_{\pm 0.3}$ & 63.1$_{\pm 0.1}$ & 35.8$_{\pm 0.4}$ & 75.7$_{\pm 0.2}$ & 40.3$_{\pm 0.3}$ & 61.2$_{\pm 0.3}$ & 54.2$_{\pm 0.1}$ \\
  & $\hat{S}$     & 37.7$_{\pm 0.2}$ & 66.1$_{\pm 0.3}$ & 63.0$_{\pm 0.2}$ & 35.6$_{\pm 0.2}$ & 75.6$_{\pm 0.2}$ & 41.2$_{\pm 0.2}$ & 61.5$_{\pm 0.1}$ & 54.4$_{\pm 0.1}$ \\
  & Random        & 37.2$_{\pm 0.2}$ & 64.5$_{\pm 0.2}$ & 63.0$_{\pm 0.0}$ & 37.6$_{\pm 0.1}$ & 75.2$_{\pm 0.2}$ & 39.4$_{\pm 0.1}$ & 61.1$_{\pm 0.2}$ & 54.0$_{\pm 0.1}$ \\
  & Full LoRA     & 37.4$_{\pm 0.3}$ & 65.7$_{\pm 0.5}$ & 63.4$_{\pm 0.1}$ & 37.0$_{\pm 0.1}$ & 75.6$_{\pm 0.2}$ & 40.8$_{\pm 0.2}$ & 61.2$_{\pm 0.1}$ & \textbf{54.5$_{\pm 0.1}$} \\
\midrule
\multirow{4}{*}{\rotatebox[origin=c]{90}{\footnotesize L3.2-3B}}
  & $\hat{F}$     & 46.0$_{\pm 0.2}$ & 77.7$_{\pm 0.3}$ & 73.0$_{\pm 0.1}$ & 54.6$_{\pm 0.3}$ & 78.2$_{\pm 0.1}$ & 45.7$_{\pm 0.3}$ & 70.6$_{\pm 0.2}$ & 63.7$_{\pm 0.1}$ \\
  & $\hat{S}$     & 46.6$_{\pm 0.1}$ & 77.0$_{\pm 0.3}$ & 73.2$_{\pm 0.1}$ & 54.5$_{\pm 0.3}$ & 78.3$_{\pm 0.1}$ & 46.3$_{\pm 0.3}$ & 70.8$_{\pm 0.1}$ & 63.8$_{\pm 0.1}$ \\
  & Random        & 46.1$_{\pm 0.1}$ & 75.5$_{\pm 0.2}$ & 72.6$_{\pm 0.0}$ & 54.9$_{\pm 0.1}$ & 78.3$_{\pm 0.1}$ & 42.5$_{\pm 0.2}$ & 70.1$_{\pm 0.0}$ & 62.9$_{\pm 0.1}$ \\
  & Full LoRA     & 46.4$_{\pm 0.2}$ & 77.8$_{\pm 0.4}$ & 73.1$_{\pm 0.0}$ & 54.6$_{\pm 0.3}$ & 77.9$_{\pm 0.1}$ & 46.2$_{\pm 0.4}$ & 70.9$_{\pm 0.1}$ & \textbf{63.9$_{\pm 0.0}$} \\
\midrule
\multirow{4}{*}{\rotatebox[origin=c]{90}{\footnotesize L3.1-8B}}
  & $\hat{F}$     & 55.4$_{\pm 0.1}$ & 83.9$_{\pm 0.1}$ & 78.4$_{\pm 0.0}$ & 63.5$_{\pm 0.0}$ & 81.8$_{\pm 0.1}$ & 50.3$_{\pm 0.2}$ & 74.4$_{\pm 0.1}$ & 69.7$_{\pm 0.0}$ \\
  & $\hat{S}$     & 55.8$_{\pm 0.2}$ & 83.8$_{\pm 0.1}$ & 78.2$_{\pm 0.0}$ & 63.2$_{\pm 0.1}$ & 81.8$_{\pm 0.1}$ & 50.4$_{\pm 0.3}$ & 74.5$_{\pm 0.1}$ & 69.6$_{\pm 0.0}$ \\
  & Random        & 54.8$_{\pm 0.1}$ & 83.5$_{\pm 0.1}$ & 78.0$_{\pm 0.0}$ & 63.4$_{\pm 0.1}$ & 80.9$_{\pm 0.1}$ & 48.3$_{\pm 0.1}$ & 74.5$_{\pm 0.1}$ & 69.1$_{\pm 0.0}$ \\
  & Full LoRA     & 55.9$_{\pm 0.5}$ & 83.9$_{\pm 0.1}$ & 78.5$_{\pm 0.1}$ & 63.4$_{\pm 0.2}$ & 81.1$_{\pm 0.2}$ & 52.4$_{\pm 0.7}$ & 74.4$_{\pm 0.2}$ & \textbf{69.9$_{\pm 0.1}$} \\
\bottomrule
\end{tabular}
\vspace{4pt}

{\scriptsize ARC-C = ARC-Challenge (acc\_norm), HellaS = HellaSwag (acc\_norm), TQA = TruthfulQA MC2, Wino = WinoGrande, PIQA (acc\_norm).  Q2.5 = Qwen2.5, L3.2 = Llama-3.2, L3.1 = Llama-3.1.}
\end{table}

%% ============================================================
\subsection{OOD Generalization: Full Results}
\label{app:ood}
All models trained with GRPO, evaluated OOD with $n{=}64$ generations per problem at temperature 0.7.  Solid lines show pass@$k$; dashed lines show maj@$k$ (AMC panels only).  7B MATH-trained models omit Random, which did not learn at this scale (55.2\%, ${\approx}$base). We show results for AIME 2024~\citep{aime_2024}, AIME 2025~\citep{aime_2025} and AMC~\citep{hendrycks2021math}.

\begin{figure}[h]
\centering
% Auto-generated by pgf_ood_math.py — do not edit by hand.
% MATH-trained models evaluated OOD. pass@k + maj@k on AMC.
\begin{tikzpicture}
\begin{groupplot}[
    tinylora,
    group style={
        group size=3 by 2,
        horizontal sep=0.8cm,
        vertical sep=0.7cm,
    },
    width=0.34\textwidth,
    height=0.24\textwidth,
    legend columns=5,
    legend style={draw=black, font=\small, column sep=6pt},
]

    \nextgroupplot[
        title={AMC 2023},
        ylabel={pass@$k$ (\%)},
        ylabel style={align=center},
        xlabel={},
        xmode=log,
        xmin=0.8, xmax=80,
        ymin=0, ymax=105,
        xtick={1, 4, 8, 16, 32, 64},
        xticklabels={},
    ]
    \addplot[colorBase, dashed, thin, mark=none, mark size=2pt, forget plot]
        coordinates {(1,20.0) (4,47.5) (8,57.5) (16,67.5) (32,80.0) (64,85.0)};
    \addplot[colorBase, dashed, thin, mark=none, forget plot]
        coordinates {(4,20.0) (8,27.5) (16,35.0) (64,42.5)};
    \addplot[colorRandom, thick, mark=o, mark size=2pt, forget plot]
        coordinates {(1,22.5) (4,40.0) (8,50.0) (16,70.0) (32,80.0) (64,87.5)};
    \addplot[colorRandom, dashed, thin, mark=none, forget plot]
        coordinates {(4,22.5) (8,25.0) (16,40.0) (64,50.0)};
    \addplot[colorSignal, thick, mark=triangle*, mark size=2pt, forget plot]
        coordinates {(1,20.0) (4,47.5) (8,57.5) (16,65.0) (32,82.5) (64,90.0)};
    \addplot[colorSignal, dashed, thin, mark=none, forget plot]
        coordinates {(4,32.5) (8,37.5) (16,42.5) (64,47.5)};
    \addplot[colorFisher, thick, mark=square*, mark size=2pt, forget plot]
        coordinates {(1,35.0) (4,62.5) (8,65.0) (16,75.0) (32,92.5) (64,100.0)};
    \addplot[colorFisher, dashed, thin, mark=none, forget plot]
        coordinates {(4,40.0) (8,42.5) (16,52.5) (64,52.5)};
    \addplot[colorFullLoRA, thick, mark=diamond*, mark size=2pt, forget plot]
        coordinates {(1,32.5) (4,50.0) (8,57.5) (16,75.0) (32,85.0) (64,95.0)};
    \addplot[colorFullLoRA, dashed, thin, mark=none, forget plot]
        coordinates {(4,40.0) (8,40.0) (16,47.5) (64,47.5)};

    \nextgroupplot[
        title={AIME 2024},
        ylabel={},
        xlabel={},
        xmode=log,
        xmin=0.8, xmax=80,
        ymin=0, ymax=50,
        xtick={1, 4, 8, 16, 32, 64},
        xticklabels={},
    ]
    \addplot[colorBase, dashed, thin, mark=none, mark size=2pt, forget plot]
        coordinates {(1,0.0) (4,3.3) (8,10.0) (16,10.0) (32,20.0) (64,23.3)};
    \addplot[colorRandom, thick, mark=o, mark size=2pt, forget plot]
        coordinates {(1,0.0) (4,3.3) (8,6.7) (16,13.3) (32,30.0) (64,36.7)};
    \addplot[colorSignal, thick, mark=triangle*, mark size=2pt, forget plot]
        coordinates {(1,0.0) (4,6.7) (8,6.7) (16,20.0) (32,20.0) (64,30.0)};
    \addplot[colorFisher, thick, mark=square*, mark size=2pt, forget plot]
        coordinates {(1,3.3) (4,10.0) (8,16.7) (16,23.3) (32,30.0) (64,36.7)};
    \addplot[colorFullLoRA, thick, mark=diamond*, mark size=2pt, forget plot]
        coordinates {(1,3.3) (4,10.0) (8,10.0) (16,20.0) (32,26.7) (64,36.7)};

    \nextgroupplot[
        title={AIME 2025},
        ylabel={},
        xlabel={},
        xmode=log,
        xmin=0.8, xmax=80,
        ymin=0, ymax=50,
        xtick={1, 4, 8, 16, 32, 64},
        xticklabels={},
    ]
    \addplot[colorBase, dashed, thin, mark=none, mark size=2pt, forget plot]
        coordinates {(1,0.0) (4,6.7) (8,13.3) (16,16.7) (32,20.0) (64,26.7)};
    \addplot[colorRandom, thick, mark=o, mark size=2pt, forget plot]
        coordinates {(1,0.0) (4,0.0) (8,3.3) (16,13.3) (32,16.7) (64,23.3)};
    \addplot[colorSignal, thick, mark=triangle*, mark size=2pt, forget plot]
        coordinates {(1,0.0) (4,6.7) (8,13.3) (16,20.0) (32,23.3) (64,36.7)};
    \addplot[colorFisher, thick, mark=square*, mark size=2pt, forget plot]
        coordinates {(1,0.0) (4,6.7) (8,10.0) (16,20.0) (32,20.0) (64,30.0)};
    \addplot[colorFullLoRA, thick, mark=diamond*, mark size=2pt, forget plot]
        coordinates {(1,3.3) (4,3.3) (8,6.7) (16,13.3) (32,16.7) (64,20.0)};

    \nextgroupplot[
        title={},
        ylabel={pass@$k$ (\%)},
        ylabel style={align=center},
        xlabel={$k$},
        xmode=log,
        xmin=0.8, xmax=80,
        ymin=0, ymax=105,
        xtick={1, 4, 8, 16, 32, 64},
        xticklabels={1, 4, 8, 16, 32, 64},
    ]
    \addplot[colorBase, dashed, thin, mark=none, mark size=2pt, forget plot]
        coordinates {(1,27.5) (4,52.5) (8,70.0) (16,85.0) (32,92.5) (64,95.0)};
    \addplot[colorBase, dashed, thin, mark=none, forget plot]
        coordinates {(4,37.5) (8,47.5) (16,45.0) (32,50.0) (64,55.0)};
    \addplot[colorSignal, thick, mark=triangle*, mark size=2pt, forget plot]
        coordinates {(1,45.0) (4,70.0) (8,82.5) (16,87.5) (32,95.0) (64,97.5)};
    \addplot[colorSignal, dashed, thin, mark=none, forget plot]
        coordinates {(4,55.0) (8,55.0) (16,57.5) (32,62.5) (64,65.0)};
    \addplot[colorFisher, thick, mark=square*, mark size=2pt, forget plot]
        coordinates {(1,47.5) (4,75.0) (8,80.0) (16,85.0) (32,87.5) (64,95.0)};
    \addplot[colorFisher, dashed, thin, mark=none, forget plot]
        coordinates {(4,62.5) (8,62.5) (16,62.5) (32,62.5) (64,62.5)};
    \addplot[colorFullLoRA, thick, mark=diamond*, mark size=2pt, forget plot]
        coordinates {(1,47.5) (4,75.0) (8,80.0) (16,85.0) (32,95.0) (64,95.0)};
    \addplot[colorFullLoRA, dashed, thin, mark=none, forget plot]
        coordinates {(4,52.5) (8,55.0) (16,62.5) (32,65.0) (64,62.5)};

    \nextgroupplot[
        title={},
        ylabel={},
        xlabel={$k$},
        xmode=log,
        xmin=0.8, xmax=80,
        ymin=0, ymax=50,
        xtick={1, 4, 8, 16, 32, 64},
        xticklabels={1, 4, 8, 16, 32, 64},
    ]
    \addplot[colorBase, dashed, thin, mark=none, mark size=2pt, forget plot]
        coordinates {(1,0.0) (4,13.3) (8,16.7) (16,23.3) (32,33.3) (64,40.0)};
    \addplot[colorSignal, thick, mark=triangle*, mark size=2pt, forget plot]
        coordinates {(1,6.7) (4,20.0) (8,30.0) (16,33.3) (32,40.0) (64,46.7)};
    \addplot[colorFisher, thick, mark=square*, mark size=2pt, forget plot]
        coordinates {(1,13.3) (4,26.7) (8,26.7) (16,33.3) (32,33.3) (64,43.3)};
    \addplot[colorFullLoRA, thick, mark=diamond*, mark size=2pt, forget plot]
        coordinates {(1,13.3) (4,16.7) (8,23.3) (16,30.0) (32,40.0) (64,46.7)};

    \nextgroupplot[
        title={},
        ylabel={},
        xlabel={$k$},
        xmode=log,
        xmin=0.8, xmax=80,
        ymin=0, ymax=50,
        xtick={1, 4, 8, 16, 32, 64},
        xticklabels={1, 4, 8, 16, 32, 64},
        legend to name=ood_math_legend,
    ]
    \addplot[colorBase, dashed, thin, mark=none, mark size=2pt, forget plot]
        coordinates {(1,0.0) (4,3.3) (8,3.3) (16,16.7) (32,23.3) (64,40.0)};
    \addplot[colorSignal, thick, mark=triangle*, mark size=2pt, forget plot]
        coordinates {(1,3.3) (4,16.7) (8,23.3) (16,33.3) (32,40.0) (64,46.7)};
    \addplot[colorFisher, thick, mark=square*, mark size=2pt, forget plot]
        coordinates {(1,3.3) (4,23.3) (8,23.3) (16,23.3) (32,33.3) (64,43.3)};
    \addplot[colorFullLoRA, thick, mark=diamond*, mark size=2pt, forget plot]
        coordinates {(1,6.7) (4,20.0) (8,30.0) (16,36.7) (32,40.0) (64,43.3)};
    \addplot[colorBase, dashed, thin, mark=none, mark size=2pt]
        coordinates {(1,-999)};
    \addlegendentry{Base}
    \addplot[colorRandom, thick, mark=o, mark size=2pt]
        coordinates {(1,-999)};
    \addlegendentry{Random}
    \addplot[colorSignal, thick, mark=triangle*, mark size=2pt]
        coordinates {(1,-999)};
    \addlegendentry{$\hat{S}$}
    \addplot[colorFisher, thick, mark=square*, mark size=2pt]
        coordinates {(1,-999)};
    \addlegendentry{$\hat{F}$}
    \addplot[colorFullLoRA, thick, mark=diamond*, mark size=2pt]
        coordinates {(1,-999)};
    \addlegendentry{Full LoRA}

\end{groupplot}
\node[anchor=east, font=\small\bfseries, rotate=90]
    at ($(group c1r1.west) + (-1.3cm,0)$) {3B};
\node[anchor=east, font=\small\bfseries, rotate=90]
    at ($(group c1r2.west) + (-1.3cm,0)$) {7B};
\node at ($(group c1r2.south)!0.5!(group c3r2.south) + (0,-1.0cm)$)
    {\ref*{ood_math_legend}};
\node[font=\scriptsize] at ($(group c1r2.south)!0.5!(group c3r2.south) + (0,-1.5cm)$)
    {Solid: pass@$k$ \qquad Dashed: maj@$k$ (AMC only)};
\end{tikzpicture}
\caption{\textbf{MATH-trained pass@$k$ and maj@$k$} (Qwen2.5-3B/7B).  $\hat{F}$ at 50K reaches 100\% pass@64 on AMC (3B) and 95\% (7B).  On AIME 2025, Full LoRA degrades below base at 3B while $\hat{S}$ reaches 36.7\% pass@64; at 7B all methods improve over base.  AIME maj@$k$ is near-zero for all methods; AMC maj@64 peaks at 52.5\% ($\hat{F}$, 3B) and 65\% ($\hat{S}$, 7B).}
\label{fig:ood_math_passk}
\end{figure}
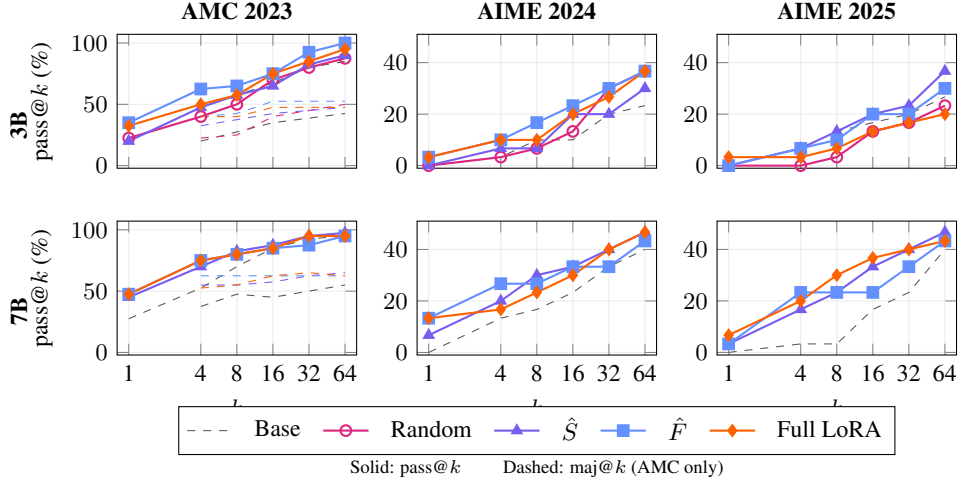

At 3B, $\hat{F}$ leads on AMC across all $k$ values and achieves 100\% pass@64.  On AIME 2025, Full LoRA falls below the base model from pass@8 onward, while $\hat{S}$ reaches 36.7\% pass@64 (vs.\ 20\% for Full LoRA)---consistent with sparse masks providing implicit regularization.  At 7B, circuits match Full LoRA across all three benchmarks; $\hat{S}$ slightly exceeds Full LoRA on AIME 2025 (46.7\% vs.\ 43.3\% pass@64).

\subsubsection{Cross-Task Transfer}
\label{app:ood_cross}

\begin{figure}[h]
\centering
% Auto-generated by pgf_cross_task.py — do not edit by hand.
% Cross-task greedy transfer: MATH <-> GSM8K.
\begin{tikzpicture}
\begin{groupplot}[
    tinylora,
    group style={
        group size=2 by 1,
        horizontal sep=1.2cm,
    },
    width=0.48\textwidth,
    height=0.36\textwidth,
    legend columns=5,
    legend style={draw=black, font=\small, column sep=6pt},
]

    \nextgroupplot[
        title={MATH $\to$ GSM8K},
        ylabel={Accuracy (\%)},
        ymin=55, ymax=95,
        ymajorgrids=true,
        ytick={55, 65, 75, 85, 95},
        symbolic x coords={3B, 7B},
        xtick=data,
        enlarge x limits=0.7,
    ]
    \addplot[ybar, bar width=7pt, fill=colorBase, draw=colorBase!70!black, bar shift=-17.0pt, forget plot]
        coordinates {(3B,62.1) (7B,79.3)};
    \addplot[ybar, bar width=7pt, fill=colorRandom, draw=colorRandom!70!black, bar shift=-8.5pt, forget plot]
        coordinates {(3B,63.6) (7B,80.4)};
    \addplot[ybar, bar width=7pt, fill=colorSignal, draw=colorSignal!70!black, bar shift=0.0pt, forget plot]
        coordinates {(3B,71.7) (7B,86.7)};
    \addplot[ybar, bar width=7pt, fill=colorFisher, draw=colorFisher!70!black, bar shift=8.5pt, forget plot]
        coordinates {(3B,74.4) (7B,86.8)};
    \addplot[ybar, bar width=7pt, fill=colorFullLoRA, draw=colorFullLoRA!70!black, bar shift=17.0pt, forget plot]
        coordinates {(3B,73.8) (7B,87.9)};

    \nextgroupplot[
        title={GSM8K $\to$ MATH-500},
        ylabel={},
        ymin=25, ymax=75,
        ymajorgrids=true,
        ytick={25, 35, 45, 55, 65, 75},
        symbolic x coords={3B, 7B},
        xtick=data,
        enlarge x limits=0.7,
        legend to name=cross_task_legend,
    ]
    \addplot[ybar, bar width=7pt, fill=colorBase, draw=colorBase!70!black, bar shift=-17.0pt, forget plot]
        coordinates {(3B,49.0) (7B,53.6)};
    \addplot[ybar, bar width=7pt, fill=colorRandom, draw=colorRandom!70!black, bar shift=-8.5pt, forget plot]
        coordinates {(3B,49.6) (7B,53.2)};
    \addplot[ybar, bar width=7pt, fill=colorSignal, draw=colorSignal!70!black, bar shift=0.0pt, forget plot]
        coordinates {(3B,53.8) (7B,57.6)};
    \addplot[ybar, bar width=7pt, fill=colorFisher, draw=colorFisher!70!black, bar shift=8.5pt, forget plot]
        coordinates {(3B,51.4) (7B,52.6)};
    \addplot[ybar, bar width=7pt, fill=colorFullLoRA, draw=colorFullLoRA!70!black, bar shift=17.0pt, forget plot]
        coordinates {(3B,34.2) (7B,64.8)};
    \addlegendimage{area legend, fill=colorBase, draw=colorBase!70!black}
    \addlegendentry{Base}
    \addlegendimage{area legend, fill=colorRandom, draw=colorRandom!70!black}
    \addlegendentry{Random}
    \addlegendimage{area legend, fill=colorSignal, draw=colorSignal!70!black}
    \addlegendentry{$\hat{S}$}
    \addlegendimage{area legend, fill=colorFisher, draw=colorFisher!70!black}
    \addlegendentry{$\hat{F}$}
    \addlegendimage{area legend, fill=colorFullLoRA, draw=colorFullLoRA!70!black}
    \addlegendentry{Full LoRA}

\end{groupplot}
\node at ($(group c1r1.south)!0.5!(group c2r1.south) + (0,-1.0cm)$)
    {\ref*{cross_task_legend}};
\end{tikzpicture}
\caption{\textbf{Cross-task greedy transfer} (Qwen2.5-3B/7B).  Gaps indicate methods not evaluated at that scale.  MATH$\to$GSM8K: circuits match Full LoRA to within 1.1pp.  GSM8K$\to$MATH-500: at 3B, Full LoRA degrades to 34.2\% ($-$14.8pp below base) while $\hat{S}$ at 10K improves to 53.8\%; at 7B, Full LoRA transfers well (64.8\%, +11.2pp).}
\label{fig:cross_task}
\end{figure}
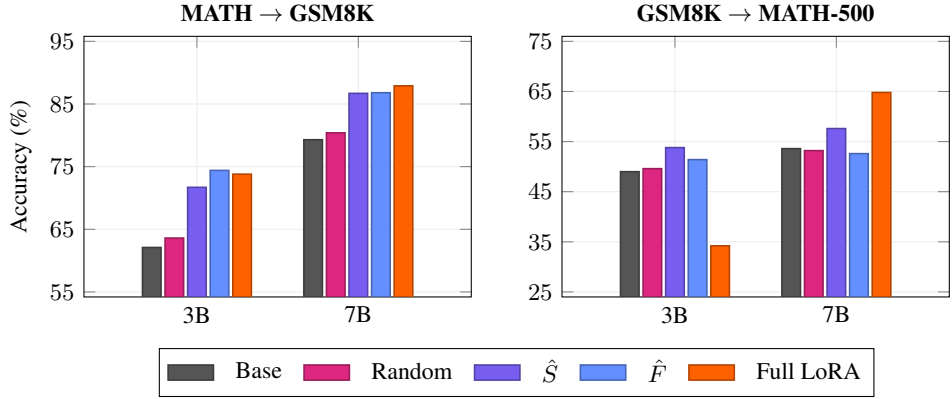

MATH$\to$GSM8K transfer is robust at both scales.  GSM8K$\to$MATH is asymmetric: at 3B, Full LoRA catastrophically degrades while $\hat{S}$ is the only method to improve over base in both directions.

\subsubsection{GSM8K-Trained Models}
\label{app:ood_gsm}

GSM8K-trained models evaluated OOD on AMC and AIME 2024.  AIME 2025 is omitted: near-zero at 3B (best: Row $\hat{F}$ 26.7\% pass@64, matching base) and weak at 7B (best: Full LoRA 43.3\%).  At 3B, different methods were tested ($\hat{S}$ 10K, Row $\hat{F}$ 1.6K); at 7B, Random 50K and $\hat{F}$ 50K.

\begin{figure}[h]
\centering
% Auto-generated by pgf_ood_gsm.py — do not edit by hand.
% GSM8K-trained models evaluated OOD. pass@k curves.
% AIME 2025 omitted (near-zero for all methods).
\begin{tikzpicture}
\begin{groupplot}[
    tinylora,
    group style={
        group size=2 by 2,
        horizontal sep=0.8cm,
        vertical sep=0.7cm,
    },
    width=0.38\textwidth,
    height=0.25\textwidth,
    legend columns=5,
    legend style={draw=black, font=\small, column sep=4pt},
]

    \nextgroupplot[
        title={AMC 2023},
        ylabel={pass@$k$ (\%)},
        ylabel style={align=center},
        xlabel={},
        xmode=log,
        xmin=0.8, xmax=80,
        ymin=0, ymax=105,
        xtick={1, 4, 8, 16, 32, 64},
        xticklabels={},
    ]
    \addplot[colorBase, dashed, thin, mark=none, mark size=2pt, forget plot]
        coordinates {(1,20.0) (4,47.5) (8,57.5) (16,67.5) (32,80.0) (64,85.0)};
    \addplot[colorBase, dashed, thin, mark=none, forget plot]
        coordinates {(4,20.0) (8,27.5) (16,35.0) (64,42.5)};
    \addplot[colorFullLoRA, thick, mark=diamond*, mark size=2pt, forget plot]
        coordinates {(1,10.0) (4,30.0) (8,40.0) (16,62.5) (32,75.0) (64,82.5)};
    \addplot[colorFullLoRA, dashed, thin, mark=none, forget plot]
        coordinates {(4,12.5) (8,20.0) (16,17.5) (64,27.5)};
    \addplot[colorSignal, thick, mark=triangle*, mark size=2pt, forget plot]
        coordinates {(1,32.5) (4,50.0) (8,55.0) (16,67.5) (32,85.0) (64,97.5)};
    \addplot[colorSignal, dashed, thin, mark=none, forget plot]
        coordinates {(4,42.5) (8,45.0) (16,45.0) (64,47.5)};
    \addplot[colorFisher, thick, dashed, mark=square, mark size=2pt, forget plot]
        coordinates {(1,27.5) (4,52.5) (8,57.5) (16,62.5) (32,75.0) (64,87.5)};
    \addplot[colorFisher, dashed, thin, mark=none, forget plot]
        coordinates {(4,32.5) (8,37.5) (16,50.0) (64,47.5)};

    \nextgroupplot[
        title={AIME 2024},
        ylabel={},
        xlabel={},
        xmode=log,
        xmin=0.8, xmax=80,
        ymin=0, ymax=50,
        xtick={1, 4, 8, 16, 32, 64},
        xticklabels={},
    ]
    \addplot[colorBase, dashed, thin, mark=none, mark size=2pt, forget plot]
        coordinates {(1,0.0) (4,3.3) (8,10.0) (16,10.0) (32,20.0) (64,23.3)};
    \addplot[colorFullLoRA, thick, mark=diamond*, mark size=2pt, forget plot]
        coordinates {(1,3.3) (4,3.3) (8,3.3) (16,6.7) (32,10.0) (64,16.7)};
    \addplot[colorSignal, thick, mark=triangle*, mark size=2pt, forget plot]
        coordinates {(1,6.7) (4,16.7) (8,20.0) (16,20.0) (32,23.3) (64,30.0)};
    \addplot[colorFisher, thick, dashed, mark=square, mark size=2pt, forget plot]
        coordinates {(1,6.7) (4,16.7) (8,20.0) (16,20.0) (32,23.3) (64,26.7)};

    \nextgroupplot[
        title={},
        ylabel={pass@$k$ (\%)},
        ylabel style={align=center},
        xlabel={$k$},
        xmode=log,
        xmin=0.8, xmax=80,
        ymin=0, ymax=105,
        xtick={1, 4, 8, 16, 32, 64},
        xticklabels={1, 4, 8, 16, 32, 64},
    ]
    \addplot[colorBase, dashed, thin, mark=none, mark size=2pt, forget plot]
        coordinates {(1,27.5) (4,52.5) (8,70.0) (16,85.0) (32,92.5) (64,95.0)};
    \addplot[colorBase, dashed, thin, mark=none, forget plot]
        coordinates {(4,37.5) (8,47.5) (16,45.0) (32,50.0) (64,55.0)};
    \addplot[colorRandom, thick, mark=o, mark size=2pt, forget plot]
        coordinates {(1,20.0) (4,55.0) (8,75.0) (16,82.5) (32,87.5) (64,95.0)};
    \addplot[colorRandom, dashed, thin, mark=none, forget plot]
        coordinates {(4,27.5) (8,42.5) (16,45.0) (32,55.0) (64,52.5)};
    \addplot[colorFisher, thick, mark=square*, mark size=2pt, forget plot]
        coordinates {(1,25.0) (4,45.0) (8,60.0) (16,75.0) (32,90.0) (64,95.0)};
    \addplot[colorFisher, dashed, thin, mark=none, forget plot]
        coordinates {(4,22.5) (8,35.0) (16,47.5) (32,52.5) (64,52.5)};
    \addplot[colorFullLoRA, thick, mark=diamond*, mark size=2pt, forget plot]
        coordinates {(1,30.0) (4,60.0) (8,72.5) (16,82.5) (32,87.5) (64,90.0)};
    \addplot[colorFullLoRA, dashed, thin, mark=none, forget plot]
        coordinates {(4,45.0) (8,57.5) (16,55.0) (32,55.0) (64,60.0)};

    \nextgroupplot[
        title={},
        ylabel={},
        xlabel={$k$},
        xmode=log,
        xmin=0.8, xmax=80,
        ymin=0, ymax=50,
        xtick={1, 4, 8, 16, 32, 64},
        xticklabels={1, 4, 8, 16, 32, 64},
        legend to name=ood_gsm_legend,
    ]
    \addplot[colorBase, dashed, thin, mark=none, mark size=2pt, forget plot]
        coordinates {(1,0.0) (4,13.3) (8,16.7) (16,23.3) (32,33.3) (64,40.0)};
    \addplot[colorRandom, thick, mark=o, mark size=2pt, forget plot]
        coordinates {(1,0.0) (4,16.7) (8,16.7) (16,20.0) (32,26.7) (64,36.7)};
    \addplot[colorFisher, thick, mark=square*, mark size=2pt, forget plot]
        coordinates {(1,3.3) (4,10.0) (8,16.7) (16,16.7) (32,20.0) (64,30.0)};
    \addplot[colorFullLoRA, thick, mark=diamond*, mark size=2pt, forget plot]
        coordinates {(1,3.3) (4,23.3) (8,30.0) (16,33.3) (32,36.7) (64,46.7)};
    \addplot[colorBase, dashed, thin, mark=none, mark size=2pt]
        coordinates {(1,-999)};
    \addlegendentry{Base}
    \addplot[colorRandom, thick, mark=o, mark size=2pt]
        coordinates {(1,-999)};
    \addlegendentry{Random}
    \addplot[colorSignal, thick, mark=triangle*, mark size=2pt]
        coordinates {(1,-999)};
    \addlegendentry{$\hat{S}$}
    \addplot[colorFisher, thick, mark=square*, mark size=2pt]
        coordinates {(1,-999)};
    \addlegendentry{$\hat{F}$}
    \addplot[colorFullLoRA, thick, mark=diamond*, mark size=2pt]
        coordinates {(1,-999)};
    \addlegendentry{Full LoRA}

\end{groupplot}
\node[anchor=east, font=\small\bfseries, rotate=90]
    at ($(group c1r1.west) + (-1.3cm,0)$) {3B};
\node[anchor=east, font=\small\bfseries, rotate=90]
    at ($(group c1r2.west) + (-1.3cm,0)$) {7B};
\node at ($(group c1r2.south)!0.5!(group c2r2.south) + (0,-1.0cm)$)
    {\ref*{ood_gsm_legend}};
\node[font=\scriptsize] at ($(group c1r2.south)!0.5!(group c2r2.south) + (0,-1.5cm)$)
    {Solid: pass@$k$ \qquad Dashed: maj@$k$ (AMC only)};
\end{tikzpicture}
\caption{\textbf{GSM8K-trained pass@$k$ and maj@$k$} (Qwen2.5-3B/7B).  At 3B, Full LoRA degrades below base on AMC while $\hat{S}$ at 10K reaches 97.5\% pass@64.  At 7B, Full LoRA transfers well and dominates on AIME 2024.}
\label{fig:ood_gsm_passk}
\end{figure}
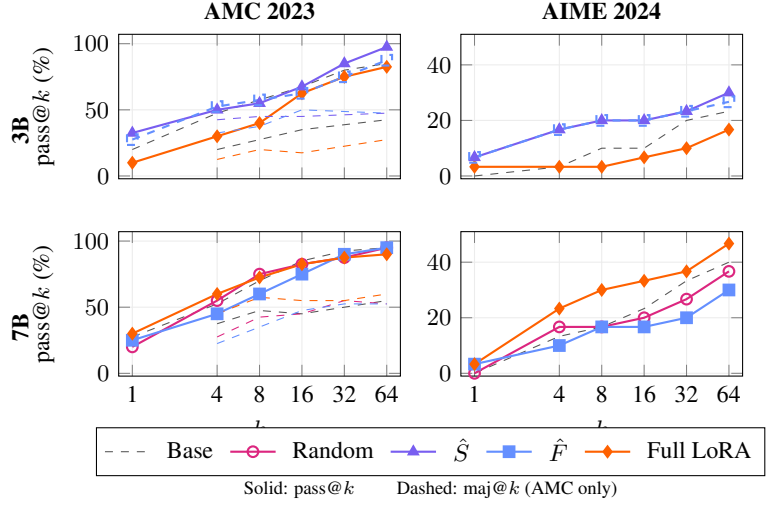

At 3B, Full LoRA degrades OOD---AMC pass@1 drops to 10\% (vs.\ 20\% base)---while $\hat{S}$ at 10K preserves and improves OOD capability (32.5\% pass@1, 97.5\% pass@64).  At 7B, Full LoRA does \emph{not} degrade and dominates on AIME, consistent with the cross-task pattern (\Cref{fig:cross_task}): sparse placement helps most when the model is capacity-constrained or the training task is narrow.

\paragraph{Why do circuits generalize?}  Circuits discovered on different math datasets (GSM8K, MATH, NuminaMath) share ${\sim}$30\% of their top-$k$ elements on Qwen2.5-1.5B, roughly 12$\times$ above the 2.4\% chance overlap; same-domain pairs share ${\sim}$40\%.  This suggests circuit discovery identifies a core subnetwork the model consistently wants to update for a given task family, constraining updates to shared directions and providing natural regularization that full LoRA lacks.

\clearpage
\section{Ablations}
\label{app:ablations}
\localtableofcontents

\subsection{Training A+B vs.\ B-Only}
\label{app:a_vs_ab}

Our method freezes $\mathbf{A}$ and selects sparse elements of $\mathbf{B}$ to train.  A natural question is whether jointly training sparse elements of both $\mathbf{A}$ and $\mathbf{B}$ at the same total budget would perform better.  We compare the two strategies on Qwen3-0.6B across budgets from 1K to 500K parameters, using both perplexity evaluation (5 seeds) and GSM8K downstream accuracy as well as Qwen2.5-1.5b on Instruction tuning.

\begin{figure}[h]
\centering
% Auto-generated by pgf_b_vs_ab.py — do not edit by hand.
% Usage: \input{figures/b_vs_ab.tex}
%   Requires: tinylora_preamble.tex, pgfplots, groupplots, calc
\begin{tikzpicture}
\begin{groupplot}[
    tinylora,
    group style={
        group size=3 by 1,
        horizontal sep=1.2cm,
    },
    width=0.35\textwidth,
    height=0.25\textwidth,
    legend columns=4,
    legend style={draw=black, font=\small, column sep=8pt},
]

    \nextgroupplot[
        title={(a) Qwen3-0.6B},
        ylabel={Perplexity $\downarrow$},
        xlabel={Budget $k$},
        xmode=log,
        xmin=700, xmax=1200000,
        xtick={1000, 10000, 100000, 500000},
        xticklabels={1K, 10K, 100K, 500K},
        ymin=4.0, ymax=9.5,
    ]
    \addplot[colorFisher, mark=square*, mark size=2pt, thick, forget plot]
        coordinates {(1000,5.59) (5000,5.23) (10000,4.97) (100000,4.99) (500000,4.57)};
    \addplot[colorRandom, mark=o, mark size=2pt, thick, forget plot]
        coordinates {(1000,8.81) (5000,8.58) (10000,8.29) (100000,6.95) (500000,5.68)};

    \nextgroupplot[
        title={(b) Qwen3-0.6B},
        ylabel={Token Accuracy (\%)},
        xlabel={Budget $k$},
        xmode=log,
        xmin=700, xmax=1200000,
        xtick={1000, 10000, 100000, 500000},
        xticklabels={1K, 10K, 100K, 500K},
        ymin=55, ymax=66,
    ]
    \addplot[colorFisher, mark=square*, mark size=2pt, thick, forget plot]
        coordinates {(1000,60.9) (5000,62.0) (10000,62.8) (100000,62.7) (500000,64.2)};
    \addplot[colorRandom, mark=o, mark size=2pt, thick, forget plot]
        coordinates {(1000,56.5) (5000,56.6) (10000,56.8) (100000,58.4) (500000,60.6)};

    \nextgroupplot[
        title={(c) Qwen2.5-1.5B},
        ylabel={7-bench Avg (\%)},
        xlabel={Budget $k$},
        xmode=log,
        xmin=700, xmax=150000,
        ymin=51, ymax=64.5,
        xtick={1000, 10000, 100000},
        xticklabels={1K, 10K, 100K},
        legend to name=bvsab_legend,
    ]
    \addplot[colorBase, dashed, thin]
        coordinates {(700,53.0) (150000,53.0)};
    \addlegendentry{Base}
    \node[anchor=south west, font=\scriptsize, text=colorBase]
        at (axis cs:900,53.0) {Base};
    \addplot[colorFullLoRA, dashed, thin]
        coordinates {(700,62.5) (150000,62.5)};
    \addlegendentry{Full LoRA}
    \node[anchor=north west, font=\scriptsize, text=colorFullLoRA]
        at (axis cs:900,62.5) {Full LoRA};
    \addplot[colorFisher, mark=square*, mark size=2pt, thick]
        coordinates {(1000,56.1) (5000,59.4) (10000,62.1) (100000,62.9)};
    \addlegendentry{$\mathbf{B}$-only}
    \addplot[colorRandom, mark=o, mark size=2pt, thick]
        coordinates {(1000,52.8) (5000,53.7) (10000,58.2) (100000,62.4)};
    \addlegendentry{$\mathbf{A}{+}\mathbf{B}$}

\end{groupplot}
\node at ($(group c1r1.south)!0.5!(group c3r1.south) + (0,-1.2cm)$)
    {\ref*{bvsab_legend}};
\end{tikzpicture}
\caption{\textbf{$\mathbf{B}$-only vs.\ $\mathbf{A}{+}\mathbf{B}$ scaling curves} at matched parameter budget.  (a,\,b)~Qwen3-0.6B perplexity and token accuracy; (c)~Qwen2.5-1.5B 7-benchmark SFT average.  $\mathbf{B}$-only (blue) dominates $\mathbf{A}{+}\mathbf{B}$ (magenta) at every budget, with the gap largest at small~$k$.  Dashed lines in~(c) show base model and full LoRA references.}
\label{fig:b_vs_ab}
\end{figure}
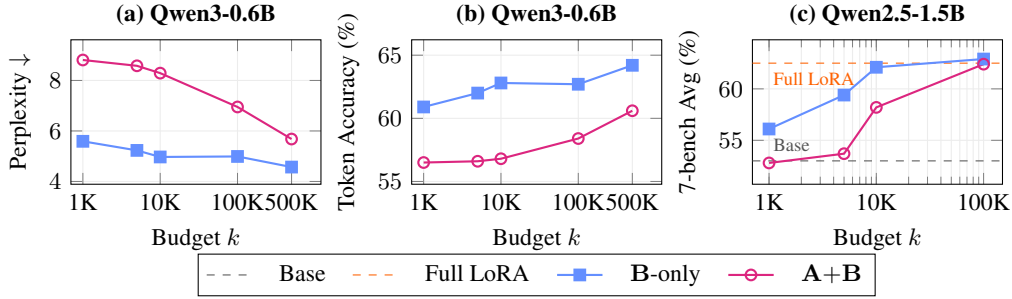

\Cref{fig:b_vs_ab} visualizes the same comparison as scaling curves.

B-only dominates A+B at every budget.  At $k{=}1$K the gap is $3.2\times$ in perplexity (5.59 vs.\ 8.81); even at $k{=}500$K it remains substantial (4.57 vs.\ 5.68).  The downstream gap on GSM8K is 4.7 points (42.5\% vs.\ 37.8\%) on 0.6B and 51.9 points (61.2\% vs.\ 9.3\%) on the larger Qwen2.5-1.5B at $k{=}1{,}600$.

The explanation is straightforward.  Since $\mathbf{B} = 0$ at initialization, the adapter output is $\mathbf{B}\mathbf{A} \mathbf{x} = 0$ regardless of $\mathbf{A}$.  Consequently, $\nabla_{\mathbf{A}} \mathcal{L} = 0$ at the start of training: changing $\mathbf{A}$ has no effect on the model output when $\mathbf{B}$ is zero.  Any parameter budget allocated to $\mathbf{A}$ is wasted until $\mathbf{B}$ moves away from zero, and even then $\mathbf{A}$ gradients scale with $\|\mathbf{B}\| \approx 0$.  Under a fixed budget, every slot given to $\mathbf{A}$ is a slot not given to $\mathbf{B}$, directly reducing the model's capacity to learn.

This validates our design choice of fixing $\mathbf{A}$ and selecting within $\mathbf{B}$. Our finding is consistent with LoRA-FA \citep{zhang2023lorafa}, which showed that freezing $\mathbf{A}$ after initialization preserves or improves performance in the full-rank setting.

%% ============================================================
\subsection{Monte Carlo Convergence}
\label{app:mc_convergence}

\Cref{tab:mc_convergence} reports top-10K element overlap with the $N{=}100$ reference circuit across three Qwen2.5 model sizes on GSM8K.  Random 10K-element subsets overlap by $< 0.1\%$ in expectation.  Convergence is consistent across models and improves with scale: the 7B model reaches 99\% overlap at $N{=}50$.  We observe the same pattern on MATH (Qwen2.5-1.5B): $\hat{S}$ overlap is 94.0\%, 95.8\%, 97.8\% at $N{=}10, 25, 50$ respectively.

\begin{figure}[t]
\centering
% Auto-generated by pgf_stability_validation.py — do not edit by hand.
% Usage: \input{figures/stability_validation.tex}
%   Requires: pgfplots, groupplots
\definecolor{plotBlue}{HTML}{4878CF}
\definecolor{plotRed}{HTML}{E8553D}
\definecolor{plotOrange}{HTML}{D65F28}
\definecolor{plotGray}{HTML}{999999}
\begin{tikzpicture}
\begin{groupplot}[
    group style={
        group size=2 by 1,
        horizontal sep=1.8cm,
    },
    width=0.50\textwidth,
    height=0.4\textwidth,
    every axis plot/.append style={semithick},
    grid style={gray!15},
    tick label style={font=\small},
    label style={font=\small},
    title style={font=\small\bfseries},
    legend style={font=\scriptsize},
]

    \nextgroupplot[
        title={(a) Monte Carlo convergence},
        xlabel={$N$ (discovery examples)},
        ylabel={Top-10K overlap with $N{=}100$},
        xtick={10, 25, 50, 100},
        ymin=0.90, ymax=1.01,
        ytick={0.90, 0.92, 0.94, 0.96, 0.98, 1.00},
        yticklabel={\pgfmathparse{\tick*100}%
            \pgfmathprintnumber[fixed, precision=0]{\pgfmathresult}\%},
        ymajorgrids=true,
        legend pos=south east,
    ]
        \addplot[plotBlue, mark=o, mark size=2.5pt, thick]
            coordinates {(10,0.948100) (25,0.966100) (50,0.981800) (100,1.000000)};
        \addlegendentry{$\hat{S}$ (grad.\ magnitude)}
        \addplot[plotRed, mark=square*, mark size=2pt, thick, dashed]
            coordinates {(10,0.948300) (25,0.964200) (50,0.979000) (100,1.000000)};
        \addlegendentry{$\hat{F}$ (Fisher)}
        \addplot[plotGray, dashed, thin, forget plot]
            coordinates {(10,1.0) (100,1.0)};
        \node[plotBlue, font=\scriptsize\bfseries, anchor=south west]
            at (axis cs:10,0.9481) [xshift=-10pt, yshift=8pt] {94.8\%};
        \node[plotRed, font=\scriptsize\bfseries, anchor=north west]
            at (axis cs:10,0.9483) [xshift=0pt, yshift=-2pt] {94.8\%};
        \node[plotBlue, font=\scriptsize\bfseries, anchor=south west]
            at (axis cs:50,0.9818) [xshift=2pt, yshift=5pt] {98.2\%};
        \node[plotRed, font=\scriptsize\bfseries, anchor=north west]
            at (axis cs:50,0.979) [xshift=2pt, yshift=-2pt] {97.9\%};

    \nextgroupplot[
        title={(b) Sensitivity to $A$ initialization},
        xlabel={Perturbation $\epsilon$ (relative to $\|A\|_F$)},
        ylabel={Overlap with unperturbed circuit},
        ymin=0, ymax=1.05,
        ytick={0, 0.20, 0.40, 0.60, 0.80, 1.00},
        yticklabel={\pgfmathparse{\tick*100}%
            \pgfmathprintnumber[fixed, precision=0]{\pgfmathresult}\%},
        ymajorgrids=true,
        legend pos=south west,
    ]
        \addplot[plotBlue, mark=o, mark size=2pt, thick]
            coordinates {(0.0,1.000000) (0.01,0.989000) (0.05,0.961000) (0.1,0.924000) (0.25,0.813000) (0.5,0.633000) (1.0,0.423000) (2.0,0.273000)};
        \addlegendentry{Top-1K}
        \addplot[plotOrange, mark=square*, mark size=2pt, thick]
            coordinates {(0.0,1.000000) (0.01,0.991000) (0.05,0.962400) (0.1,0.923600) (0.25,0.816600) (0.5,0.663000) (1.0,0.466000) (2.0,0.336600)};
        \addlegendentry{Top-5K}
        \addplot[plotRed, mark=diamond*, mark size=2pt, thick]
            coordinates {(0.0,1.000000) (0.01,0.991500) (0.05,0.963000) (0.1,0.927700) (0.25,0.823300) (0.5,0.672900) (1.0,0.482000) (2.0,0.353800)};
        \addlegendentry{Top-10K}
        \addplot[plotGray, dashed, thin, forget plot]
            coordinates {(0,1.0) (2.0,1.0)};
        \node[plotRed, font=\scriptsize\bfseries, anchor=west]
            at (axis cs:0.1,0.928) [xshift=4pt] {92\%};

\end{groupplot}
\end{tikzpicture}
\caption{\textbf{Circuit discovery is stable.}  (a)~Top-10K overlap with the $N{=}100$ reference circuit as a function of discovery examples $N$.  Both $\hat{S}$ and $\hat{F}$ reach 98\% overlap at $N{=}50$.  Random circuits overlap by $<0.1\%$.  (b)~$\hat{S}$-circuit overlap under perturbation of $\mathbf{A}$.  The circuit degrades smoothly; at 10\% perturbation, 92\% of elements are unchanged.}
\label{fig:stability}
\end{figure}
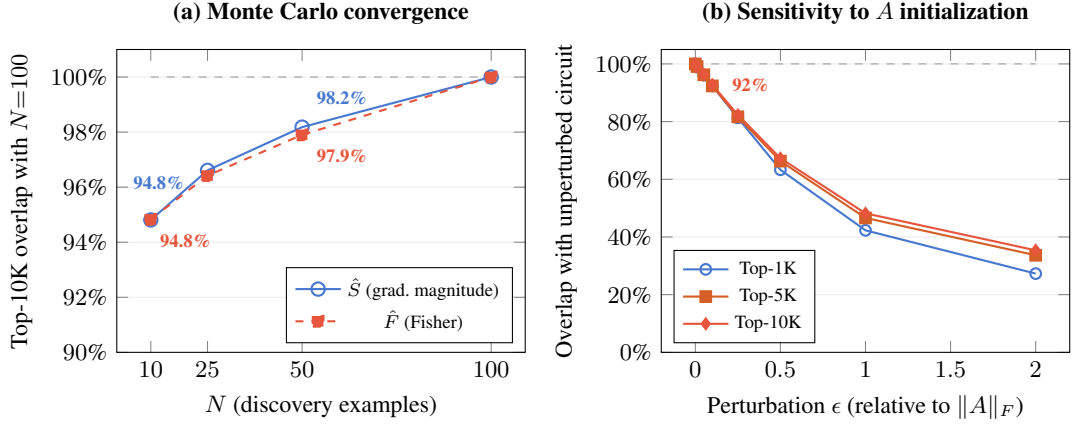

\begin{table}[h]
\centering
\caption{\textbf{Monte Carlo convergence.}  Top-10K overlap (\%) with the $N{=}100$ circuit (GSM8K).}
\label{tab:mc_convergence}
\vspace{4pt}
\small
\setlength{\tabcolsep}{5pt}
\begin{tabular}{l cc cc cc}
\toprule
& \multicolumn{2}{c}{\textbf{Qwen2.5-1.5B}} & \multicolumn{2}{c}{\textbf{Qwen2.5-3B}} & \multicolumn{2}{c}{\textbf{Qwen2.5-7B}} \\
\cmidrule(lr){2-3} \cmidrule(lr){4-5} \cmidrule(lr){6-7}
\textbf{$N$} & $\hat{S}$ & $\hat{F}$ & $\hat{S}$ & $\hat{F}$ & $\hat{S}$ & $\hat{F}$ \\
\midrule
10  & 94.8 & 94.8 & 92.3 & 90.4 & 96.8 & 95.9 \\
25  & 96.6 & 96.4 & 95.4 & 94.1 & 98.4 & 98.0 \\
50  & 98.2 & 97.9 & 97.6 & 97.0 & 99.1 & 98.8 \\
100 & 100.0 & 100.0 & 100.0 & 100.0 & 100.0 & 100.0 \\
\bottomrule
\end{tabular}
\end{table}

%% ============================================================
\subsection{Sensitivity to $\mathbf{A}$ Initialization}
\label{app:a_perturbation}

We test whether the discovered circuit depends on the specific random initialization of $\mathbf{A}$.  Starting from the same base model, we perturb $\mathbf{A}' = \mathbf{A} + \varepsilon \|\mathbf{A}\|_F \cdot \boldsymbol{\Delta} / \|\boldsymbol{\Delta}\|_F$ (where $\boldsymbol{\Delta} \sim \mathcal{N}(0, \mathbf{I})$) and re-run circuit discovery.  \Cref{tab:a_perturbation} reports top-10K overlap with the unperturbed circuit.

\begin{table}[h]
\centering
\caption{\textbf{Sensitivity to $\mathbf{A}$ perturbation.}  Top-10K overlap (\%) with the unperturbed ($\varepsilon{=}0$) circuit across three model sizes (GSM8K, $N{=}50$ batches).}
\label{tab:a_perturbation}
\vspace{4pt}
\small
\setlength{\tabcolsep}{6pt}
\begin{tabular}{l ccc}
\toprule
\textbf{$\varepsilon$} & \textbf{1.5B} & \textbf{3B} & \textbf{7B} \\
\midrule
0.01 & 98.9 & 99.3 & 99.4 \\
0.05 & 96.1 & 96.7 & 96.8 \\
0.1  & 92.4 & 93.5 & 93.7 \\
0.25 & 82.3 & 84.0 & 84.0 \\
0.5  & 63.3 & 70.4 & 68.2 \\
1.0  & 42.3 & 53.8 & 49.0 \\
2.0  & 27.3 & 42.0 & 42.3 \\
\bottomrule
\end{tabular}
\end{table}

At small perturbations ($\varepsilon \le 0.1$), over 92\% of the top-10K elements are preserved across all model sizes.  This confirms that our circuit reflects the base model's structure rather than the random $\mathbf{A}$ initialization.  Degradation at larger $\varepsilon$ is expected: $\mathbf{A}$ scales the gradient signal reaching $\mathbf{B}$, so large changes to $\mathbf{A}$ change which $\mathbf{B}$ elements receive the strongest gradients.

%% ============================================================
\subsection{Circuit Discovery Cost}
\label{app:discovery_cost}

We benchmark the cost of circuit discovery with $N{=}50$ forward-backward passes, batch size 1, max length 1024, bfloat16) across Qwen2.5 model sizes on a single NVIDIA RTX PRO 6000 GPU (96\,GB).  No optimizer state is required---only the accumulated gradient tensor (float32) is stored on CPU.

\begin{table}[h]
\centering
\caption{\textbf{Circuit discovery cost} ($N{=}50$, single GPU). Discovery time excludes model loading. Peak VRAM includes model weights, LoRA adapters, activations, and gradients. Grad tensor is CPU memory for accumulated scores.}
\label{tab:discovery_cost}
\vspace{4pt}
\small
\setlength{\tabcolsep}{4pt}
\begin{tabular}{l rr rrrr}
\toprule
\textbf{Model} & \textbf{Params} & \textbf{LoRA-$\mathbf{B}$} & \textbf{Time (s)} & \textbf{Peak VRAM} & \textbf{Grad Tensor} & \textbf{\% SFT / GRPO} \\
\midrule
Qwen2.5-0.5B & 494M & 9.7M & 4.9 & 2.6\,GB & 39\,MB  & 0.49 / 0.18 \\
Qwen2.5-1.5B & 1.5B & 20.6M & 5.5 & 5.5\,GB & 83\,MB  & 0.49 / 0.18 \\
Qwen2.5-3B   & 3.1B & 33.0M & 7.3 & 9.7\,GB & 132\,MB & 0.49 / 0.18 \\
Qwen2.5-7B   & 7.6B & 44.5M & 8.5 & 19.6\,GB & 178\,MB & 0.49 / 0.18 \\
\bottomrule
\end{tabular}
\end{table}

Discovery time scales sub-linearly with model size: 7B takes only 1.7$\times$ longer than 0.5B (8.5s vs.\ 4.9s), since kernel launch overhead dominates at smaller scales.

\paragraph{FLOP comparison with training.}  We compare theoretical FLOPs using the standard approximation (forward $\approx 2PS$, backward $\approx 4PS$).  Circuit discovery uses 50 passes with average sequence length 201 (empirical on GSM8K).  SFT uses 1{,}000 steps $\times$ gradient accumulation 4 = 4{,}000 passes at sequence length 512.  GRPO uses 1{,}000 steps $\times$ gradient accumulation 8 = 8{,}000 policy gradient passes plus generation cost (1{,}000 prompts $\times$ 8 generations $\times$ 512 tokens).

\begin{table}[h]
\centering
\caption{\textbf{Circuit discovery FLOPs relative to training.}  Discovery costs $<$0.5\% of SFT and $<$0.2\% of GRPO at all scales.}
\label{tab:discovery_flops}
\vspace{4pt}
\small
\setlength{\tabcolsep}{5pt}
\begin{tabular}{l rrr cc}
\toprule
\textbf{Model} & \textbf{Discovery} & \textbf{SFT} & \textbf{GRPO} & \textbf{\% of SFT} & \textbf{\% of GRPO} \\
\midrule
Qwen2.5-0.5B & 30.9\,TF & 6.3\,PF & 16.7\,PF & 0.49 & 0.18 \\
Qwen2.5-1.5B & 95.3\,TF & 19.4\,PF & 51.7\,PF & 0.49 & 0.18 \\
Qwen2.5-3B   & 189.8\,TF & 38.6\,PF & 102.9\,PF & 0.49 & 0.18 \\
Qwen2.5-7B   & 464.1\,TF & 94.6\,PF & 252.2\,PF & 0.49 & 0.18 \\
\bottomrule
\end{tabular}
\end{table}

The FLOP ratio is constant across model sizes because both discovery and training scale linearly with parameters.  The ratio is small because discovery uses far fewer passes (50 vs.\ 4{,}000--8{,}000) and shorter sequences (201 vs.\ 512).

\paragraph{Practical implications.}  (1)~Circuit discovery is amortizable: once computed, the same circuit can be reused across training runs with different hyperparameters, learning rates, or training durations.  (2)~No optimizer state is needed, so peak VRAM is comparable to inference with a backward pass.  (3)~The accumulated gradient tensor is small (39--178\,MB on CPU), making discovery feasible even on memory-constrained hardware where training would require offloading or sharding.

\subsection{Stability Across Seeds}
\label{app:seed_stability}
Circuit placement is stable across A-initializations: over 8 random seeds, early-layer concentration varies by $\leq3$pp (e.g., 99.0±0.2\% in layers 0–4 for Qwen-7B) and module preferences are consistent within each architecture (V+O: 83±2\% on Qwen-7B, Down: 92±5\% on Llama-3B).

  \begin{table}[h]                                                                 
  \centering                                                                      
  \caption{\textbf{Circuit placement stability across $\mathbf{A}$-seeds} ($\hat{S}$, top 10K, Alpaca).  Mean ± std over 8 random $\mathbf{A}$ initializations (4  
  seeds $\times$ 2 scoring methods).  Layer and module preferences are consistent regardless of the random projection.}               
  \label{tab:seed_stability}                                     
  \vspace{4pt}                                                       
  \small                                                             
  \setlength{\tabcolsep}{4pt}
  \begin{tabular}{l cc ccc}                                        
  \toprule                                                       
  & \multicolumn{2}{c}{\textbf{Layer concentration}} & \multicolumn{3}{c}{\textbf{Top modules (\% of circuit)}} \\                                     
  \cmidrule(lr){2-3} \cmidrule(lr){4-6}                                                
  \textbf{Model} & \textbf{Early (0--4)} & \textbf{std} & \textbf{1st} & \textbf{2nd} & \textbf{3rd} \\                               
  \midrule                                                                                 
  Qwen2.5-3B  & 71.7\% & $\pm$3.3 & V\,(35.3) & Gate\,(23.5) & Up\,(16.1) \\                                                                            
  Qwen2.5-7B  & 98.9\% & $\pm$0.2 & V\,(53.9) & O\,(29.5) & Gate\,(8.3) \\                                                                          
  Llama-3.2-3B & 97.4\% & $\pm$0.6 & Down\,(91.5) & V\,(6.6) & Up\,(0.9) \\                                                                                        
  Llama-3.1-8B & 96.8\% & $\pm$1.8 & V\,(51.3) & Down\,(46.7) & Up\,(1.3) \\                                                                                       
  \bottomrule                                                                      
  \end{tabular}                                                                        
  \end{table}

%% ============================================================
\subsection{Synthetic Experiment Details}
\label{app:synthetic}

\paragraph{Model.}  A 2-layer MLP with ReLU: $\mathbf{y} = \mathbf{W}_2 \, \text{ReLU}(\mathbf{W}_1 \mathbf{x})$, where $\mathbf{W}_1 \in \mathbb{R}^{64 \times 128}$ and $\mathbf{W}_2 \in \mathbb{R}^{32 \times 64}$.

\paragraph{LoRA adapter.}  We adapt $\mathbf{W}_1$ with a rank-16 LoRA: $\mathbf{W}_1^{\text{eff}} = \mathbf{W}_1 + \mathbf{B}\mathbf{A}$, where $\mathbf{A} \in \mathbb{R}^{16 \times 128}$ is frozen (Kaiming initialization) and $\mathbf{B} \in \mathbb{R}^{64 \times 16}$ is initialized to zero.  The total adapter has $|\mathbf{B}| = 1{,}024$ parameters.  A mask $\mathbf{m} \in \{0,1\}^{64 \times 16}$ selects $k$ elements for training.  Full LoRA trains both $\mathbf{A}$ and $\mathbf{B}$ jointly.

\paragraph{Targets.}  The two panels use different targets to model the SFT/RL gradient structure.  \textbf{SFT-like:} the target network differs from the base by a rank-2 additive residual on $\mathbf{W}_1$.  Since the residual is dense and full-rank in the hidden dimension, gradients on $\mathbf{B}$ are distributed across all entries---any parameter subset captures useful signal.  \textbf{RL-like:} the target is a near-sparse $\mathbf{B}^*$ where only 5\% of entries are large (0.3 std) and the rest are near-zero (0.01 std).  Gradients concentrate on the few active entries, modeling the concentrated gradient structure we observe under GRPO (\Cref{sec:signal_retention}).

\paragraph{Circuit discovery.}  We run 100 forward-backward passes at initialization (batch size 128) and select the top-$k$ elements of $\mathbf{B}$ by $|\nabla_{\mathbf{B}} \mathcal{L}|$---the same procedure as for LLMs (\Cref{sec:circuit_discovery}).

\paragraph{Training.}  \textbf{SFT-like:} Adam, batch size 128, learning rate $5 \times 10^{-2}$, 5{,}000 steps, clean MSE gradients.  \textbf{RL-like:} SGD with additive Gaussian noise ($\sigma = 0.005$) and gradient clipping (norm $\leq 1.0$), batch size 4, learning rate $3 \times 10^{-3}$, 15{,}000 steps.  The noise and small batch model the high-variance gradients characteristic of GRPO, where the reward signal is binary and most generations produce near-zero advantage.

\paragraph{Evaluation.}  We report MSE on a held-out test set (4{,}096 examples), normalized by the baseline MSE (no adaptation, $\mathbf{B} = 0$).  All results are averaged over 10 seeds.

%% ============================================================
\subsection{Budget Selection}
\label{app:budget_selection}

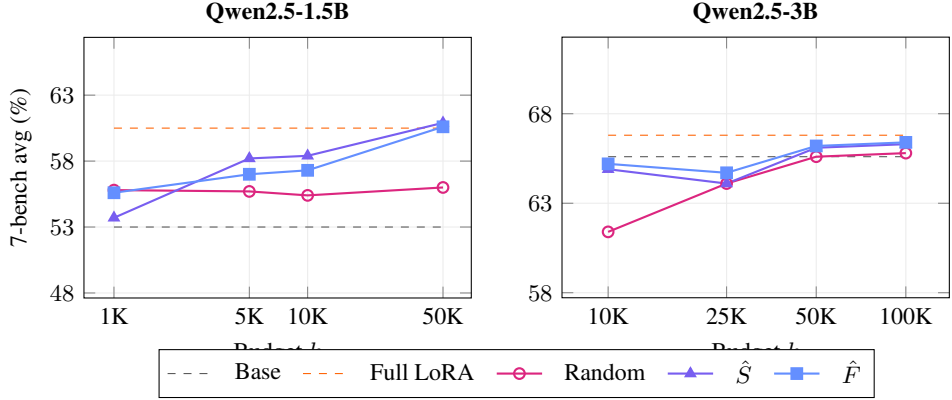
\begin{figure}[t]
\centering
% Auto-generated by pgf_budget_sweep.py — do not edit by hand.
% Budget sweep: 7-bench avg vs parameter budget k.
\begin{tikzpicture}
\begin{groupplot}[
    tinylora,
    group style={
        group size=2 by 1,
        horizontal sep=1.2cm,
    },
    width=0.48\textwidth,
    height=0.36\textwidth,
    legend columns=5,
    legend style={draw=black, font=\small, column sep=6pt},
]

    \nextgroupplot[
        title={Qwen2.5-1.5B},
        ylabel={7-bench avg (\%)},
        xlabel={Budget $k$},
        xmode=log,
        xmin=700.0, xmax=70000.0,
        ymin=48, ymax=67,
        ymajorgrids=true,
        ytick={48, 53, 58, 63},
        xtick={1000, 5000, 10000, 50000},
        xticklabels={1K, 5K, 10K, 50K},
    ]
    \addplot[colorBase, dashed, thin, forget plot]
        coordinates {(1000,53.0) (50000,53.0)};
    \addplot[colorFullLoRA, dashed, thin, forget plot]
        coordinates {(1000,60.5) (50000,60.5)};
    \addplot[colorRandom, thick, mark=o, mark size=2pt, forget plot]
        coordinates {(1000,55.8) (5000,55.7) (10000,55.4) (50000,56.0)};
    \addplot[colorSignal, thick, mark=triangle*, mark size=2pt, forget plot]
        coordinates {(1000,53.7) (5000,58.2) (10000,58.4) (50000,60.9)};
    \addplot[colorFisher, thick, mark=square*, mark size=2pt, forget plot]
        coordinates {(1000,55.6) (5000,57.0) (10000,57.3) (50000,60.6)};

    \nextgroupplot[
        title={Qwen2.5-3B},
        ylabel={},
        xlabel={Budget $k$},
        xmode=log,
        xmin=7000.0, xmax=140000.0,
        ymin=58, ymax=72,
        ymajorgrids=true,
        ytick={58, 63, 68},
        xtick={10000, 25000, 50000, 100000},
        xticklabels={10K, 25K, 50K, 100K},
        legend to name=budget_sweep_legend,
    ]
    \addplot[colorBase, dashed, thin, forget plot]
        coordinates {(10000,65.6) (100000,65.6)};
    \addplot[colorFullLoRA, dashed, thin, forget plot]
        coordinates {(10000,66.8) (100000,66.8)};
    \addplot[colorRandom, thick, mark=o, mark size=2pt, forget plot]
        coordinates {(10000,61.4) (25000,64.1) (50000,65.6) (100000,65.8)};
    \addplot[colorSignal, thick, mark=triangle*, mark size=2pt, forget plot]
        coordinates {(10000,64.9) (25000,64.1) (50000,66.1) (100000,66.3)};
    \addplot[colorFisher, thick, mark=square*, mark size=2pt, forget plot]
        coordinates {(10000,65.2) (25000,64.7) (50000,66.2) (100000,66.4)};
    \addplot[colorBase, dashed, thin]
        coordinates {(1,-999)};
    \addlegendentry{Base}
    \addplot[colorFullLoRA, dashed, thin]
        coordinates {(1,-999)};
    \addlegendentry{Full LoRA}
    \addplot[colorRandom, thick, mark=o, mark size=2pt]
        coordinates {(1,-999)};
    \addlegendentry{Random}
    \addplot[colorSignal, thick, mark=triangle*, mark size=2pt]
        coordinates {(1,-999)};
    \addlegendentry{$\hat{S}$}
    \addplot[colorFisher, thick, mark=square*, mark size=2pt]
        coordinates {(1,-999)};
    \addlegendentry{$\hat{F}$}

\end{groupplot}
\node at ($(group c1r1.south)!0.5!(group c2r1.south) + (0,-1.0cm)$)
    {\ref*{budget_sweep_legend}};
\end{tikzpicture}
\caption{\textbf{Budget sweep under Alpaca SFT} (7-benchmark avg).
Dashed lines mark base and full-LoRA references.
On 1.5B, random placement is flat across budgets (55--56\%) due to persistent MMLU collapse, while both circuits scale steadily and match full LoRA by $k{=}50$K.
On 3B, circuits lead random by 3--4pp at $k{=}10$K; all methods converge toward full LoRA by $k{=}100$K.}
\label{fig:budget_sweep}
\end{figure}

\Cref{fig:budget_sweep} shows how performance scales with the parameter budget~$k$ under SFT for two model sizes.

Informed placement is most valuable at low budgets: circuits separate from random as early as $k{=}5$K (1.5B) and $k{=}10$K (3B), while random requires $5{-}10{\times}$ more parameters to reach the same accuracy.  At moderate budgets ($k{=}100$K, 0.1--0.3\% of adapter parameters), Fisher and gradient circuits still outperform random on Llama-3.1-8B (69.8/69.7 vs.\ 69.6) and come within 0.1pp of full LoRA.

\paragraph{VLM SFT hyperparameter sensitivity.}
\label{app:vlm_sft}

At $k{=}5$K, $\hat{S}$ scores 66.9\% and $\hat{F}$ scores 66.3\%, matching or exceeding full LoRA.  Random at 5K scores 66.2\%.  The circuit methods are stable across the 5K--10K budget range, and we report the numbers for the 10K budget in the main paper as a primary reference.

\begin{table}[t]
\centering
\caption{\textbf{VLM SFT on MathVista} (Qwen2.5-VL-7B). 
Circuit methods ($\hat{S}, \hat{F}$) across parameter budgets. 
Reference: Base = 62.5, Full LoRA = 66.9.}
\label{tab:sft_vlm}
\vspace{4pt}
\small
\setlength{\tabcolsep}{5pt}
\begin{tabular}{l cccc}
\toprule
\textbf{$k$} & $\hat{S}$ & $\hat{F}$ & \textbf{Random} \\
\midrule
5K  & \textbf{66.9} & 66.3 & 66.2 \\
10K & \textbf{67.8} & 66.8 & 66.4 \\
\bottomrule
\end{tabular}
\end{table}

\paragraph{GRPO budget scaling.}  \Cref{tab:grpo_budget_all} reports circuit accuracy under GRPO across budgets from 1K to 10K on both GSM8K and MATH-500 for Qwen2.5-1.5b.  Unlike SFT, where random placement works at all budgets (\Cref{fig:budget_sweep}), under GRPO random fails uniformly while circuit methods scale smoothly with budget.

\begin{table}[t]
\centering
\caption{\textbf{GRPO budget scaling across models and tasks.} 
Circuit methods improve consistently with budget, while random placement fails. 
At extreme sparsity ($k{=}1$K), $\hat{F}$ can match full LoRA on larger models.}
\label{tab:grpo_budget_all}
\vspace{4pt}
\small
\setlength{\tabcolsep}{5pt}
\begin{tabular}{llcccc}
\toprule
\textbf{Model} & \textbf{Task} & \textbf{$k$} & $\hat{S}$ & $\hat{F}$ & \textbf{Random} \\
\midrule
\multirow{6}{*}{Qwen2.5-1.5B}
& GSM8K    & 1K  & 43.2 & \textbf{47.5} & 9.5 \\
&          & 5K  & 62.2 & \textbf{62.8} & 9.8 \\
&          & 10K & \textbf{64.2} & 63.2 & 9.5 \\
\cmidrule(l){2-6}
& MATH-500 & 1K  & \textbf{32.8} & 29.2 & 18.0 \\
&          & 5K  & \textbf{37.0} & 35.0 & 18.4 \\
&          & 10K & \textbf{43.6} & 42.2 & 18.8 \\
\midrule
Qwen2.5-3B & MATH-500 & 1K & 58.4 & \textbf{62.4} & 51.2 \\
\midrule
\multicolumn{3}{l}{\textit{Full LoRA (1.5B)}} & \multicolumn{3}{c}{GSM8K: 62.8 \quad MATH: 41.4} \\
\multicolumn{3}{l}{\textit{Full LoRA (3B)}}   & \multicolumn{3}{c}{MATH: 62.6} \\
\bottomrule
\end{tabular}
\end{table}

% \begin{table}[h]
% \centering
% \caption{\textbf{GRPO budget scaling: Qwen2.5-1.5B} (GSM8K and MATH-500).  Circuit methods scale with budget; random fails at every tier.  $\hat{F}$ leads $\hat{S}$ at 1K on GSM8K but the ranking reverses on MATH-500, consistent with task-dependent scoring at extreme sparsity.}
% \label{tab:grpo_budget_15b}
% \vspace{4pt}
% \small
% \setlength{\tabcolsep}{4pt}
% \begin{tabular}{l r cccc}
% \toprule
% \textbf{Task} & \textbf{$k$} & $\hat{S}$ & $\hat{F}$ & \textbf{Random} & \textbf{Full LoRA} \\
% \midrule
% \multirow{3}{*}{GSM8K} & 1K & 43.2 & \textbf{47.5} & 9.5 & --- \\
%  & 5K & 62.2 & \textbf{62.8} & 9.8 & --- \\
%  & 10K & \textbf{64.2} & 63.2 & 9.5 & 62.8 \\
% \midrule
% \multirow{3}{*}{MATH-500} & 1K & \textbf{32.8} & 29.2 & 18.0 & --- \\
%  & 5K & \textbf{37.0} & 35.0 & 18.4 & --- \\
%  & 10K & \textbf{43.6} & 42.2 & 18.8 & 41.4 \\
% \bottomrule
% \end{tabular}
% \end{table}

On GSM8K, $\hat{F}$ outperforms $\hat{S}$ at 1K (47.5\% vs.\ 43.2\%) but the gap closes at 5K and reverses at 10K.  On MATH-500, $\hat{S}$ leads at every budget (32.8\% vs.\ 29.2\% at 1K).  The scoring function ranking is task-dependent at extreme sparsity, consistent with the hypothesis that $\hat{S}$ (directional consistency) is more robust on harder tasks with more diffuse gradient signal.  Random placement fails uniformly at all budgets on both tasks (9.5--18.8\%), confirming that the collapse is structural.

We also show the 1K budget tier on MATH-500 for Qwen2.5-3b.  $\hat{F}$ at just 1{,}000 parameters (0.002\% of the adapter) matches full LoRA, the most extreme sparsity result in this paper. The 3B result demonstrates that circuit discovery scales across model sizes: informed placement at 0.002\% of the adapter recovers full LoRA performance under GRPO.

\paragraph{SFT budget scaling.}  \Cref{app:tab:sft_7bench:budget} scores over different rank and budget values for models trained on Bespoke-Stratos and tested on GPQADiamond. In all cases, performance is similar. The selected circuits tend to improve performance; the random circuit also improves modestly.

\begin{table}[t]
\centering
\caption{\textbf{Budget for SFT on Bespoke-Stratos (Reasoning); Accuracy on GPQADiamond.}}
\label{app:tab:sft_7bench:budget}
\vspace{4pt}
\small
\setlength{\tabcolsep}{4.5pt}
\begin{tabular}{ll ccccc}
\toprule
\textbf{Model} & \textbf{Base} & \textbf{Random} & $\hat{S}$ & $\hat{F}$ & \textbf{LoRA} & \textbf{$k$} \\
\midrule
\addlinespace[2pt]
Qwen2.5-7B-it r32 & 31.6 & \textbf{37.4$_{115\%}$} & 35.6$_{80\%}$ & 33.3$_{34\%}$ & 36.6 & 5K \\
Qwen2.5-7B-it r32 & 31.6 & 37.4 & \textbf{38.9$_{146\%}$} & \textbf{38.9$_{146\%}$} & 36.6 & 50K \\
Qwen2.5-7B-it & 31.6 & {41.1} & {43.3} & 31.1 & \textbf{50.7} & 5K \\
Qwen2.5-7B-it & 31.6 & 36.5 & {40.6} & 35.4 & 50.7 & 50K \\
\bottomrule
\end{tabular}
\end{table}

% \begin{table}[h]
% \centering
% \caption{\textbf{GRPO budget scaling: Qwen2.5-3B, MATH-500} ($k{=}1$K, rank=8).  $\hat{F}$ at 1{,}000 parameters matches full LoRA (47M parameters), a 47{,}000$\times$ compression.}
% \label{tab:grpo_budget_3b_math}
% \vspace{4pt}
% \small
% \setlength{\tabcolsep}{5pt}
% \begin{tabular}{l rc}
% \toprule
% \textbf{Method} & \textbf{$k$} & \textbf{MATH-500 (\%)}\\
% \midrule
% $\hat{F}$ & 1K & \textbf{62.4} \\
% $\hat{S}$ & 1K & 58.4\\
% Random & 1K & 51.2\\
% \midrule
% Full LoRA & 47M & 62.6 \\
% Base & --- & 50.6  \\
% \bottomrule
% \end{tabular}
% \end{table}

\subsection{Random Placement under GRPO}
\label{app:random_grpo}

\Cref{tab:grpo_budget_all,tab:grpo} shows that random placement fails under GRPO.  A natural question is that random placement might succeed with a different budget or LoRA alpha.  Here we show that the failure is robust: we swept budgets from 1K to 10K and alpha values from 32 to 4{,}096, and random placement fails uniformly across all configurations.

\begin{table}[t]
\centering
\caption{\textbf{Random placement fails under GRPO.} 
Across models, budgets, and hyperparameters, random updates produce no learning signal.}
\label{tab:random_failure}
\vspace{4pt}
\small
\setlength{\tabcolsep}{6pt}
\begin{tabular}{lccc}
\toprule
\textbf{Model} & \textbf{Setting} & \textbf{Random (\%)} & \textbf{Base (\%)} \\
\midrule
Qwen2.5-1.5B & multiple $(k,\alpha)$ & 9.3--9.8 & 44.5 \\
Qwen2.5-3B   & multiple $(k,\alpha)$ & 61.9--62.3 & 62.1 \\
\bottomrule
\end{tabular}
\end{table}

On Qwen2.5-1.5B, all five configurations produce GSM8K accuracy of 9.3--9.8\%---the model collapses to degenerate outputs regardless of whether $k{=}1$K or $k{=}10$K, and regardless of whether alpha is 32 or 4{,}096.  Increasing alpha by $128\times$ does not compensate for random placement; the failure is not a scaling issue but a structural one.  On Qwen2.5-3B, random placement is less catastrophic but equally uninformative: all configurations score within 0.4pp of the base model, indicating that GRPO produces no meaningful parameter updates through randomly selected elements.  In contrast, the circuits at the same budget reach 64.2\% (1.5B) and 70.6\% (3B), demonstrating that the difference is entirely in which parameters are trained.

\clearpage
\section{Additional Circuit Analysis}
\label{app:circuits}
\localtableofcontents

\subsection{Circuit Anatomy: Full Results}
\label{app:circuit_anatomy}

\paragraph{Circuits for Qwen and Llama Models}
\Cref{tab:circuit_composition:main} shows the placement density for the $\hat{S}$ circuit for the Qwen model.  Here we provide the placements for the other circuits for other model families. 

\begin{figure}[h]
\centering
\includegraphics[width=\linewidth]{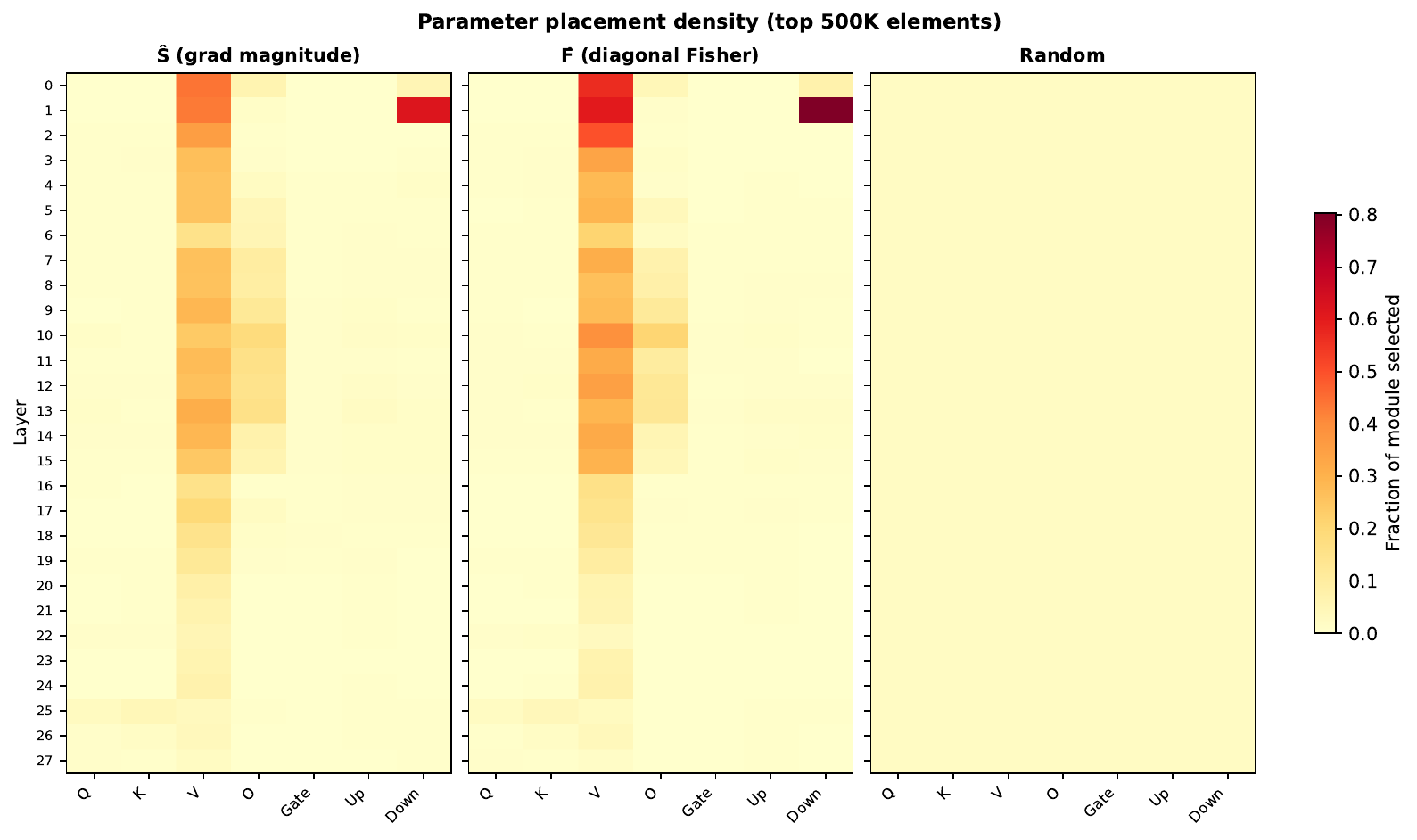}
\caption{\textbf{Llama-3.2-3B-Instruct, GSM8K} (top 500K).  At the larger budget (2\% of $B$), the circuit is more distributed but still favors V/O in the first half of layers.  The instruct model shows similar module preferences to base Qwen models.}
\label{fig:anatomy_llama3b_gsm8k}
\end{figure}

\Cref{tab:circuit_composition_fhat} shows the concentration of $\hat{F}$ for the Qwen models. The results mimic that of the $\hat{S}$ placements (\Cref{tab:circuit_composition:main}). $\hat{F}$ circuits show similar early-layer concentration (88–99\%) except on 3B MATH, where budget distributes more evenly across layers 0–23.\Cref{fig:anatomy_llama3b_gsm8k} shows the placement of the circuits for the Llama-3.2-3B-Instruct for GSM8k. The circuit is more distributed partly due to the larger budget, but like the Qwen models still favors the V/O modules. 

\begin{table}[h]
\centering
\caption{\textbf{$\hat{F}$-circuit composition} (top 10K entries, \% of circuit).  ``Param'' shows each module type's share of all $\mathbf{B}$ parameters.  V and O dominate at both scales, with O concentration increasing at 7B.  Gate and Up are consistently underrepresented despite holding the majority of parameters.}
\label{tab:circuit_composition_fhat}
\vspace{4pt}
\small
\setlength{\tabcolsep}{4.5pt}
\begin{tabular}{@{}l c cccc@{}}
\toprule
& & \multicolumn{4}{c}{\textbf{\% of circuit (top 10K)}} \\
\cmidrule(lr){3-6}
\textbf{Module} & \textbf{Param \%} & \textbf{3B MATH} & \textbf{3B GSM8K} & \textbf{7B MATH} & \textbf{7B GSM8K} \\
\midrule
Q    & \phantom{0}7.1 & \phantom{0}3.6 & \phantom{0}0.4 & \phantom{0}0.9 & \phantom{0}0.2 \\
K    & \phantom{0}0.9 & \phantom{0}3.1 & \phantom{0}6.4 & \phantom{0}3.5 & \phantom{0}0.9 \\
\rowcolor{black!4}
V    & \phantom{0}0.9 & 17.6 & 37.1 & 34.9 & 30.6 \\
\rowcolor{black!4}
O    & \phantom{0}7.1 & \phantom{0}8.9 & \phantom{0}1.3 & 40.4 & 54.0 \\
Gate & 38.4 & 30.3 & 30.9 & 10.7 & \phantom{0}5.4 \\
Up   & 38.4 & 23.5 & 16.4 & \phantom{0}9.3 & \phantom{0}8.9 \\
\rowcolor{black!4}
Down & \phantom{0}7.1 & 12.9 & \phantom{0}7.5 & \phantom{0}0.3 & \phantom{0}0.1 \\
\bottomrule
\end{tabular}
\end{table}

On base models, $\hat{S}$ circuits concentrate 70–95\% of budget in layers 0–2, with V and O projections dominant at 7B scale; on instruct models (including Qwen2.5-7b-instruct), budget redistributes toward mid-to-late layers (18\% early on Qwen-7B-Instruct, 90\% early on Llama-3B-Instruct concentrated in a single layer's Down projection. $\hat{F}$ shows the same early-layer pattern on base models (92–98\% in layers 0–4) but distributes more broadly on instruct models (15\% early on Qwen-7B-Instruct), consistent with instruct models requiring adjustments across a wider range of layers.

\paragraph{Vision-Language Models} We see similar preference for V and O modules. Most of the budget is now allocated to the vision encoder with concentration on the earlier layers. The random placement is heavily allocated towards the language model simply it has more parameters.

\begin{figure}[h]
\centering
\includegraphics[width=\linewidth]{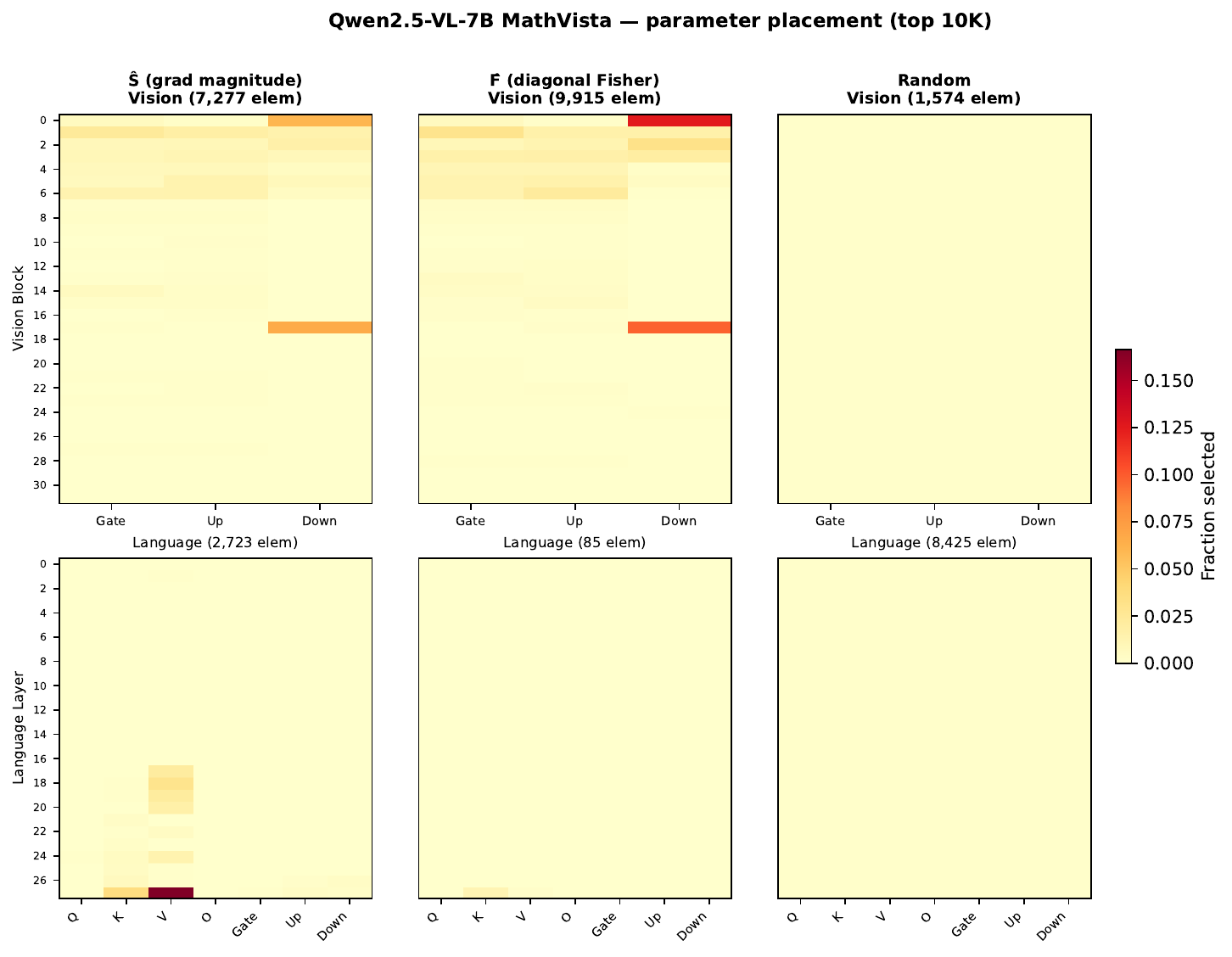}
\caption{\textbf{Qwen2.5-VL-7B, MathVista} (top 10K).  Circuit discovery automatically routes budget to the vision encoder: $\hat{S}$ allocates 73\% to vision (vs.\ 16\% for random), and $\hat{F}$ allocates 99\%.  Within vision, both methods concentrate on early blocks (0--6) and block 16.  The few language elements selected by $\hat{S}$ go to V/O in late layers (16--19, 26)---the same module preference as LLM-only circuits.  Random placement inverts this, allocating 84\% to language simply because it has more parameters.}
\label{fig:anatomy_vlm}
\end{figure}

The VLM result demonstrates that circuit discovery can identify which component of a multi-modal architecture is the training bottleneck.  For a visual math task, the language model already has strong mathematical reasoning; the gradient signal concentrates in the vision encoder, which must learn to extract task-relevant information from images.  A practitioner choosing modules to fine-tune would need domain expertise to reach this conclusion; the circuit scores reveal it automatically.

% \subsubsection{Score Distributions}

\begin{figure}[h]
\centering
\includegraphics[width=0.48\linewidth]{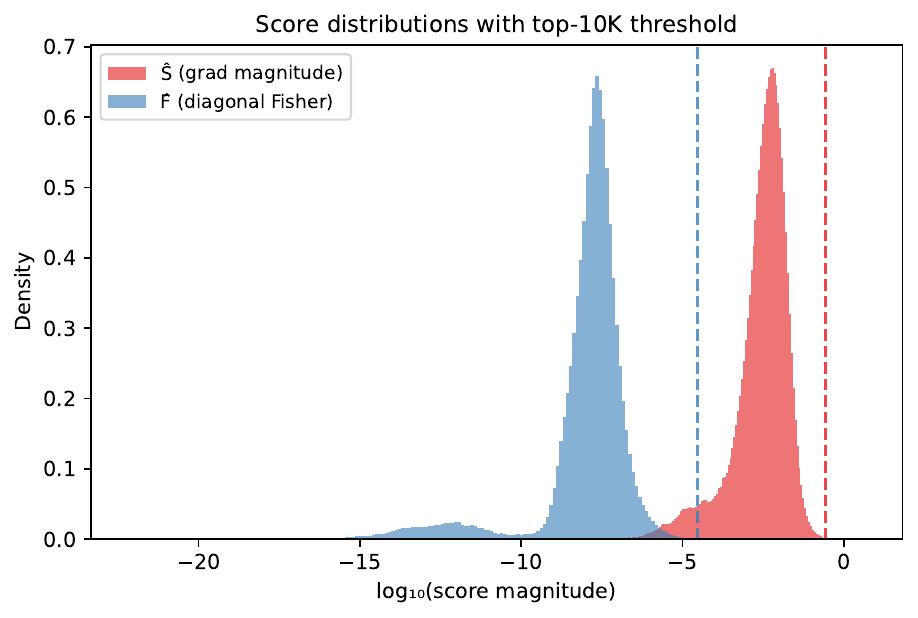}
\hfill
\includegraphics[width=0.48\linewidth]{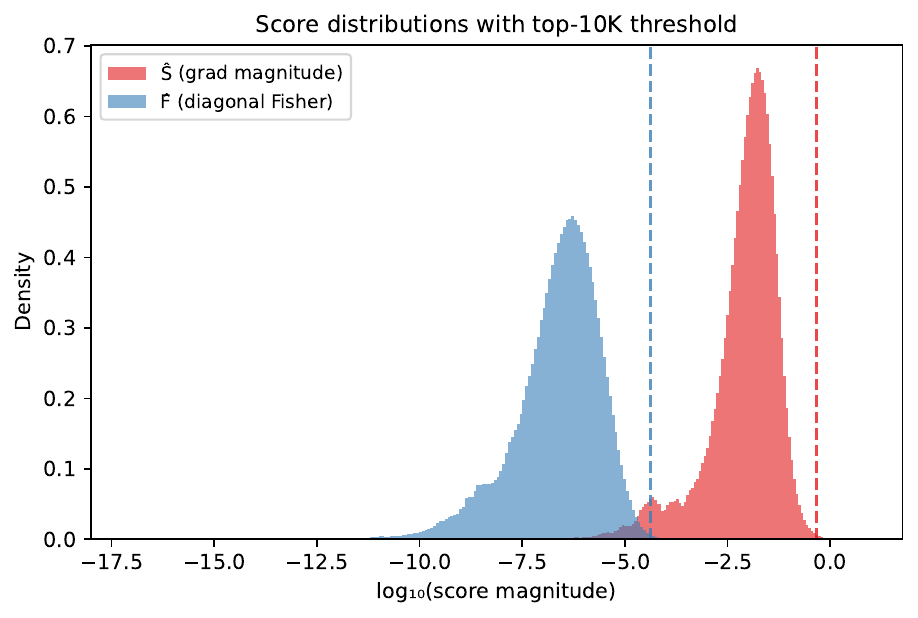}
\caption{\textbf{Score distributions} for Qwen2.5-3B on GSM8K (left) and MATH (right).  Both $\hat{S}$ and $\hat{F}$ exhibit heavy-tailed distributions; the top-10K threshold (dashed lines) selects from the extreme tail.  $\hat{F}$ scores span a wider dynamic range (${\sim}$10 orders of magnitude) than $\hat{S}$ (${\sim}$4 orders).}
\label{fig:score_distributions}
\end{figure}

\begin{figure}[h]
\centering
\includegraphics[width=0.48\linewidth]{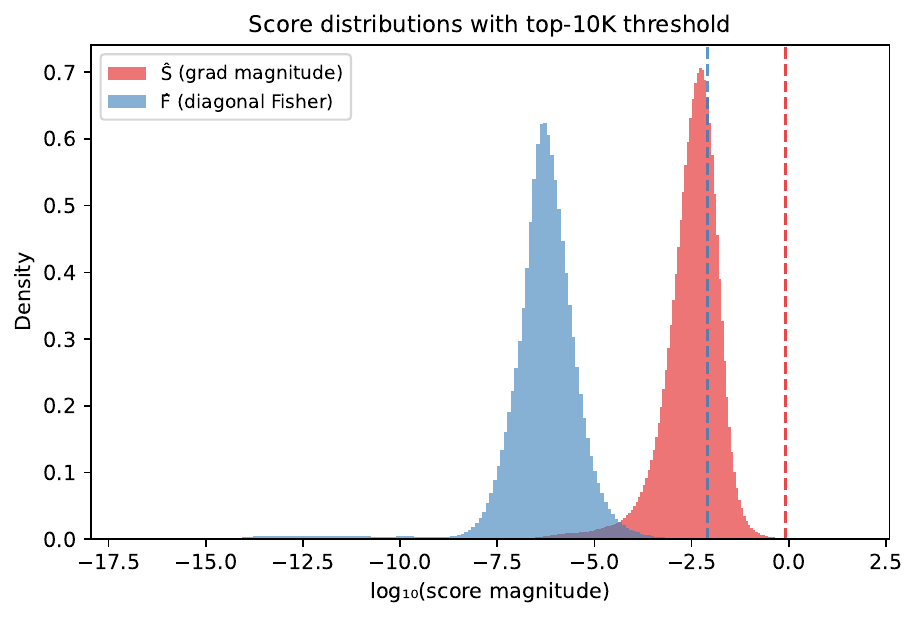}
\hfill
\includegraphics[width=0.48\linewidth]{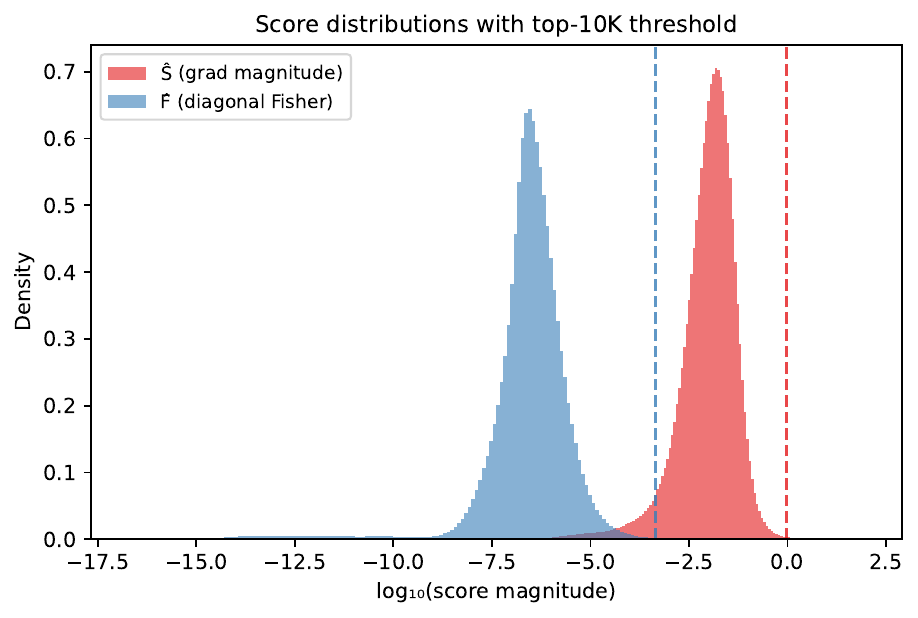}
\caption{\textbf{Score distributions} for Qwen2.5-7B on GSM8K (left) and MATH (right).}
\label{fig:score_distributions_7b}
\end{figure}

\subsection{Per-Layer Score Concentration}
\label{app:perlayer_concentration}

\begin{figure}[h]
\centering
\includegraphics[width=\linewidth]{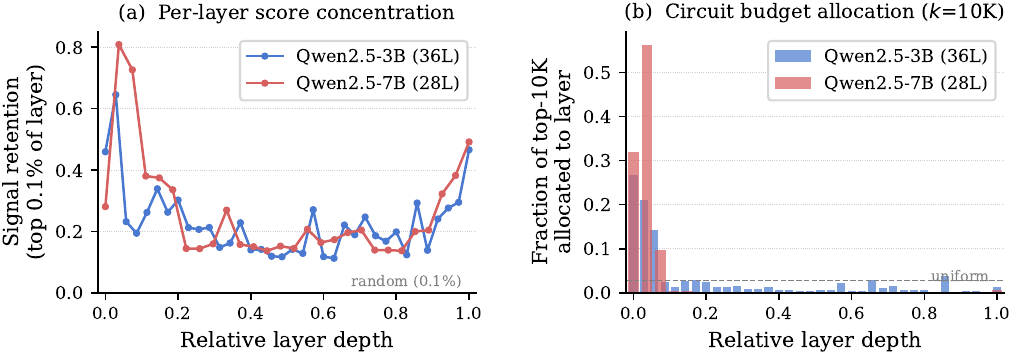}
\caption{\textbf{Per-layer circuit analysis} for Qwen2.5-3B (36L) and Qwen2.5-7B (28L) on GSM8K with budget $k{=}10$K.
\textbf{(a)}~Signal retention: fraction of each layer's gradient energy captured by the top 0.1\% of elements.  Both models show strong concentration in early layers (${\sim}40{-}80\times$ above the random baseline), confirming that circuit scores are far from uniform.
\textbf{(b)}~Circuit budget allocation: fraction of the global top-$k$ budget assigned to each layer.  The budget clusters heavily in the first ${\sim}20\%$ of layers, with a secondary peak near the output.}
\label{fig:perlayer_circuit}
\end{figure}

The heatmaps in \Cref{app:circuit_anatomy} show which modules receive budget; \Cref{fig:perlayer_circuit} shows how concentrated that budget is across depth.  Panel~(a) confirms that even within a single layer, gradient energy is far from uniformly distributed---the top 0.1\% of elements capture 40--80$\times$ more energy than a random subset of the same size.  Panel~(b) shows that global top-$k$ selection allocates budget non-uniformly across layers: early layers (depth $< 0.2$) receive a disproportionate share.  Both patterns are consistent across 3B and 7B, and both scoring functions ($\hat{S}$, $\hat{F}$) produce similar layer profiles.

\subsection{Cross-dataset circuit overlap.}   
\label{app:cross_dataset_overlap} 
We measure top-$k$ element overlap between $\hat{S}$ circuits discovered on three math datasets (GSM8K, MATH, NuminaMath) using Qwen2.5-1.5B.  Two random  $k$-element subsets of the 20.6M adapter parameters overlap by $k/20.6\text{M}$ in expectation. \Cref{tab:cross_dataset_overlap} shows that circuits share 21--30\% of elements across datasets, hundreds of times above chance, indicating that circuit discovery identifies a stable task-general subnetwork rather than dataset-specific artifacts.                   \begin{table}[h]                                                      \centering                                                            \caption{\textbf{Cross-dataset circuit overlap} (Qwen2.5-1.5B, $\hat{S}$).  Fraction of top-$k$ elements shared between circuits discovered on different math datasets.  Chance overlap is $k/20.6$M.}                                                           \label{tab:cross_dataset_overlap}                                                        
  \vspace{4pt}                                                                                                                                                     
  \small                                                                                                                                                           
  \setlength{\tabcolsep}{5pt}                                                                                                                                      
  \begin{tabular}{l ccc c}                                                                                                                                         
  \toprule                                                                                                                                                         
  \textbf{$k$} & \textbf{GSM8K--MATH} & \textbf{GSM8K--Numina} & \textbf{MATH--Numina} & \textbf{Chance} \\                                                        
  \midrule                                                                                                                                                         
  1K   & 21.1\% & 22.1\% & 16.7\% & 0.005\% \\                                                                                                                     
  5K   & 29.7\% & 29.1\% & 28.2\% & 0.024\% \\                                                                                                                     
  10K  & 29.6\% & 29.8\% & 29.6\% & 0.049\% \\                                                                                                                     
  50K  & 29.5\% & 30.9\% & 32.1\% & 0.24\%  \\                                                                                                                     
  500K & 35.6\% & 35.8\% & 37.3\% & 2.4\%   \\                                                                                                                     
  \bottomrule                                                                                                                                                      
  \end{tabular}                                                                                                                                                    
  \end{table}                                                        

%% ============================================================
\subsection{Scoring Functions: Formal Details}
\label{app:scoring_theory}

We provide formal definitions and convergence properties of the two circuit scoring functions introduced in \Cref{sec:circuit_discovery}.

\begin{definition}
The \textbf{Fisher score} for element $(i,j)$ of $\mathbf{B}$ is:
\[
\hat{F}_{ij} = \frac{1}{N} \sum_{k=1}^{N} g_{k,ij}^2.
\]
\end{definition}

As a sample mean of i.i.d.\ non-negative random variables, $\hat{F}_{ij} \to F_{ij} = \mathbb{E}[g_{ij}^2]$ almost surely by the strong law of large numbers, with $\mathrm{Var}(\hat{F}_{ij}) = O(1/N)$ by Chebyshev's inequality.

\begin{definition}
The \textbf{gradient magnitude score} (b\_zero) for element $(i,j)$ of $\mathbf{B}$ is:
\[
\hat{S}_{ij} = \left|\frac{1}{N}\sum_{k=1}^{N} g_{k,ij}\right|.
\]
\end{definition}

As a sample mean passed through the absolute value, $\hat{S}_{ij} \to |\mu_{ij}|$ where $\mu_{ij} = \mathbb{E}[g_{ij}]$, with convergence at rate $O(1/\sqrt{N})$ by the CLT.

\paragraph{Relationship.}  The two scores are connected by the bias-variance decomposition:
\[
\mathbb{E}[g_{ij}^2] = \mu_{ij}^2 + \sigma_{ij}^2.
\]
Fisher equals the squared mean gradient plus the gradient variance.  An element can have high Fisher but low $\hat{S}$ if its gradients are large but oscillate in sign; high $\hat{S}$ requires directional consistency across examples.

\paragraph{Why small $N$ suffices.}  For ranking stability, what matters is separation between top-ranked and median scores.  Empirically, the score distribution is heavy-tailed: top elements have scores 100--1000$\times$ above the median.  This large gap means even noisy estimates at small $N$ produce stable rankings.  We verify in \Cref{app:mc_convergence} that $N{=}50$ yields ${\geq}97\%$ overlap with the $N{=}100$ reference circuit across all model sizes.

%% ============================================================
\subsection{Cumulative Update Alignment}
\label{app:update_alignment}

Since $\mathbf{B} = 0$ at initialization, the trained $\mathbf{B}$ weights are the total displacement $\Delta \mathbf{B}$.  For a given circuit mask (top-$k$ elements by score) and a size-matched random mask, we measure what fraction of $\|\Delta \mathbf{B}\|^2$ falls on each subset.  The ratio $\text{Circuit} / \text{Random}$ tells us whether training concentrates weight changes onto the elements that circuit discovery identified.  A ratio above 1 means circuit elements move more than expected; below 1 means they move less.

\Cref{tab:alignment} reports the ratio of per-element update energy (mean squared displacement over training) for circuit elements vs.\ random elements at $k{=}10$K.

\begin{table}[h]
\centering
\caption{\textbf{Cumulative update alignment} ($k{=}10{,}000$, circuit/random energy ratio).  Under SFT, circuit elements move less than random ($<1\times$); under GRPO, they move more ($>1\times$).  The reversal holds across models, scoring methods, and tasks.}
\label{tab:alignment}
\vspace{4pt}
\small
\setlength{\tabcolsep}{5pt}
\begin{tabular}{ll cc cc}
\toprule
& & \multicolumn{2}{c}{\textbf{GSM8K}} & \multicolumn{2}{c}{\textbf{MATH-500}} \\
\cmidrule(lr){3-4} \cmidrule(lr){5-6}
\textbf{Model} & \textbf{Circuit} & \textbf{SFT} & \textbf{GRPO} & \textbf{SFT} & \textbf{GRPO} \\
\midrule
Qwen2.5-1.5B & $\hat{S}$ & $0.41\times$ & $\mathbf{2.04\times}$ & $0.41\times$ & $\mathbf{1.38\times}$ \\
Qwen2.5-1.5B & $\hat{F}$ & $0.40\times$ & $\mathbf{2.05\times}$ & $0.46\times$ & $1.35 \times$ \\
Qwen2.5-3B   & $\hat{S}$ & $0.24\times$ & $\mathbf{1.98\times}$ & $0.39\times$ & $\mathbf{1.60\times}$ \\
Qwen2.5-3B   & $\hat{F}$ & $0.26\times$ & $\mathbf{1.62\times}$ & $0.44\times$ & $1.53\times$ \\
\bottomrule
\end{tabular}
\end{table}

Under SFT, coherent gradients spread updates broadly---any subset accumulates comparable displacement, so circuit elements are not preferentially updated (0.24--0.41$\times$).  Under GRPO, incoherent gradients produce meaningful cumulative displacement only at elements with consistent gradient direction, exactly what $\hat{S}$ and $\hat{F}$ identify.  Circuit elements move 1.4--2.1$\times$ more than random, confirming that training concentrates learning onto the discovered circuit.

\paragraph{Weight-statistic scores are unreliable.}  The Wanda scoring function---a product of pretrained weight magnitude and mean gradient (\Cref{app:scoring_defs})---is inconsistent across architectures (\Cref{tab:scoring_consistency}).  Weight-magnitude methods do not reliably identify the elements onto which GRPO concentrates updates, further motivating gradient-based scoring.

%% ============================================================
\subsection{Gradient Structure: Full Results}
\label{app:gradient_full}

\Cref{tab:gradient_properties} (main text) reports gradient structure on MATH across three Qwen2.5 scales.  \Cref{tab:gradient_full} provides the measurements on GSM8K.  The SFT/GRPO dichotomy is consistent: SFT gradients are low-rank, directionally stable, and accumulate coherently, while GRPO gradients are higher-rank, near-orthogonal across steps, and largely cancel.
\begin{table}[h]                                                                                                                                                 
  \centering
  \caption{\textbf{Gradient structure under SFT vs.\ GRPO} (matched gradient steps at initialization).  Three metrics across Qwen2.5 base and instruct models on   
  GSM8K.  The SFT/GRPO contrast is consistent on base models; on instruct models both regimes produce diffuse gradients.}                                          
  \label{tab:gradient_full}
  \vspace{4pt}                                                                                                                                                     
  \small                                                                                                                                                           
  \setlength{\tabcolsep}{4pt}
  \begin{tabular}{@{}l cc cc cc@{}}                                                                                                                                
  \toprule                                                                                                                                                         
  & \multicolumn{2}{c}{\textbf{Eff.\ Gradient Rank}} & \multicolumn{2}{c}{\textbf{Cosine Similarity}} & \multicolumn{2}{c}{\textbf{Accum.\ Efficiency}} \\
  \cmidrule(lr){2-3} \cmidrule(lr){4-5} \cmidrule(lr){6-7}                                                                                                         
  \textbf{Model} & \textbf{SFT} & \textbf{GRPO} & \textbf{SFT} & \textbf{GRPO} & \textbf{SFT} & \textbf{GRPO} \\                                                   
  \midrule                                                                                                                                                         
  Qwen2.5-1.5B & 2.3 & 24.4 & 0.87 & 0.08 & 0.93 & 0.34 \\                                                                                                         
  Qwen2.5-3B   & 4.0 & 6.6  & 0.78 & $-$0.04 & 0.88 & 0.22 \\                                                                                                      
  Qwen2.5-7B   & 1.5 & 2.7  & 0.96 & 0.41 & 0.97 & 0.78 \\                                                                                                         
  \midrule                                                                                                                                                         
  \rowcolor{blue!5}                                                                                                                                                
  Qwen2.5-7B-Instruct & 12.6 & 5.9 & 0.54 & 0.27 & 0.73 & 0.52 \\                                                                                                  
  \bottomrule                                                                                                                                                      
  \end{tabular}                                                                                                                                                    
  \end{table}                                                                                                                                                      
\subsection{Knockout Sweep Validation}
\label{app:knockout}

The training experiments (\Cref{sec:exp_grpo}) demonstrate that circuit-placed parameters learn effectively while random-placed parameters do not.  Knockout sweeps provide independent, training-free confirmation: we zero out the top fraction of trained $\mathbf{B}$ entries by score, merge the residual LoRA into the base model, and evaluate accuracy.  If circuits identify functionally critical parameters, circuit-ordered knockout should degrade accuracy faster than removing entries at random.

\begin{figure}[h]
\centering
\includegraphics[width=\linewidth]{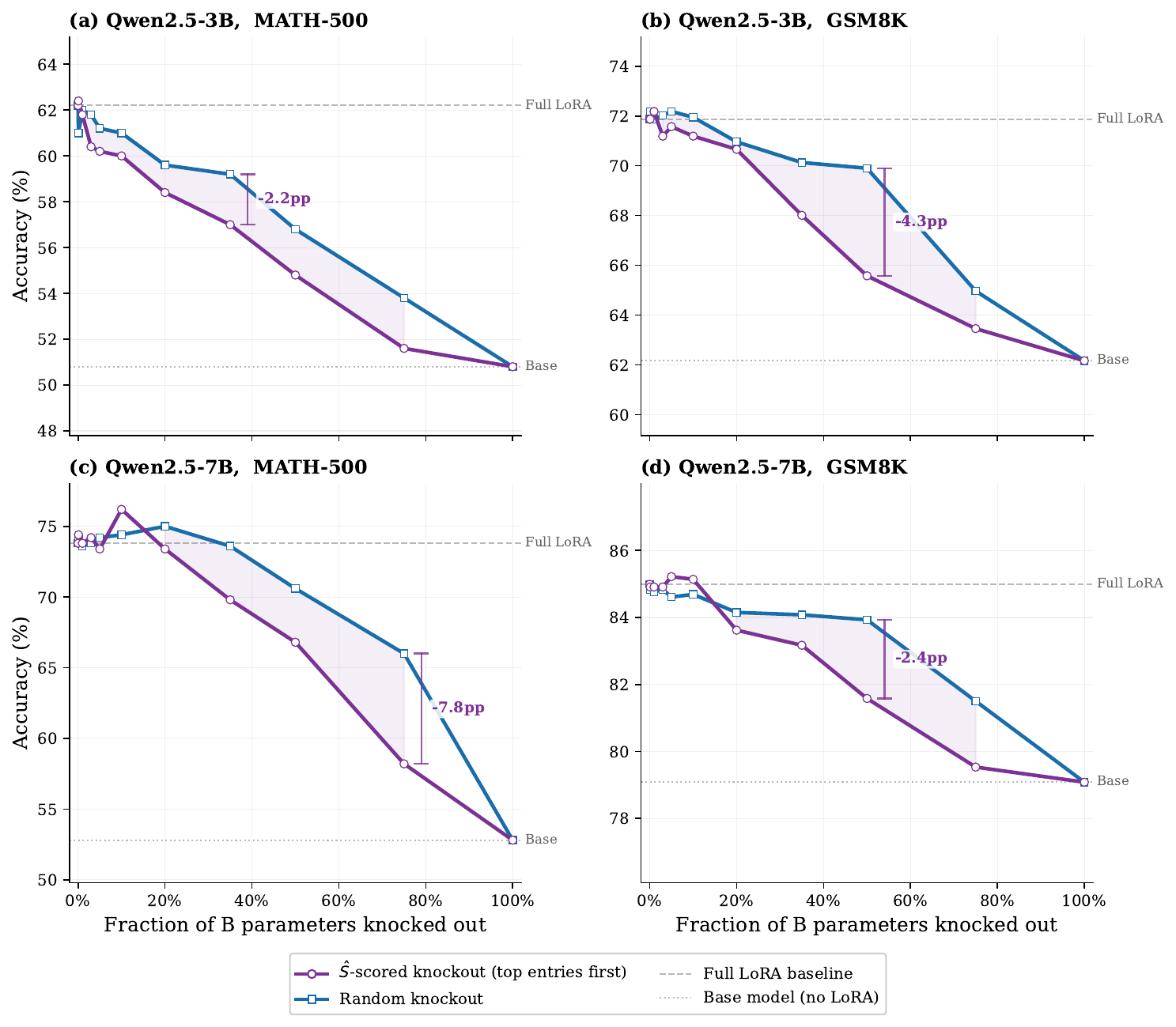}
\caption{\textbf{$\hat{S}$ knockout sweep}).  $\hat{S}$-ranked entries also degrade accuracy faster than random, but with smaller gaps: the peak is $-$7.8pp (7B MATH at 75\%) vs.\ $-$12.2pp for $\hat{F}$.}
\label{fig:knockout_shat}
\end{figure}

$\hat{S}$ knockout (\Cref{fig:knockout_shat}) shows the same qualitative pattern with uniformly smaller gaps: $-$7.8pp vs.\ $-$12.2pp on 7B MATH, $-$4.3pp vs.\ $-$5.0pp on 3B GSM8K.  This ranking matches the training performance comparison (\Cref{tab:grpo}) and reflects the scores' statistical properties.  $\hat{S}$ captures directional consistency ($|\mathbb{E}[g]|$), while $\hat{F}$ captures sensitivity ($\mathbb{E}[g^2] = \mu^2 + \sigma^2$).  Elements with large but sign-varying gradients score high on $\hat{F}$ but low on $\hat{S}$.  The knockout results confirm that these high-variance elements are functionally important, explaining why $\hat{F}$ identifies more concentrated circuits.

\begin{table}[h]
\centering
\caption{\textbf{Knockout sweep summary} across four configurations.  LoRA effect is the accuracy span between the full adapter and the base model.  Peak gap is the maximum difference between circuit-ordered and random knockout at any fraction.  $\hat{F}$ consistently produces larger gaps than $\hat{S}$, and the gap scales with LoRA effect size.}
\label{tab:knockout_summary}
\vspace{4pt}
\small
\setlength{\tabcolsep}{5pt}
\begin{tabular}{@{}ll ccc cc@{}}
\toprule
& & \textbf{Full LoRA} & \textbf{Base} & \textbf{LoRA} & \multicolumn{2}{c}{\textbf{Peak gap (pp)}} \\
\cmidrule(lr){6-7}
\textbf{Model} & \textbf{Task} & \textbf{(\%)} & \textbf{(\%)} & \textbf{effect} & $\hat{F}$ & $\hat{S}$ \\
\midrule
Qwen2.5-3B & MATH  & 62.8 & 49.4 & 13.4 & $-$7.2 & $-$2.2 \\
Qwen2.5-3B & GSM8K & 72.0 & 62.2 & \phantom{0}9.8 & $-$5.0 & $-$4.3 \\
Qwen2.5-7B & MATH  & 74.4 & 52.8 & 21.6 & $-$12.2 & $-$7.8 \\
Qwen2.5-7B & GSM8K & 85.2 & 79.5 & \phantom{0}5.7 & $-$2.5 & $-$2.4 \\
\bottomrule
\end{tabular}
\end{table}

%% ============================================================
\subsection{Module-Type Dissociation: Readers and Writers}
\label{app:module_dissociation}

The circuit heatmaps (\Cref{app:circuit_anatomy}) show which modules are selected; this section asks why.  Two analyses---gradient sign consistency and SVD alignment---reveal that selected parameters fall into two functional classes with distinct structural signatures: \textbf{attention readers} (V, K) and \textbf{residual writers} (O, Down), following the work of~\citep{elhage2021mathematical}.

\subsubsection{Gradient Sign Consistency}
\label{app:sign_consistency}

For each element of $\mathbf{B}$, we compute the fraction of training examples ($N{=}100$) on which the gradient sign agrees with the majority sign, then average across all elements of a given module type.  Higher consistency means the loss surface pushes that element in a stable direction---a prerequisite for accumulating meaningful displacement under noisy RL gradients (\Cref{sec:signal_retention}).

\begin{figure}[h]
\centering
\includegraphics[width=\linewidth]{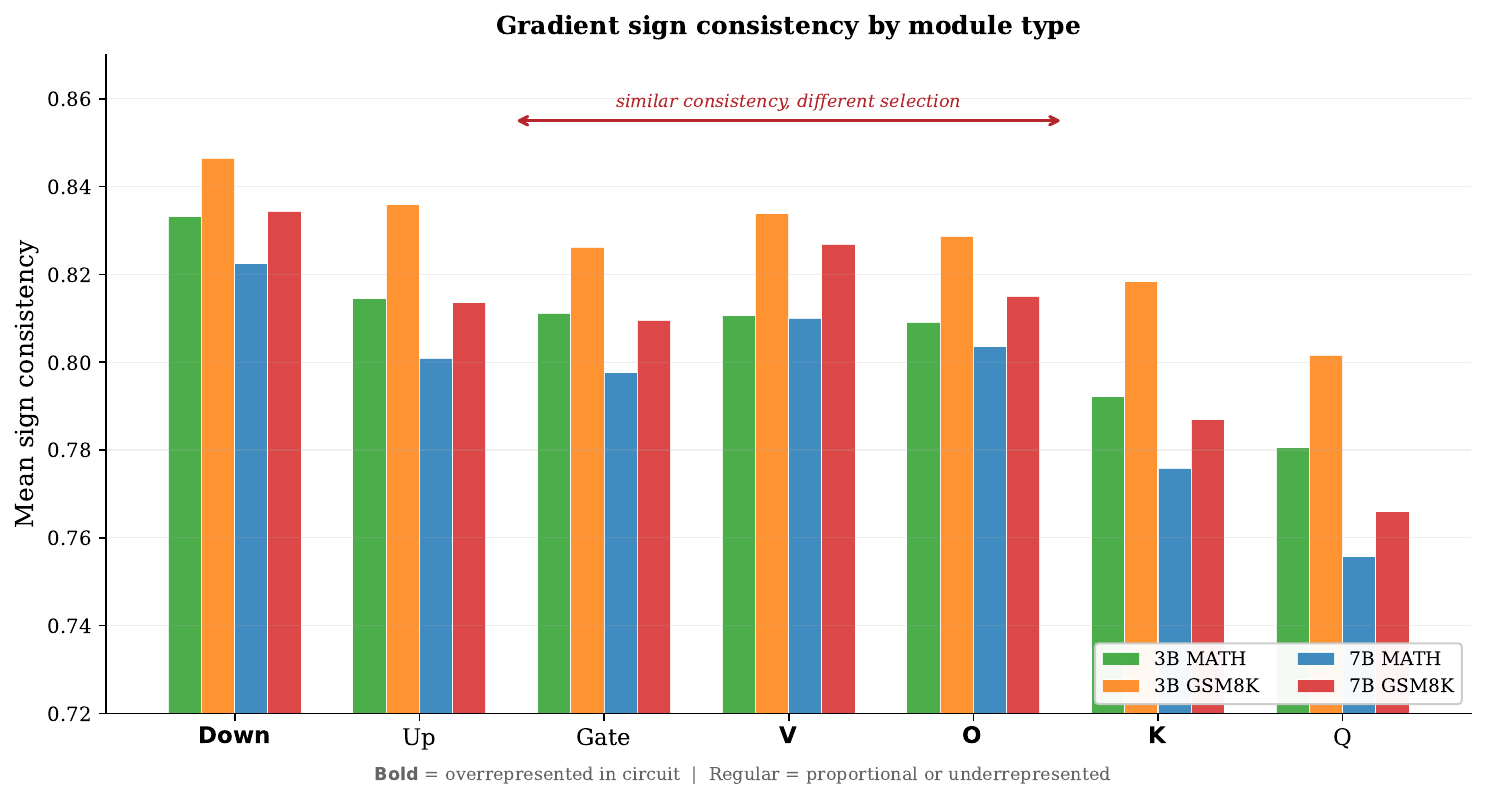}
\caption{\textbf{Gradient sign consistency by module type} (3B/7B $\times$ MATH/GSM8K).  Modules ordered by descending consistency; bold labels indicate modules overrepresented in circuits.  Down is consistently highest ($0.82{-}0.85$).  Gate and Up achieve consistency comparable to the selected modules V and O, yet are underrepresented in circuits---consistency is necessary but not sufficient for selection.}
\label{fig:sign_consistency}
\end{figure}

\Cref{fig:sign_consistency} reveals a stable hierarchy: Down $>$ \{V, Up, Gate, O\} $>$ K $>$ Q, where the middle group clusters within ${\sim}$0.02 of each other.  The ordering is consistent across model scale and task; only the absolute level shifts (GSM8K runs ${\sim}$0.02 higher, reflecting easier optimization).

The critical observation is the \textbf{Gate/Up anomaly}: Gate and Up achieve consistency comparable to V and O ($0.80{-}0.84$ vs.\ $0.80{-}0.83$) yet are consistently underrepresented in circuits (4--9\% capture rate vs.\ 20--87\% for V, O, Down).  Consistency is therefore necessary---Q, with the lowest consistency ($0.76{-}0.80$), is never overrepresented---but not sufficient.  MLP hidden activations are highly stable because the model reliably activates similar features across reasoning examples, producing consistent gradients at Gate and Up.  But scoring functions implicitly weight where the update lands: updates to V or O directly modify what the attention layer reads from or writes to the residual stream, while updates to Gate or Up modify an internal computation one step removed.

\subsubsection{SVD Alignment}
\label{app:svd_alignment}

Sign consistency measures how reliably each module's gradient points in one direction; SVD alignment measures where that direction lies relative to the pretrained weight structure.  For each trained LoRA adapter $\Delta \mathbf{W} = \mathbf{B}\mathbf{A}$, we compute the SVD of the base weight $\mathbf{W}_0 = \mathbf{U} \boldsymbol{\Sigma} \mathbf{V}^\top$ and measure two quantities:

\begin{itemize}[leftmargin=*,itemsep=2pt]
\item \textbf{Left alignment}: how much of $\Delta \mathbf{W}$'s column space overlaps with the top-$r$ left singular vectors of $\mathbf{W}_0$.  High left alignment means the adapter modifies the same output directions that the pretrained weight already emphasizes.
\item \textbf{Spectral ratio}: the fraction of $\|\Delta \mathbf{W}\|_F^2$ captured by the top-$r$ singular values of $\Delta \mathbf{W}$.  A high spectral ratio means the update is spectrally concentrated and effectively low-rank.
\end{itemize}

\begin{figure}[h]
\centering
\includegraphics[width=\linewidth]{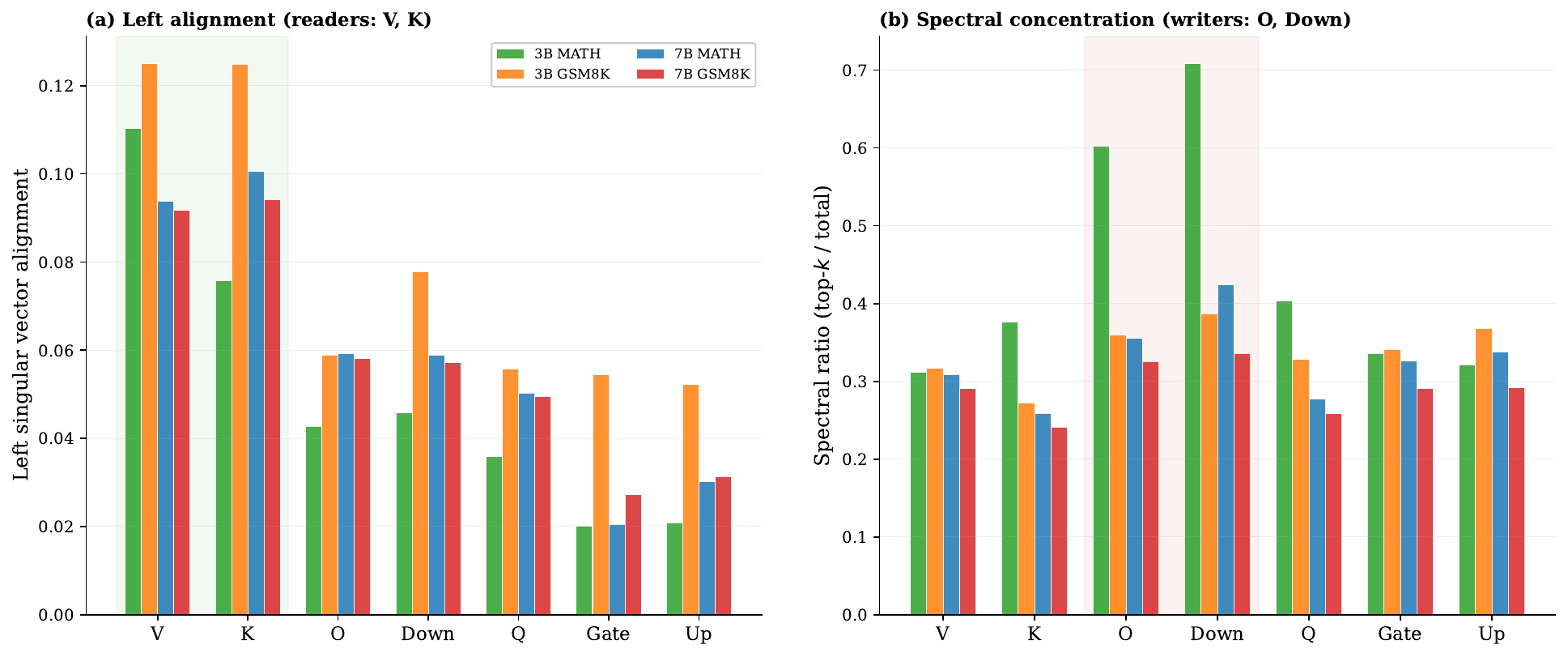}
\caption{\textbf{SVD alignment by module type} (3B/7B $\times$ MATH/GSM8K).  (a)~Left singular vector alignment: V and K dominate, identifying them as attention readers whose updates reinforce existing information-selection directions.  (b)~Spectral concentration: O and Down dominate, identifying them as residual writers whose updates are low-rank.  Shaded bands highlight the dominant pair in each panel.}
\label{fig:svd_alignment}
\end{figure}

\Cref{fig:svd_alignment} reveals a clean dissociation across all four configurations.  Panel~(a): V and K have the highest left alignment---their LoRA updates align with the dominant input-selection directions of the pretrained attention weights.  We term these \textbf{attention readers}: they refine what information the model attends to.  Panel~(b): O and Down have the highest spectral ratio---their updates are low-rank and concentrated.  We term these \textbf{residual writers}: they control what gets written back to the residual stream.

The dissociation is structurally grounded.  V and K sit at the input of the attention mechanism, projecting token representations into query and key spaces; high left alignment means the adapter adjusts attention patterns within the subspace the pretrained model already uses.  O and Down sit at the output of attention and MLP blocks respectively, projecting back into the residual stream; high spectral concentration means the adapter targets a narrow set of output directions, consistent with steering predictions through a low-rank perturbation.  Gate, Up, and Q---which operate on internal representations and neither read from nor write to the residual stream directly---show neither elevated left alignment nor spectral concentration.

\paragraph{Two-factor selection model.}  Sign consistency and SVD alignment provide complementary selection criteria.  Consistency is a necessary condition: an element must have a stable gradient direction to accumulate displacement under RL (Down, V, Up, Gate, O all satisfy this).  Proximity to the residual stream is a sufficiency condition: among consistent elements, those that read from the stream (V, K: high left alignment) or write to it (O, Down: high spectral concentration) are functionally most impactful.  The Gate/Up anomaly resolves under this two-factor model: high consistency but low functional impact, because their updates target internal MLP activations rather than the residual stream interface.

%% ============================================================
\subsection{Cross-Architecture Validation: Llama-3.2-3B}
\label{app:llama_dissociation}

The analyses in \Cref{app:module_dissociation} use Qwen2.5 models exclusively.  To test whether the Readers/Writers dissociation is architecture-specific or reflects a general property of transformer fine-tuning under RL, we replicate all three measurements on Llama-3.2-3B-Instruct trained with GRPO on MATH.

\begin{figure}[h]
\centering
\includegraphics[width=\linewidth]{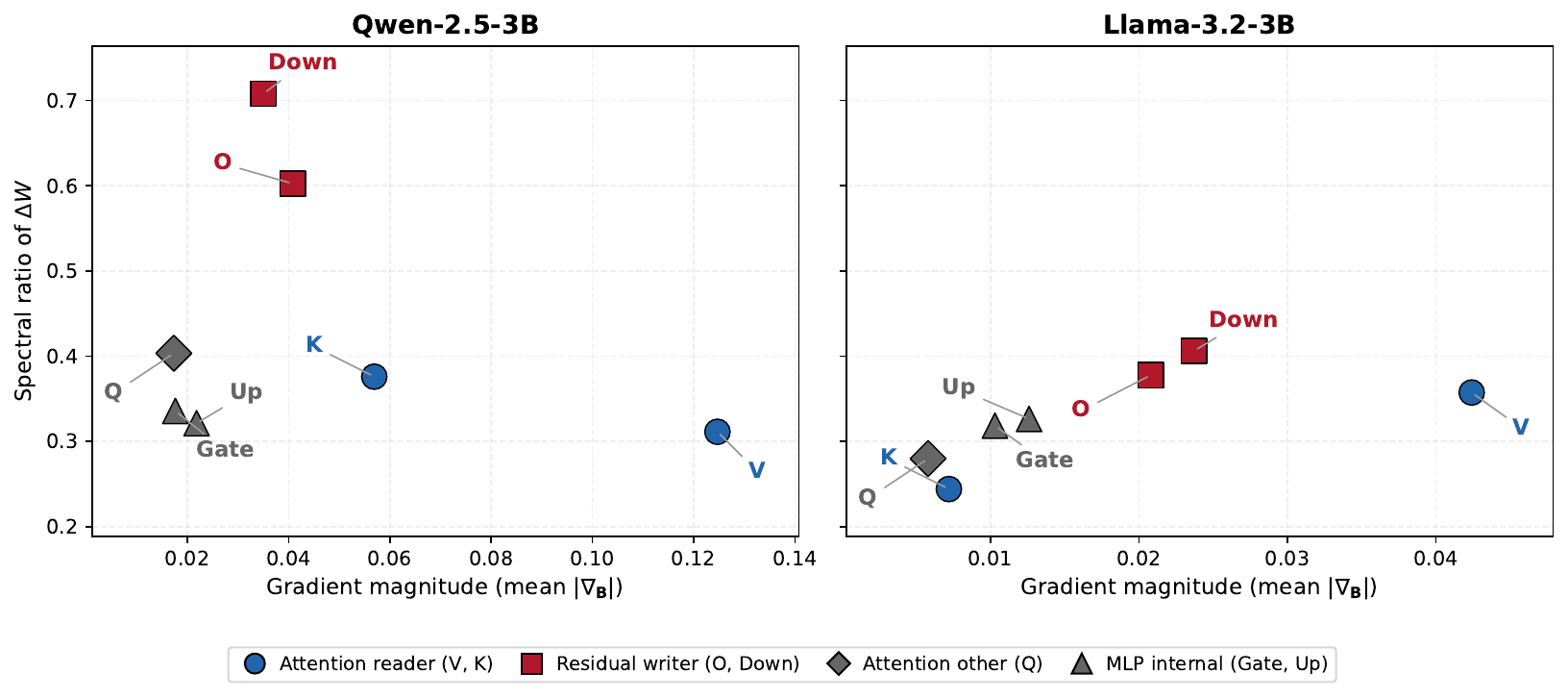}
\caption{\textbf{Cross-architecture dissociation.}  Each point is one module type, positioned by its gradient magnitude (mean $|\nabla_\mathbf{B}|$, $x$-axis) and spectral concentration of $\Delta W$ ($y$-axis).  In both Qwen-2.5-3B and Llama-3.2-3B, residual writers (O, Down; red squares) cluster at high spectral ratio while attention readers (V, K; blue circles) receive the strongest gradient signal.  The spatial separation of these two functional roles is preserved across architectures despite differences in attention implementation (multi-head vs.\ grouped-query).}
\label{fig:cross_arch_dissociation}
\end{figure}

\begin{table}[h]
\centering
\caption{\textbf{Module-level analysis: Llama-3.2-3B-Instruct on MATH} (GRPO, 28 layers).  Three complementary measurements per module type, averaged across all layers.  The Readers/Writers dissociation transfers from Qwen to Llama: V and K show the highest left alignment (attention readers), O and Down show the highest spectral concentration (residual writers), and Gate/Up show high sign consistency but neither structural signature---the same anomaly observed in Qwen (\Cref{fig:sign_consistency}).}
\label{tab:llama_module_analysis}
\vspace{4pt}
\small
\setlength{\tabcolsep}{5pt}
\begin{tabular}{@{}l ccc cc@{}}
\toprule
& \multicolumn{3}{c}{\textbf{Sign Consistency}} & \multicolumn{2}{c}{\textbf{SVD Alignment}} \\
\cmidrule(lr){2-4} \cmidrule(lr){5-6}
\textbf{Module} & \textbf{Mean} & \textbf{Frac $>$0.8} & \textbf{SNR} & \textbf{Left Align} & \textbf{Spec.\ Ratio} \\
\midrule
\rowcolor{black!4}
V    & 0.780 & 0.482 & 0.354 & \textbf{0.066} & 0.357 \\
\rowcolor{black!4}
K    & 0.763 & 0.425 & 0.313 & \textbf{0.081} & 0.244 \\
Q    & 0.761 & 0.419 & 0.301 & 0.049 & 0.280 \\
\rowcolor{black!4}
O    & 0.778 & 0.476 & 0.352 & 0.056 & \textbf{0.378} \\
Gate & 0.792 & 0.518 & 0.358 & 0.022 & 0.318 \\
Up   & 0.785 & 0.498 & 0.353 & 0.024 & 0.326 \\
\rowcolor{black!4}
Down & 0.787 & 0.505 & 0.371 & 0.061 & \textbf{0.406} \\
\bottomrule
\end{tabular}
\end{table}

\begin{table}[h]
\centering
\caption{\textbf{Gradient magnitude by module type} (Llama-3.2-3B-Instruct, MATH, $\mathbf{B}$-matrix only).  Mean $|\nabla_{\mathbf{B}}|$ averaged across all layers.  V dominates ($4{-}6\times$ over Gate/Up), followed by O and Down---the same ordering as Qwen2.5, and consistent with the Readers/Writers interpretation: modules at the residual stream interface receive the strongest gradient signal under RL.}
\label{tab:llama_gradient_magnitude}
\vspace{4pt}
\small
\setlength{\tabcolsep}{8pt}
\begin{tabular}{@{}l c l@{}}
\toprule
\textbf{Module} & $\mathbb{E}[|\nabla_\mathbf{B}|]$ & \textbf{Role} \\
\midrule
\rowcolor{black!4}
V    & 0.042 & Attention reader \\
\rowcolor{black!4}
Down & 0.024 & Residual writer \\
\rowcolor{black!4}
O    & 0.021 & Residual writer \\
Up   & 0.013 & MLP internal \\
Gate & 0.010 & MLP internal \\
\rowcolor{black!4}
K    & 0.007 & Attention reader \\
Q    & 0.006 & --- \\
\bottomrule
\end{tabular}
\end{table}

\Cref{fig:cross_arch_dissociation} visualizes the dissociation: writers occupy the upper region (high spectral ratio) while readers extend rightward (high gradient magnitude), with the same spatial layout in both architectures.  \Cref{tab:llama_module_analysis} confirms that all three structural signatures transfer:

\begin{enumerate}[leftmargin=*,itemsep=2pt]
\item \textbf{Sign consistency}: The same Gate/Up anomaly appears---Gate (0.792) and Up (0.785) match or exceed V (0.780) and O (0.778), yet gradient magnitude (\Cref{tab:llama_gradient_magnitude}) shows V receives $4\times$ the gradient signal.  Consistency is necessary but not sufficient, exactly as in Qwen.
\item \textbf{Left alignment}: V (0.066) and K (0.081) have the highest alignment with the pretrained weight's dominant singular vectors, confirming their role as attention readers that refine existing information-selection directions.
\item \textbf{Spectral concentration}: Down (0.406) and O (0.378) produce the most concentrated (low-rank) updates, confirming their role as residual writers that steer output through narrow perturbations.
\end{enumerate}

The dissociation is not an artifact of the Qwen architecture or its specific attention implementation (grouped-query vs.\ multi-head).  Llama-3.2-3B uses grouped-query attention (8 KV heads vs.\ 24 query heads), giving K and V matrices $3\times$ fewer parameters than Q and O, yet V still dominates gradient magnitude and K still shows the highest left alignment.  The functional roles---reading from vs.\ writing to the residual stream---are determined by the transformer's computational graph, not by architecture-specific hyperparameters.

\subsection{Weight Norm and Behavior Change Dynamics}
\label{app:dynamics}
Here we continue the analysis on the dynamics between how selected neurons change and how much model predictions change, across training checkpoints (top) and across model layers (bottom)--but extend it to Llama models and using KL instead of Jaccard. This was first discussed in \Cref{sec:dynamics}. We repeat the setup below for reader convenience.

\paragraph{Setup} We focus on the dynamics for Qwen2.5-1.5B/3B/7B and Llama-3.2-3B/3.1-8B (GRPO on MATH) over 100 test examples from MATH, focused only on the processing of the prompt and problem definition, compared to the respective base models. We report $1 -$ \textit{top-25 overlap}, where overlap is the Jaccard overlap of the top-25 predicted tokens between the fine-tuned and base model, averaged over positions and examples. The \textit{top} row x-axis reports the effective norm of the LoRA trainable weights, $\|\mathbf{B}_{\mathrm{nz}}\|_2 \cdot \alpha/r$. Since $\mathbf{B} = 0$ at initialization, the non-zero $\mathbf{B}_{\mathrm{nz}}$ weights are the total displacement. The y-axis captures some of the behavioral change, as reflected by the change in output token distributions. For top, this is the projection across the vocabulary across training checkpoints; for bottom, these values are across model layers (x-axis) using logit lens \citep{belrose2023eliciting}, projecting residual hidden states to the vocabulary logits.

The two metrics we use to capture how much the model's behavior is changing are:
 \begin{enumerate}
     \item  $1 -$ \textit{top-25 overlap}, where overlap is the Jaccard-style fraction of top-25 predicted tokens shared between each pair of models. Higher values indicate greater divergence. A value of 0 indicates identical behavior to the base model.
     \item KL divergence between each pair of models.
 \end{enumerate}
 Both metrics are averaged across positions and examples. Higher values indicate greater divergence.

\begin{figure}[t]
\centering
\input{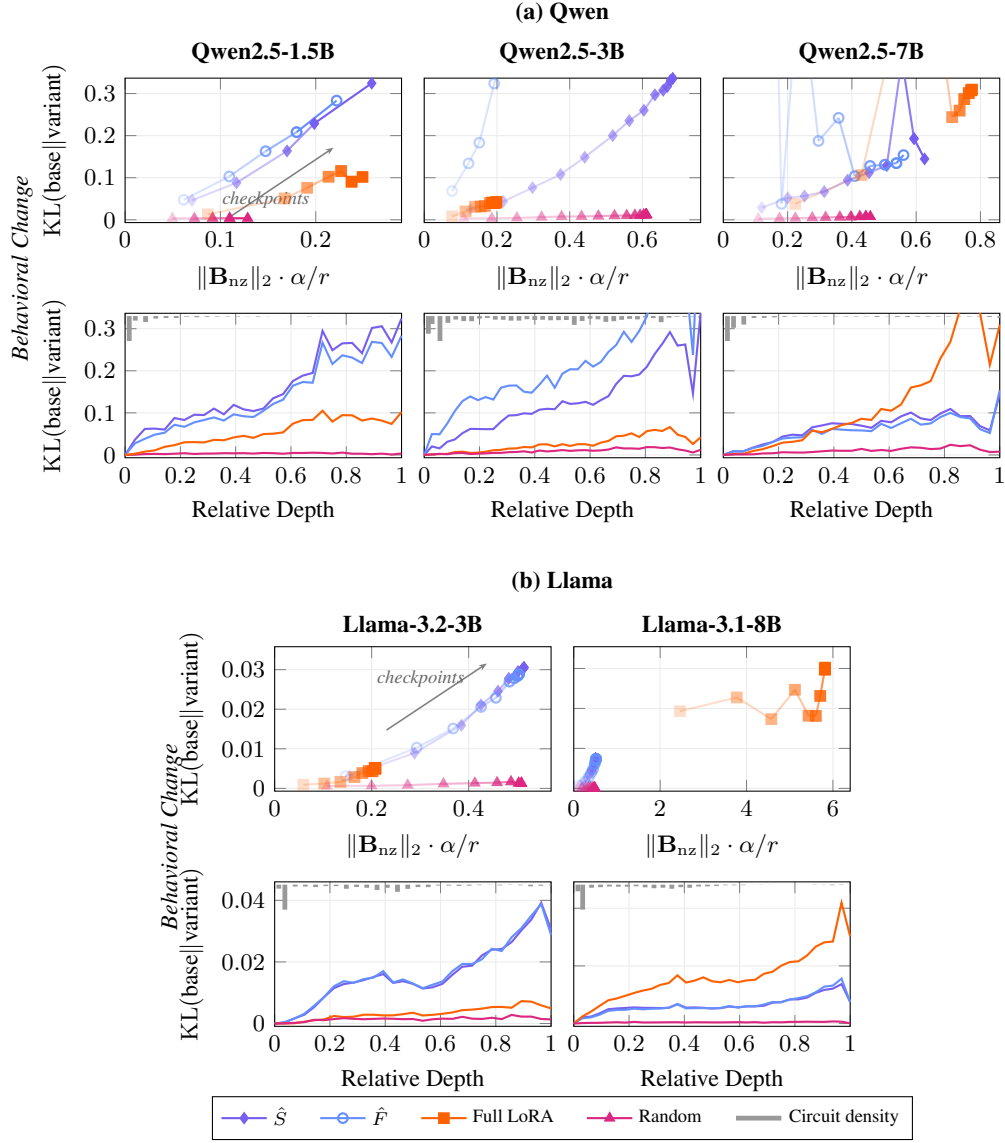}
\caption{\textbf{Divergence vs.\ effective update norm} (a: Qwen2.5-1.5B/3B/7B, MATH-500, GRPO); (B: Llama-3.2-3B/Llama-3.18B, MATH-500, GRPO). For both rows, the $y$-axis is $1 - \textit{top-25 token overlap}$, measuring behavioral change via logit lens relative to the base model. Top: each trajectory traces training checkpoints, opacity increases with step. $x$-axis: $|\mathbf{B}_{\mathrm{nz}}|_2 \cdot \alpha/r$, the scaled norm of non-zero $\mathbf{B}$ weights. Bottom: per-layer divergence at the final checkpoint. $x$-axis: relative layer depth. Gray bars report the relative circuit density.  }
\label{fig:analysis_app_kl}
\end{figure}

In \Cref{fig:analysis_app_kl}, we replicate the analysis from \Cref{fig:analysis}, using KL divergence instead of top-25 overlap. 

\Cref{tab:divergence} shows the final-layer (final training checkpoint) output distribution metrics across a number of settings. This is over 100 test examples from MATH, focused only on the processing of the prompt and problem definition, for Qwen2.5-1.5B/3B/7B and Llama-3.2-3B/3.1-8B. The two metrics we report are:

% Auto-generated by pgf_divergence_table_appendix.py — do not edit by hand.
\begin{table*}[t]
\centering
\caption{\textbf{Distributional divergence at final checkpoints (all families).} Models Qwen2.5-1.5B, -3B, -7B and Llama-3.2-3B, -3.1-8B are marked by Q- and L-, respectively. We report the  KL$(M_1 \| M_2)$ and $1{-}\text{Top-25 overlap}$ at the final transformer layer between different model pairs ($M_1, M_2$). This is approximated from top-25 token distributions over 100 MATH-500 prompts and problem texts.}
\label{tab:divergence}
\vspace{4pt}
\small
\setlength{\tabcolsep}{3pt}
\begin{tabular}{l ccccc ccccc}
\toprule
& \multicolumn{5}{c}{\textbf{KL divergence}} & \multicolumn{5}{c}{\textbf{1\,--\,Top-25 overlap}} \\
\cmidrule(lr){2-6} \cmidrule(lr){7-11}
& Q-1.5B & Q-3B & Q-7B & L-3B & L-8B & Q-1.5B & Q-3B & Q-7B & L-3B & L-8B \\
\midrule
Full-LoRA vs $\hat{F}$ & 0.161 & 0.154 & 0.180 & 0.019 & 0.244 & 0.185 & 0.153 & 0.150 & 0.117 & 0.274 \\
Full-LoRA  vs $\hat{S}$ & 0.186 & 0.257 & 0.166 & 0.021 & 0.290 & 0.204 & 0.185 & 0.142 & 0.121 & 0.277 \\
Full-LoRA  vs Random & 0.122 & 0.033 & 0.273 & 0.006 & 0.337 & 0.142 & 0.069 & 0.192 & 0.030 & 0.312 \\
Full-LoRA  vs Pre-trained & 0.125 & 0.047 & 0.290 & 0.007 & 0.341 & 0.144 & 0.085 & 0.197 & 0.033 & 0.314 \\
\midrule
Pre-trained vs $\hat{F}$ & 0.283 & 0.240 & 0.154 & 0.029 & 0.074 & 0.241 & 0.196 & 0.122 & 0.134 & 0.172 \\
Pre-trained vs $\hat{S}$ & 0.324 & 0.339 & 0.145 & 0.031 & 0.075 & 0.263 & 0.226 & 0.132 & 0.136 & 0.180 \\
Pre-trained vs Random & 0.004 & 0.012 & 0.008 & 0.001 & 0.001 & 0.013 & 0.027 & 0.020 & 0.014 & 0.014 \\
\bottomrule
\end{tabular}
\end{table*}

\paragraph{More Findings} Beyond the findings discussed in the main body of the paper, \Cref{tab:divergence} suggests a number of takeaways: (1) selected circuits find new solutions rather than replicating LoRA behavior; (2) random circuits do not diverge far from the base model; (3) the final divergence between full LoRA and the circuits is smaller than that between the circuits and the base model.

\begin{figure}[t]
\centering
    \input{figures-pgf/divergence_vs_norm_appendix}
\caption{\textbf{Divergence vs.\ effective update norm} (a: Qwen2.5-1.5B/3B/7B, MATH-500, GRPO); (B: Llama-3.2-3B/Llama-3.18B, MATH-500, GRPO). For both rows, the $y$-axis is $1 - \textit{top-25 token overlap}$, measuring behavioral change via logit lens relative to the base model. Top: each trajectory traces training checkpoints, opacity increases with step. $x$-axis: $|\mathbf{B}_{\mathrm{nz}}|_2 \cdot \alpha/r$, the scaled norm of non-zero $\mathbf{B}$ weights. Bottom: per-layer divergence at the final checkpoint. $x$-axis: relative layer depth. Gray bars report the relative circuit density.  }
\label{fig:analysis_app}
\end{figure}

\clearpage
\section{Hyperparameters}
\label{app:hyperparams}
\localtableofcontents

\subsection{Common Setup}
All experiments run on NVIDIA RTX PRO 6000 GPUs (96\,GB each).  Qwen2.5 base-models GRPO uses TRL \citep{trl} with PEFT in bf16 across two GPUs via Accelerate.  Llama instruct GRPO uses Unsloth \citep{unsloth} with 4-bit quantization on a single GPU.  SFT uses the HuggingFace SFTTrainer on a single GPU.

\paragraph{Greedy evaluation.}  All main results (\Cref{tab:sft_7bench,tab:grpo}) use greedy decoding: temperature 0.0, no nucleus sampling.  We use vllm and \texttt{lm\_eval\_harness}~\citep{eval-harness} for SFT. 

%% ============================================================
\subsection{GRPO Training: Qwen2.5 Base Models}
\label{app:hp_grpo_qwen}

\Cref{tab:hp_grpo_qwen} reports the configuration shared across all Qwen2.5 base-model GRPO runs (GSM8K and MATH-500).  Where 7B differs from 1.5B/3B, we note the deviation. \Cref{tab:hp_lora_rank} reports the LoRA rank used for each model--task combination.  The 1.5B GSM8K results use rank 32; all other GRPO configurations use rank 8.  In all cases, LoRA is applied to every linear layer in the transformer block (q, k, v, o, gate, up, down projections), with $\mathbf{A} \sim \mathcal{N}(0, 2/d)$ and $\mathbf{B} = \mathbf{0}$.

\begin{table}[h]
\centering
\caption{\textbf{LoRA rank and $\alpha$ per GRPO configuration.}  Sparse circuits scale $\alpha = 32 \times r$ to compensate for the reduced number of active parameters.}
\label{tab:hp_lora_rank}
\vspace{4pt}
\small
\setlength{\tabcolsep}{5pt}
\begin{tabular}{ll ccc}
\toprule
\textbf{Model} & \textbf{Task} & \textbf{$r$} & \textbf{$\alpha$ (full)} & \textbf{$\alpha$ (sparse)} \\
\midrule
Qwen2.5-1.5B & GSM8K  & 32 & 32  & 1024 \\
Qwen2.5-1.5B & MATH   & 32 & 32  & 1024 \\
Qwen2.5-3B   & GSM8K  & 32 & 32  & 1024 \\
Qwen2.5-3B   & MATH   &  8 & 32  & 256 \\
Qwen2.5-7B   & GSM8K  &  8 & 32  & 256 \\
Qwen2.5-7B   & MATH   &  8 & 32  & 256 \\
\bottomrule
\end{tabular}
\end{table}

\Cref{tab:hp_grpo_qwen} reports the remaining training configuration, shared across all Qwen2.5 base-model GRPO runs.  All runs train for 2{,}000 steps.

\begin{table}[h]
\centering
\caption{\textbf{GRPO training hyperparameters (Qwen2.5 base models).}  Where 7B differs, the deviation is noted.}
\label{tab:hp_grpo_qwen}
\vspace{4pt}
\small
\setlength{\tabcolsep}{5pt}
\begin{tabular}{l l l}
\toprule
\textbf{Hyperparameter} & \textbf{1.5B / 3B} & \textbf{7B} \\
\midrule
\multicolumn{3}{l}{\emph{Optimization}} \\
\midrule
Loss type                & DAPO                   & DR-GRPO \\
KL coefficient ($\beta$) & 0.0                   & 0.0 \\
Learning rate            & $1 \times 10^{-6}$     & $1 \times 10^{-6}$ \\
LR scheduler             & constant               & constant \\
Warmup steps             & 25                     & 25 \\
Optimizer                & AdamW          & AdamW \\
Adam $(\beta_1, \beta_2)$ & (0.9, 0.999)         & (0.9, 0.999) \\
Weight decay             & 0.0                    & 0.0 \\
Max gradient norm        & 0.1                    & 0.1 \\
Precision                & bf16                   & bf16 \\
\midrule
\multicolumn{3}{l}{\emph{GRPO generation}} \\
\midrule
Generations per prompt   & 4                      & 4 \\
Max completion length    & 1024                   & 1024 \\
Sampling temperature     & 1.0                    & 1.0 \\
\midrule
\multicolumn{3}{l}{\emph{Batching and schedule}} \\
\midrule
Per-device batch size    & 1                      & 1 \\
Gradient accumulation    & 4                      & 4 \\
Gradient checkpointing   & \multicolumn{2}{l}{yes} \\
\bottomrule
\end{tabular}
\end{table}

\paragraph{Loss types.}  DAPO \citep{dapo} removes the KL penalty ($\beta = 0$) and applies clipped importance weights.  DR-GRPO \citep{drgrpo} additionally length-normalizes the per-token log-probabilities to remove the bias toward shorter completions.  We use DR-GRPO for 3B MATH, 7B GSM8K, and 7B MATH; while 1.5B and 3B GSM8K train with DAPO.

\paragraph{Alpha scaling for sparse circuits.}  When only $k$ of the $\mathbf{B}$-matrix elements are trainable, the effective adapter norm shrinks proportionally.  To compensate, we scale $\alpha$ to $32 \times r$ (\Cref{tab:hp_lora_rank}).  This scaling is applied uniformly to all sparse and random runs; full LoRA runs use $\alpha = 32$.  We ablate alpha scaling in \Cref{app:random_grpo}.

\paragraph{Reward function.}  All GRPO runs use binary correctness rewards.  For GSM8K, the model is prompted with ``Put your final numerical answer after \texttt{\#\#\#\#}'' and the answer is extracted and compared symbolically using \texttt{math\_verify}.  For MATH, the prompt requests the answer in \verb|\boxed{}| format and the same \texttt{math\_verify} library grades symbolic equivalence.

\paragraph{Checkpoint selection.}  We evaluate every saved checkpoint (every 200 steps on MATH, final checkpoint on GSM8K) on the corresponding test set with greedy decoding and report the best checkpoint per method.  All methods within a comparison use the same evaluation protocol.

\paragraph{Training data.}  GSM8K uses the standard 7{,}473-example training split \citep{cobbe2021gsm8k}.  MATH uses all 7{,}500 training examples from EleutherAI/hendrycks\_math (7 subjects concatenated) \citep{hendrycks2021math}.

%% ============================================================
\subsection{GRPO Training: Llama Instruct Models}
\label{app:hp_grpo_llama}

Llama instruct models (Llama-3.2-3B-Instruct, Llama-3.1-8B-Instruct) use BNPO loss \citep{bnpo} throughout, with 4-bit quantized base weights via Unsloth \citep{unsloth}.  All runs train for 2{,}000 steps with checkpoints saved every 200 steps.  \Cref{tab:hp_grpo_llama} reports the full configuration.

\begin{table}[h]
\centering
\caption{\textbf{GRPO training hyperparameters (Llama instruct models).}}
\label{tab:hp_grpo_llama}
\vspace{4pt}
\small
\setlength{\tabcolsep}{5pt}
\begin{tabular}{l l l}
\toprule
\textbf{Hyperparameter} & \textbf{3B} & \textbf{8B} \\
\midrule
\multicolumn{3}{l}{\emph{LoRA configuration}} \\
\midrule
Rank ($r$)               & 32                     & 32 \\
$\alpha$                 & 32                     & 32 \\
Target modules           & \multicolumn{2}{l}{q, k, v, o, gate, up, down projections} \\
Quantization             & \multicolumn{2}{l}{4-bit (Unsloth)} \\
\midrule
\multicolumn{3}{l}{\emph{Optimization}} \\
\midrule
Loss type                & BNPO                   & BNPO \\
KL coefficient ($\beta$) & 0.002                 & 0.002 \\
Learning rate            & $5 \times 10^{-6}$     & $2 \times 10^{-6}$ \\
LR scheduler             & cosine                 & cosine \\
Warmup ratio             & 5\%                    & 5\% \\
Optimizer                & AdamW 8-bit            & AdamW 8-bit \\
Adam $(\beta_1, \beta_2)$ & (0.9, 0.95)          & (0.9, 0.95) \\
Weight decay             & 0.1                    & 0.1 \\
Max gradient norm        & 0.1                    & 0.1 \\
\midrule
\multicolumn{3}{l}{\emph{GRPO generation}} \\
\midrule
Generations per prompt   & 4                      & 4 \\
Max completion (GSM8K)   & 1024                   & 1024 \\
Max completion (MATH)    & ---                    & 2048 \\
Sampling temperature     & 1.0                    & 1.0 \\
\midrule
\multicolumn{3}{l}{\emph{Batching}} \\
\midrule
Per-device batch size    & 4                      & 4 \\
Gradient accumulation    & 4                      & 4 \\
% Gradient checkpointing   & \multicolumn{2}{l}{yes} \\
\bottomrule
\end{tabular}
\end{table}

%% ============================================================
\subsection{Supervised Fine-Tuning}
\label{app:hp_sft}

All SFT experiments use the HuggingFace TRL SFTTrainer \citep{trl} with AdamW optimizer, linear warmup over 20 steps, per-device batch size 1, max sequence length 1024.  Training data is the cleaned version of Alpaca \citep{alpaca}.

\paragraph{Default configuration.}  LoRA rank 32, $\alpha = 32$, applied to all linear layers.  Learning rate $2 \times 10^{-4}$, no weight decay, 4 gradient accumulation steps (effective batch size 4), 1{,}000 training steps.

\paragraph{Per-model deviations.}  \Cref{tab:sft_hparams} reports deviations from the default.  Larger models (Qwen2.5-7B, Llama-3.1-8B) and all Llama models required hyperparameter adjustments.  For Llama-3.2-1B/3B, adding weight decay ($\lambda{=}0.01$) stabilized training.  For Qwen2.5-7B and Llama-3.1-8B at 100K budgets, we reduced the learning rate to $5 \times 10^{-5}$ and weight decay of 0.01.

Llama-3.1-8B full LoRA baseline uses the default learning rate ($2{\times}10^{-4}$), no weight decay; the circuit and random conditions use the values shown.  All other models use the same hyperparameters across all placement methods

\begin{table}[h]
\centering
\caption{\textbf{Per-model SFT hyperparameter deviations.}  All models use rank 32, $\alpha{=}32$, warmup 20, max sequence length 1024.  ``---'' indicates the default applies ($\text{lr}{=}2{\times}10^{-4}$, $\text{wd}{=}0$, grad.\ accum.\ 4, 1{,}000 steps).}
\label{tab:sft_hparams}
\vspace{4pt}
\small
\setlength{\tabcolsep}{4pt}
\begin{tabular}{l cccc}
\toprule
\textbf{Model} & \textbf{lr} & \textbf{wd} & \textbf{grad.\ accum.} & \textbf{steps} \\
\midrule
Qwen2.5-1.5B   & ---    & ---   & ---  & --- \\
Qwen2.5-3B     & ---    & ---  & ---   & --- \\
Qwen2.5-7B     & $5{\times}10^{-5}$ & 0.01  & ---  & --- \\
Qwen3-8B       & ---    & ---    & 8    & 2{,}000 \\
Llama-3.2-1B   & ---    & 0.01 & ---   & --- \\
Llama-3.2-3B   & ---    & 0.01 & ---  & --- \\
Llama-3.1-8B   & $5{\times}10^{-5}$ & 0.01  & 8 & --- \\
\bottomrule
\end{tabular}
\end{table}

\paragraph{Alpha scaling.}  For sparse circuits, $\alpha$ is scaled from 32 to 1024 ($= 32 \times r$) to compensate for the reduced effective adapter norm when only $k \ll |\mathbf{B}|$ elements are trainable.  This matches the GRPO alpha scaling (\Cref{app:hp_grpo_qwen}).  Full LoRA and random baselines at the main-table budgets (\Cref{tab:sft_7bench}) use $\alpha = 32$.

%% ============================================================
\subsection{Circuit Discovery}
\label{app:hp_circuits}

Circuit discovery runs $N$ forward-backward passes at LoRA initialization on the training distribution to accumulate gradient statistics for the $\mathbf{B}$ matrix.  Gradients are computed with the standard supervised cross-entropy loss, regardless of whether the downstream task uses RL or SFT.  This is computationally efficient---requiring only 50 forward-backward passes---and produces circuits that transfer across both training regimes.  \Cref{tab:hp_circuit} reports the configuration.

\begin{table}[h]
\centering
\caption{\textbf{Circuit discovery hyperparameters.}  The same configuration is used for all models and tasks.}
\label{tab:hp_circuit}
\vspace{4pt}
\small
\setlength{\tabcolsep}{5pt}
\begin{tabular}{l l}
\toprule
\textbf{Parameter} & \textbf{Value} \\
\midrule
Forward-backward passes ($N$) & 50 \\
Batch size                     & 1 \\
Max sequence length            & 1024 \\
LoRA initialization            & $\mathbf{A} \sim \mathcal{N}(0, 2/d)$, $\mathbf{B} = \mathbf{0}$ \\
Loss                           & Cross-entropy (supervised) \\
Method                         & $\mathbf{B}$-zero (accumulate $\nabla_{\mathbf{B}} \mathcal{L}$ with $\mathbf{B} = \mathbf{0}$) \\
Precision                      & bfloat16 \\
\midrule
\multicolumn{2}{l}{\emph{Scoring functions}} \\
\midrule
$\hat{S}$ (gradient magnitude) & $1/N\left|\sum_{n=1}^{N} \nabla_{\mathbf{B}}^{(n)}\right|$ \\[2pt]
$\hat{F}$ (diagonal Fisher)    & $1/N\sum_{n=1}^{N} \left(\nabla_{\mathbf{B}}^{(n)}\right)^2$ \\
\midrule
\multicolumn{2}{l}{\emph{Mask application}} \\
\midrule
Selection                      & Top-$k$ elements globally across all layers \\
Masking mechanism              & Gradient hook: $\nabla \mapsto \nabla \odot \mathbf{m}$ (binary mask) \\
Frozen layers                  & Layers with zero selected elements have \texttt{requires\_grad=False} \\
\bottomrule
\end{tabular}
\end{table}

The discovery dataset matches the GRPO training distribution (GSM8K train or MATH train).  A single circuit discovery pass takes $<$10 seconds on any model up to 7B (\Cref{tab:discovery_cost}).

%% ============================================================

\paragraph{OOD evaluation (pass@$k$).}  AMC and AIME evaluations (\Cref{app:ood}) generate $n{=}64$ completions per problem at temperature 0.7 with top-$p = 0.95$ and max 2{,}048 tokens.  Pass@$k$ is computed as the fraction of problems where at least one of the first $k$ samples is correct.  Majority vote (maj@$k$) takes the most common answer among $k$ samples.

\paragraph{Grading.}  GSM8K answers are extracted after the \texttt{\#\#\#\#} delimiter.  MATH-500 answers are extracted from \verb|\boxed{}|.  Both are verified using \texttt{math\_verify} for symbolic equivalence (e.g., $\frac{1}{2}$ matches $0.5$).

\paragraph{Knockout sweep.}  Starting from a fully trained LoRA (merged into float16), we zero the top fraction of $\mathbf{B}$ entries by score and evaluate accuracy.  Fractions tested: $\{0.1\%, 1\%, 3\%, 5\%, 10\%, 20\%, 35\%, 50\%, 75\%\}$.  Random knockout uses seed 42 for reproducibility.

\clearpage
\section{Additional Experiments}
\label{app:additional_experiments}

This appendix reports experiments that supplement the main paper's SFT evaluation.  We compare additional scoring functions beyond $\hat{S}$ and $\hat{F}$, test element- vs.\ row-level selection granularity, examine budget scaling on Llama, and extend the instruction-following evaluation to larger budgets and additional benchmarks and models.
\localtableofcontents

\subsection{Extended Scoring Function Comparison}
\label{app:scoring_comparison}

The main paper focuses on two circuit scoring functions: gradient magnitude ($\hat{S}$) and diagonal Fisher ($\hat{F}$).  Here we evaluate additional importance metrics under SFT across multiple model families and tasks, at both element-level and row-level selection granularities.  All methods are computable at initialization from $N{=}50$ forward-backward passes in under 10 seconds (\Cref{app:discovery_cost}).  We find that $\hat{F}$ is the most consistent method across architectures; weight-statistic methods and row-level variants are unreliable.  This motivates the main paper's focus on $\hat{F}$ and $\hat{S}$.

\subsubsection{Additional Scoring Functions}
\label{app:scoring_defs}

All scoring functions operate on the LoRA $\mathbf{B}$ matrix at initialization ($\mathbf{B} = 0$).  We distinguish two selection granularities:

\paragraph{Element-level methods} select individual entries of $\mathbf{B}$:
\begin{itemize}[nosep,leftmargin=*]
  \item \textbf{Gradient magnitude} ($\hat{S}$): $\left|\frac{1}{N}\sum_{k=1}^N g_{k,ij}\right|$ --- directional consistency of per-example gradients.  Defined in \Cref{sec:circuit_discovery}.
  \item \textbf{Diagonal Fisher} ($\hat{F}$): $\frac{1}{N}\sum_{k=1}^N g_{k,ij}^2$ --- gradient sensitivity (second moment).  Defined in \Cref{sec:circuit_discovery}.
  \item \textbf{Weight magnitude} (Mag): $|[\mathbf{W}_0 \mathbf{A}^\top]_{ij}|$ --- selects $\mathbf{B}$ entries where the pretrained weight projected through $\mathbf{A}$ is largest.  Requires no gradient computation.
  \item \textbf{Wanda}: $|[\mathbf{W}_0 \mathbf{A}^\top]_{ij} \cdot \bar{g}_{ij}|$ --- product of projected weight magnitude and mean gradient, adapted from the structured pruning method of \citet{sun2024wanda}.
\end{itemize}

\paragraph{Row-level methods} select entire rows of $\mathbf{B}$ (all $r$ entries per hidden dimension train together):
\begin{itemize}[nosep,leftmargin=*]
  \item \textbf{Row Fisher / Row Magnitude / Row Wanda}: aggregate the element-level score across the rank dimension for each row, then select the top-$k$ rows.  This coarser granularity selects entire hidden dimensions rather than individual elements, making the selection independent of the randomly initialized $\mathbf{A}$ matrix.  Row-level selection connects to structured pruning methods that operate on neurons or channels.
\end{itemize}

\subsubsection{Alternative Scoring Methods (GSM8K)}
\label{app:scoring_elem}

\Cref{tab:scoring_elem} compares element-level scoring methods on GSM8K under both SFT and GRPO.  All SFT rows use $k{=}5$K; the GRPO row uses $k{=}10$K.

\begin{table}[h]
\centering
\caption{\textbf{Element-level scoring on GSM8K} (accuracy, \%).
Gradient-based scores ($\hat{S}$, $\hat{F}$) consistently outperform random; weight-statistic methods (Mag, Wanda) are unreliable.
Under GRPO the gap is extreme: random stays at base while circuits recover full LoRA.}
\label{tab:scoring_elem}
\vspace{4pt}
\small
\setlength{\tabcolsep}{3.5pt}
\begin{tabular}{ll ccccccc}
\toprule
\textbf{Model} & \textbf{Regime} & \textbf{Base} & \textbf{Rand} & $\hat{F}$ & $\hat{S}$ & \textbf{Mag} & \textbf{Wanda} & \textbf{Full LoRA} \\
\midrule
Qwen2.5-1.5B & SFT  & 9.5  & 13.2 & 62.2 & \textbf{64.3} & 64.6 & 63.6 & 61.4 \\
Qwen2.5-3B   & SFT  & 62.1 & 62.9 & 66.3 & 67.5 & 66.9 & 62.5 & \textbf{70.0} \\
Qwen3-4B     & SFT  & 37.6 & 78.2 & 78.7 & 79.1 & 69.4 & 75.6 & \textbf{81.4} \\
\midrule
Qwen2.5-1.5B & GRPO & 9.5  & 9.8  & 63.2 & \textbf{64.2} & 34.2 & 44.8 & 62.8 \\
\bottomrule
\end{tabular}
\end{table}

Under SFT, all informed methods beat random on every model, though Wanda drops to random-level on 3B (62.5 vs.\ 62.9).  Under GRPO the contrast sharpens: random is stuck at base (9.8\%), Magnitude recovers only half the gap (34.2\%), and Wanda reaches 44.8\%, while $\hat{S}$ and $\hat{F}$ match full LoRA.  This motivates the main paper's focus on gradient-based scores.

\subsubsection{Row-Level Scoring (GSM8K)}
\label{app:scoring_row}

Row-level methods select entire rows of $\mathbf{B}$ ($r$ entries per hidden dimension), making selection independent of the random $\mathbf{A}$ initialization.  \Cref{tab:scoring_row} reports accuracy at three budgets under SFT and one budget under GRPO.

\begin{table}[h]
\centering
\caption{\textbf{Row-level scoring on GSM8K} (accuracy, \%).
No single row method is reliable across models and regimes.
Results below base are \colorbox{red!10}{highlighted}.}
\label{tab:scoring_row}
\vspace{4pt}
\small
\setlength{\tabcolsep}{4pt}
\begin{tabular}{ll ccc cc}
\toprule
& & \multicolumn{3}{c}{\textbf{Row Budget}} & & \\
\cmidrule(lr){3-5}
\textbf{Model} & \textbf{Method} & \textbf{1.6K} & \textbf{4.8K} & \textbf{16K} & \textbf{Base} & \textbf{Full LoRA} \\
\midrule
\multicolumn{7}{l}{SFT} \\[2pt]
\multirow{3}{*}{\shortstack[l]{Qwen2.5\\-1.5B}}
  & Row Fisher     & 63.8 & \textbf{68.2} & 62.4 & \multirow{3}{*}{9.5} & \multirow{3}{*}{61.4} \\
  & Row Magnitude  & 63.2 & 67.6 & 65.7 & & \\
  & Row Wanda      & 59.4 & 65.4 & 64.8 & & \\
\midrule
\multirow{3}{*}{\shortstack[l]{Qwen2.5\\-3B}}
  & Row Fisher     & \cellcolor{red!10}59.1 & 64.3 & 66.8 & \multirow{3}{*}{62.1} & \multirow{3}{*}{70.0} \\
  & Row Magnitude  & 62.7 & \cellcolor{red!10}61.0 & 64.7 & & \\
  & Row Wanda      & 66.9 & \cellcolor{red!10}59.4 & 65.3 & & \\
\midrule
\multirow{3}{*}{\shortstack[l]{Qwen3\\-4B}}
  & Row Fisher     & 61.1 & 60.5 & 58.6 & \multirow{3}{*}{37.6} & \multirow{3}{*}{81.4} \\
  & Row Magnitude  & \textbf{79.1} & 76.8 & 62.3 & & \\
  & Row Wanda      & 70.8 & 69.3 & 62.6 & & \\
\midrule
\multicolumn{7}{l}{GRPO ($k{\approx}10$K)} \\[2pt]
\multirow{3}{*}{\shortstack[l]{Qwen2.5\\-1.5B}}
  & Row Fisher     & \multicolumn{3}{c}{41.6} & \multirow{3}{*}{9.5} & \multirow{3}{*}{62.8} \\
  & Row Magnitude  & \multicolumn{3}{c}{33.1} & & \\
  & Row Wanda      & \multicolumn{3}{c}{43.2} & & \\
\bottomrule
\end{tabular}
\end{table}

Under SFT, row methods beat base on 1.5B but are erratic on 3B (several entries below base).  Under GRPO, all row methods fall far short of element-level circuits (41--43\% vs.\ 64\%), confirming that the finer granularity of element selection is critical when gradients are noisy.

\subsubsection{Additional Budget Sweeps}
\label{app:budget_extra}

\paragraph{Qwen2.5-1.5B (OpenHermes).}
Random placement scales slowly: 55.6\% at $k{=}1$K, 58.3\% at 10K, 61.6\% at 100K, reaching full LoRA (62.6\%) only at $k{=}500$K (${\sim}$2.4\% of $B$).  At 500K, $\hat{S}$ (62.4\%) and $\hat{F}$ (62.2\%) match full LoRA.

\paragraph{Qwen3-8B (Alpaca).}
$\hat{F}$ leads at every budget tested: 70.9\% at $k{=}5$K and 71.2\% at $k{=}10$K, matching full LoRA (71.4\%) at 0.02\% of adapter parameters.  Random (70.5\%) and Row Fisher (70.4\%) trail by 0.7pp at 10K.

\paragraph{Llama-3.2-1B (Alpaca).}
At $k{=}50$K, $\hat{S}$ reaches 55.1\% (7-bench avg), above full LoRA (54.8\%) and random (53.9\%, equal to base).

\paragraph{Training dataset sensitivity.}
The 1.5B results above use OpenHermes.  On larger models (3B, 7B), OpenHermes shows mixed effects---MMLU improves but BoolQ and TruthfulQA degrade---yielding no net gain.  We use Alpaca for 3B and larger in the main paper.

\subsubsection{IFEval}
\label{app:scoring_ifeval}

\Cref{tab:scoring_ifeval} reports IFEval accuracy for Qwen2.5-1.5B under SFT at $k{=}5$K.

\begin{table}[h]
\centering
\caption{\textbf{IFEval accuracy (\%), Qwen2.5-1.5B} (SFT, $k{=}5$K).  $\hat{F}$ recovers 71--77\% of the full LoRA gain; random is indistinguishable from base.}
\label{tab:scoring_ifeval}
\vspace{4pt}
\small
\setlength{\tabcolsep}{4pt}
\begin{tabular}{l cccccc}
\toprule
\textbf{Train Data} & \textbf{Base} & \textbf{Rand} & $\hat{F}$ & \textbf{Mag} & \textbf{Wanda} & \textbf{Full LoRA} \\
\midrule
OpenHermes  & 17.0 & 16.8 & \textbf{21.4} & 21.4 & 18.5 & 22.7 \\
Alpaca      & 17.0 & 16.3 & \textbf{22.4} & 20.9 & 18.3 & 24.6 \\
\bottomrule
\end{tabular}
\end{table}

\subsubsection{Consistency Summary}
\label{app:scoring_consistency}

\Cref{tab:scoring_consistency} tallies how often each scoring method outperforms random across all configurations in \Cref{tab:scoring_elem,tab:scoring_row}.

\begin{table}[h]
\centering
\caption{\textbf{Scoring method consistency across models and training regimes.}
$\hat{S}$ and $\hat{F}$ are the only methods that never fail.}
\label{tab:scoring_consistency}
\vspace{4pt}
\small
\setlength{\tabcolsep}{5pt}
\begin{tabular}{l cc}
\toprule
\textbf{Method} & \textbf{Wins / Total} & \textbf{Failure Modes} \\
\midrule
\multicolumn{3}{l}{Element-level} \\[2pt]
$\hat{S}$            & \textbf{4/4} & --- \\
$\hat{F}$            & \textbf{4/4} & --- \\
Magnitude             & 3/4 & Collapses under GRPO (34.2\% vs.\ 9.8 rand) \\
Wanda                 & 2/4 & $\approx$ random on 3B SFT; partial under GRPO \\
\midrule
\multicolumn{3}{l}{Row-level} \\[2pt]
Row Fisher            & 2/4 & Below base on 3B SFT; 41.6\% under GRPO \\
Row Magnitude         & 2/4 & Below base on 3B SFT; 33.1\% under GRPO \\
Row Wanda             & 2/4 & Below base on 3B SFT; budget-sensitive \\
\bottomrule
\end{tabular}
\end{table}

The pattern is clear: $\hat{S}$ and $\hat{F}$ are the only scoring methods that outperform random on every architecture and training regime tested.  Weight-statistic methods (Magnitude, Wanda) work under SFT's stable gradients on some models but break under GRPO, where Magnitude collapses to 34.2\%---worse than Wanda (44.8\%) and far below circuits (64.2\%).  Row-level methods, which select entire hidden dimensions independent of $\mathbf{A}$, are inconsistent even under SFT (multiple entries below base on 3B) and fall far short under GRPO (33--43\% vs.\ 64\% for element-level $\hat{S}$).

This motivates the main paper's focus on $\hat{S}$ and $\hat{F}$: they are the only methods that reliably identify which parameters matter, regardless of whether the training signal is stable (SFT) or noisy (GRPO).

\clearpage

\section{Limitations}
\label{app:limitations}
The GRPO experiments focus on mathematical reasoning (GSM8K, MATH-500); whether the same gradient structure dichotomy holds for other RL domains like code generation, dialogue, or open-ended tasks with softer reward signals, remains to be tested.  More broadly, the SFT/RL contrast we observe may depend on properties of the reward function and RL algorithm; different reward designs or policy gradient estimators could produce gradient structures that fall between the extremes we characterize.

Looking forward, custom sparse kernels for the LoRA layer could realize the $O(k)$ adapter cost in practice, and scaling circuit discovery beyond 7B, where memory and communication savings become more impactful is a natural next step.

\section{Broader Impact}
\label{app:broad}

This work studies how parameter selection affects the efficiency and effectiveness of fine-tuning large language models. By identifying small subsets of parameters (“circuits”) that capture useful training signal, our approach can significantly reduce the compute and memory required for adaptation. This may lower the barrier to entry for researchers and practitioners with limited resources, enabling broader participation in developing and customizing language models.

At the same time, improving the efficiency of model adaptation can accelerate the deployment of domain-specific systems, including those used in high-stakes settings such as education, healthcare, and decision support. As with prior work on parameter-efficient fine-tuning, our method does not inherently address issues such as bias, hallucination, or misuse; these risks remain and may be amplified if models are more easily adapted without careful oversight.

Our findings also suggest that training dynamics differ substantially between supervised and reinforcement learning settings. While this may guide the design of more robust and efficient training methods, it could also be used to optimize systems for specific objectives (e.g., reward maximization) without improving general reliability. We therefore recommend that applications using such techniques be evaluated carefully, particularly in settings where alignment with human intent is critical.

Overall, we view this work as a step toward more efficient and accessible model adaptation, with benefits for research accessibility and deployment efficiency, alongside the need for continued attention to safety and responsible use.

%%%%%%%%%%%%%%%%%%%%%%%%%%%%%%%%%%%%%%%%%%%%%%%%%%%%%%%%%%%%

\end{document}